\documentclass[lettersize,journal,twoside]{IEEEtran}
\usepackage{amsmath,amsfonts}
\usepackage{algorithm}
\usepackage[noend]{algpseudocode}
\usepackage{array}
\usepackage{textcomp}
\usepackage{stfloats}
\usepackage{url}
\usepackage{verbatim}
\usepackage{graphicx}
\usepackage{booktabs, multirow, makecell, subcaption}
\usepackage{cite}
\usepackage{hyperref}
\usepackage[table, svgnames, dvipsnames]{xcolor}
\usepackage[frozencache,cachedir=.]{minted}
\usepackage[figuresright]{rotating}

\usepackage{amsthm}
\usepackage{xcolor}

\begin{document}
\newcommand{\blue}{\textcolor{black}}
\newcommand{\red}{\textcolor{red}}
\newcommand{\GR}[1]{\cellcolor{Gainsboro!60}} % coloring table 

\title{NeFL: Nested Model Scaling for Federated Learning with System Heterogeneous Clients}

\author{Honggu~Kang, Seohyeon~Cha, Jinwoo~Shin, Jongmyeong~Lee, Joonhyuk~Kang % ,~\IEEEmembership{Staff,~IEEE,}
        % <-this % stops a space
\thanks{Manuscript received XXX, XX, 2024; revised XXX, XX, 2024; accepted XXX, XX, 2024.
% This work was supported in part by the Ministry of Science and ICT (MSIT), South Korea, through the Information Technology Research Center (ITRC) Support Program supervised by the Institute of Information and Communications Technology Planning and Evaluation (IITP) under Grant IITP-2024-2020-0-01787 and Grant IITP-2024-RS-2023-00259991.
Recommended for acceptance by Dr. X. \textit{(Corresponding author: Joonhyuk
Kang.)}}
\thanks{H.~Kang is now with Samsung Electronics, Suwon 16677, South Korea (e-mail: honggu.kang@samsung.com).}
% He was with the School of Electrical Engineering, Korea Advanced Institute of Science and Technology (KAIST), Daejeon 305-701, South Korea (e-mail: honggu.kang@samsung.com).}
\thanks{S.~Cha is now with the University of Texas at Austin, TX 78712, USA (e-mail: seohyeon.cha@utexas.edu)}
\thanks{H.~Kang, S.~Cha and J.~Lee were with the School of Electrical Engineering, Korea Advanced Institute of Science and Technology (KAIST), Daejeon 305-701, South Korea.}
\thanks{J.~Shin and J. Kang are with the School of Electrical Engineering, Korea Advanced Institute of Science and Technology (KAIST), Daejeon 305-701, South Korea (e-mail: jinwoos@kaist.ac.kr; jhkang@ee.kaist.ac.kr).}
}

\markboth{IEEE Transactions on Mobile Computing,~Vol.~X, No.~X, X~2024}%
{Kang \MakeLowercase{\textit{et al.}}: NeFL: Nested Model Scaling for Federated Learning with System Heterogeneous Clients}


\maketitle

\begin{abstract}
Federated learning (FL) enables distributed training while preserving data privacy, but stragglers—slow or incapable clients—can significantly slow down the total training time and degrade performance. To mitigate the impact of stragglers, system heterogeneity, including heterogeneous computing and network bandwidth, has been addressed.
While previous studies have addressed system heterogeneity by splitting models into submodels, they offer limited flexibility in model architecture design, without considering potential inconsistencies arising from training multiple submodel architectures.
We propose \textit{nested federated learning (NeFL)}, a generalized framework that efficiently divides deep neural networks into submodels using both depthwise and widthwise scaling.
% NeFL interprets forward propagation as solving ordinary differential equations (ODEs) with adaptive step sizes, allowing for dynamic submodel architectures.
To address the \textit{inconsistency} arising from training multiple submodel architectures, NeFL decouples a subset of parameters from those being trained for each submodel. An averaging method is proposed to handle these decoupled parameters during aggregation.
NeFL enables resource-constrained devices to effectively participate in the FL pipeline, facilitating larger datasets for model training. \blue{Experiments demonstrate that NeFL achieves performance gain, especially for the worst-case submodel compared to baseline approaches (7.63\% improvement on CIFAR-100)}. Furthermore, NeFL aligns with recent advances in FL, such as leveraging pre-trained models and accounting for statistical heterogeneity. Our code is available online \cite{Kang}.

% Federated learning (FL) is a promising approach in distributed learning keeping privacy. However, during the training pipeline of FL, slow or incapable clients (i.e., stragglers) slow down the total training time and degrade performance. To mitigate the impact of stragglers, system heterogeneity, including heterogeneous computing and network bandwidth, has been addressed.
% Previous studies tackle the system heterogeneity by splitting a model into submodels, but with less degree-of-freedom in terms of model architecture. We propose {\it nested federated learning (NeFL)}, a generalized framework that efficiently divides a model into submodels using both depthwise and widthwise scaling. NeFL is implemented by interpreting forward propagation of models as solving ordinary differential equations (ODEs) with adaptive step sizes. To address the {\it inconsistency} that arises when training multiple submodels of different architecture, we decouple a few parameters from parameters being trained for each submodel. Furthermore, to address the decoupled parameters, we propose averaging method for NeFL.
% NeFL enables resource-constrained clients to effectively join the FL pipeline and the model to be trained with a larger amount of data. Through a series of experiments, we demonstrate that NeFL leads to significant performance gains, especially for the worst-case submodel.
% Furthermore, we demonstrate NeFL aligns with recent studies in FL, regarding pre-trained models of FL and the statistical heterogeneity. Our code is available online\footnote{https://honggkang.github.io/nefl}.
\end{abstract}

\begin{IEEEkeywords}
Federated learning, system heterogeneity, pruning, ordinary differential equations, inconsistency.
\end{IEEEkeywords}

\section{Introduction}

\IEEEPARstart{T}{he} success of deep learning owes much to vast amounts of training data where a large amount of data comes from mobile and internet-of-things (IoT) devices. However, privacy regulations on data collection have become a critical concern, potentially limiting further advances in deep learning. % \cite{DataProtection, dou2021federated}.
A distributed machine learning framework, federated learning (FL) is getting attention to address these privacy concerns. FL enables model training by collaboratively leveraging the vast amount of data on clients while preserving data privacy. Rather than centralizing raw data, FL collects trained model weights from clients, that are subsequently aggregated on a server by a method such as FedAvg \cite{mcmahan2017communication}. \blue{FL has shown its potential, and several studies have explored to utilize it more practically \cite{hong2022efficient, he2020fedgkt, tang2024reliable, makhija2022federated, zhuang2022divergenceaware, he2021ssfl}.}

Despite its promising prospects, challenges related to system heterogeneity have been identified \cite{kairouz2021advances,Li2020federated}. Clients with varying resources such as computing power, communication bandwidth, and memory introduce stragglers, i.e., slow or incapable clients, which can degrade FL performance. While waiting for stragglers can elongate training time, dropping out these clients may bias the model towards data from resource-rich clients. Hence, extensive studies have focused on FL frameworks that accomodate clients with heterogeneous resources. A naive approach involves reducing the global model size to accommodate resource-poor clients; however, this may lead to performance degradation due to limited model capacity \cite{kaplan2020scaling, zhai2022scaling}. Clearly, FL with a single global model does not efficiently meet the needs of heterogeneous clients.

While vanilla FL trains a model with a fixed size and structure, we propose {\it Nested Federated Learning (NeFL)}, a method that trains multiple submodels of adaptive sizes to meet dynamic requirements (e.g., memory, computing and bandwidth dynamics) of heterogeneous clients.
\blue{Although previous studies have explored training submodels \cite{horvath2021fjord,diao2021heterofl,kim2023depthfl,ilhan2023scalefl}, they have limitations. These limitations include unbalanced scaling and the lack of consideration for discrepancies between submodels.
Our approach scales down the global model into submodels either widthwise, depthwise, or both, to achieve flexibility.
% While NeFL embraces existing studies of federated learning in a nested manner \cite{horvath2021fjord,diao2021heterofl,kim2023depthfl,ilhan2023scalefl}, the proposed scaling method provides more degrees-of-freedom (DoF) to scale down a model than previous studies.
The increased flexibility enhances the efficiency of submodels in terms of the number of parameters \cite{mingxing2019efficient} and offers greater adaptability in model size and computation cost.
The scaling is motivated by interpreting model forward propagation as solving ordinary differential equations (ODEs). This ODE-based perspective led us to introduce submodels with {\it learnable step size parameters} and the concept of {\it inconsistency}.
Additionally, we propose a parameter averaging method within NeFL, which includes \textit{Nested Federated Averaging (NeFedAvg)} for averaging \textit{consistent parameters} and FedAvg for averaging \textit{inconsistent parameters} across submodels.}
%  \cite{horvath2021fjord,diao2021heterofl,kim2023depthfl,ilhan2023scalefl}.
% Notably, NeFL outperforms the Top-1 accuracy of baselines by \textbf{7.63\%} for the worst-case submodel of ResNet110.

Furthermore, we verify if NeFL aligns with the recently proposed ideas in FL: (i) pre-trained models improve the performance of FL in both identically independently distributed (IID) and non-IID settings \cite{chen2023on} and (ii) simply rethinking the model architecture improves the performance, especially in non-IID settings \cite{qu2022rethinking}.
Through a series of experiments we observe that NeFL outperforms state-of-the-art (SOTA) baselines sharing the advantages of recent studies.

The main contributions of this study are summarized:
as follows:
\blue{
\begin{itemize}
    \item We introduce a novel model scaling technique that adjusts both the width and depth of neural networks, inspired by ODE solvers.
    \item We propose a parameter averaging method that accommodates architectural discrepancies among diverse submodels by incorporating inconsistent parameters.
    \item We provide comprehensive performance evaluations of NeFL through a series of experiments, demonstrating its effectiveness and validating its applicability.
\end{itemize}
}

% \begin{itemize}
%     \item We introduce a novel model scaling method--across both width and depth-- inspired by ODE solvers to effectively address system heterogeneity.
%     % The method generally scales a model , ensuring balanced and efficient learning.
%     \item We propose a parameter averaging method designed for different submodels. Our approach uniquely incorporates inconsistent parameters to manage architectural discrepancies among diverse submodels.
%     \item We provide comprehensive performance evaluations of NeFL through a series of experiments, demonstrating its effectiveness and validating its applicability.
% \end{itemize}

The remainder of this paper is organized as follows. Section \ref{sec:related_works} briefly reviews related works and Section \ref{sec:background} describes the background. We present our proposed method, \textit{NeFL} consisting of model scaling method and the submodel averaging method in Section \ref{sec:method}. Section \ref{sec:experiment} presents the experimental results and Section\ref{sec:conclusion} concludes the paper.
\begin{table}[t]
\caption{Summarization of main notations}
\centering
{\renewcommand{\arraystretch}{1}
\begin{tabular}{cc}
\toprule
Symbol & Definition \\
\midrule
$T$ & Number of total communication rounds \\
$E$ & Number of local epochs in each round \\
$N_s$ & Number of submodels \\
$\mathcal{C}_t$ & Clients set in communication round t\\
$\theta_j=\{\theta_{\textrm{c},j}, \theta_{\textrm{ic},j} \}$ & Parameters of submodel $j$ \\
$\theta_\textrm{c}$ & Concistent parameters of submodel $j$ \\
$\theta_\textrm{ic}$ & Inconsistent parameters of submodel $j$ \\
$\textbf{w}_{i}$ & Weights of client $i$\\
$\textbf{F}_{G}$ & Global model \\
$\textbf{F}_{j}$ & $j$-th submodel \\
$\gamma_W$ & Widthwise scaling ratio \\
$\gamma_D$ & Depthwise scaling ratio \\
$s_j$ & step size parameter of $j$-th block \\
$\mathcal{M}_k$ & Weights set from clients who trained $k$-th submodel \\
$\mathcal{M}$ & $\{\mathcal{M}_k\}_{k=1}^{N_s}$ \\
$\phi_{j,k}$ &$j$-th block weights of $k$-th submodel\\
$\eta$ &Learning rate\\
$J$ &Loss function\\
\bottomrule
\end{tabular}}
\vspace{-0.15in}
\end{table}

\section{Related Works}\label{sec:related_works}
\blue{In this section, we introduce several methods for reducing model size and present FL studies that incorporate these techniques. Previous FL studies have limitations or take approaches that are orthogonal to our closely related works, which we discuss in Section \ref{sec:closely-related}. We then highlight our contributions in relation to these closely related works.}

\subsection{Model Size Reduction}
The real-world deployment of deep neural networks depends on their efficiency such as low latency and high energy efficiency. There have been rigorous studies to increase the efficiency by reducing the model size.

Knowledge distillation (KD) aims to compress the model by transferring knowledge from a large teacher model to a smaller student model \cite{hinton2015distilling}. KD trains the student model to mimic the output predictions of the teacher model.
% on a given dataset.
% This knowledge transfer process enables the student model to achieve better performance compared to directly training it on the original data.

Pruning reduces the size and computational complexity of neural networks by removing redundant or less significant parameters \cite{han2015learning, he2017channel, blalock2020state,lee2020signal}. This reduces the model's memory footprint by removing storage for pruned (zeroed-out) weights and improves computational efficiency during inference by leveraging sparsity in weight matrices.
% However, pruning may introduce accuracy trade-offs despite its significant compression and inference acceleration benefits.

Quantization aims to reduce the numerical precision of neural network weights and activations represented in high-precision floating-point formats by mapping them to lower bitwidth fixed point or integer formats \cite{jacob2018quantization, krishnamoorthi2018quantizing}. This reduction in bitwidth results in memory savings and requires less storage space compared to higher-precision representations. Moreover, lower-precision computations often demand fewer computational resources, leading to speedups and reductions in energy consumption.
% However, similar to pruning, quantization introduces accuracy trade-offs, necessitating careful strategies to balance compression and model performance.

Split learning (SL) \cite{vepakomma2018split} utilizes external computing resources like cloud or edge servers. It partitions the neural network into client-side and server-side segments, separated by a cut layer. Clients train their part of the network up to the cut layer with local data, sending intermediate outputs to the server. The server then trains the remaining layers, backpropagates gradients to the cut layer, and communicates them back to clients.
% However, SL requires significant communication of intermediate outputs and gradients between clients and servers.

One study introduced the concept of \textit{hypernetworks}, smaller neural networks used to generate weights for larger neural networks \cite{ha2017hypernetworks}. This approach, explored in the context of recurrent neural networks, generates adaptive weights for each time step or layer, rather than using shared weights across all steps, thus leading to memory savings.
Similarly, NeuralODEs represent deep neural network models that parameterize the derivative of the hidden state using a neural network, unlike traditional architectures with discrete hidden layers. This enables a constant memory cost, as the same network is reused at each solver step. Additionally, NeuralODEs provide adaptive computation, balancing computational speed and numerical precision \cite{chen2018node, ODEbook}.

\subsection{Federated Learning with Resource Constraints}
% FL, with its decentralized approach to model training across multiple client devices, brings about various overheads. These include communication costs incurred in exchanging model updates between clients and the central server, alongside increased memory and computational capability on the client devices required for training models locally using their data. Here, we introduce various techniques that have been developed to enhance FL under such resource constraints.

% KD
Several studies have explored the integration of KD within FL, primarily focusing on two aspects: (i) reducing communication cost \cite{wu2022fedkd,seo2020federated} and (ii) combining knowledge from clients with different architectures \cite{zhu2021datafree, he2020fedgkt, lin2020ensemble, afonin2022towards}.
FedKD \cite{wu2022fedkd} suggests sharing only the student model to reduce communication, employing adaptive mutual distillation based on prediction performance. FD \cite{seo2020federated} uploads locally averaged logit vectors periodically, allowing clients to learn from globally averaged logits for knowledge transfer.
FedGKT \cite{he2020fedgkt} presents an edge-computing method that employs KD. This involves edge nodes transferring their knowledge to a large server-side model. Moreover, the server sends its predicted soft labels to the edge clients.
% , which then utilize these soft labels to train their local models using a KD-based loss function, thus facilitating the transfer of knowledge from the server to the edge clients.
FedDF \cite{lin2020ensemble} introduces a model fusion that employs ensemble distillation to combine models with different architecture. However, it necessitates the availability of a shared proxy dataset, which could be a public dataset or synthetic data generated from well-trained generative models, to serve as a reference for KD.
% It demonstrates the limitations of KD that require the common data samples for teacher model and student model.
FedGen \cite{zhu2021datafree} overcomes dataset limitations by training a generative model but relies on a well-trained predictor. In \cite{afonin2022towards}, clients transmit their models instead of datasets for knowledge transfer, mitigating architectural differences but increasing communication burden. \blue{However, those KD studies introduce additonal burdens such as another teacher model, logits, proxy dataset and generative model.}

% pruning
There has been research employing pruning within FL.
PruneFL \cite{jiang2022pruningFL} adapts model size during FL.
% to reduce communication/computation overhead and minimize training time while maintaining accuracy. 
Specifically, it involves initial pruning at a selected client and further pruning during the FL process, adapting model size to reduce approximate empirical risk.
FedLTN \cite{Mugunthan2022fedltn} leverages the Lottery Ticket Hypothesis to learn sparse and personalized lottery ticket networks (LTNs).
% FedLTN preserves clients' batch-norm statistics, performs post-pruning without rewinding, and aggregates LTNs.
FedMP \cite{jiang2024fedmp} leverages adaptive model pruning 
% to improve computation and communication efficiency across heterogeneous workers. It uses
by using a Multi-Armed Bandit algorithm to adaptively determine pruning ratios for heterogeneous clients. This allows each worker to train and transmit a sub-model tailored to its capabilities, accelerating training without sacrificing accuracy. 
% However, these previous works evaluates the performance on a single model and the effectiveness of a each submodel is not evaluated.
% However, these previous works \cite{jiang2022pruningFL,Mugunthan2022fedltn,jiang2024fedmp} evaluate the performance on a global model architecture, without assessing the effectiveness of the submodels in test time.
\blue{These existing pruning works \cite{jiang2022pruningFL,Mugunthan2022fedltn,jiang2024fedmp} evaluate the performance of a global model, which is fundamentally different from closely related works in Section \ref{sec:closely-related}.}

% [quantization]
Several studies have focused on compressing uploaded gradients by quantization or sparsification to deal with the communication bottleneck \cite{rothchild2020fetchsgd, haddadpour2021federated}. FetchSGD \cite{rothchild2020fetchsgd} proposes a method that compresses model updates using a Count Sketch algorithm.
Meanwhile, FL with bitwidth heterogeneity across devices has been studied. ProWD \cite{yoon2022bitwidth} introduces a trainable weight dequantizer at the central server that progressively reconstructs low-bitwidth weights into higher bitwidths and full-precision weights. ProWD further selectively aggregates model parameters across bit-heterogeneous weights.
\blue{Note that quantization and NeFL are orthogonal, saying that our method combined with quantization can be conducted.}

% [splitfed]
\blue{While there have been studies exploring the integration of SL with FL \cite{thapa2022splitfed,shin2023fedsplitx,he2020fedgkt,jiang2024federated}}, these approaches introduce additional communication overhead compared to traditional FL. Unlike FL, where only trained model parameters are transmitted, these SL-FL approaches necessitate the transmission of intermediate outputs corresponding to the number of data samples, resulting in a higher communication cost.
One study \cite{liang2020think} has introduced a method called LG-FedAvg to split the model and decouple model layers into global and local layers, reducing the number of parameters involved in communication.
\blue{However, these SL-FL require the computing server to offload training tasks which makes burden.}

% HyperNetworks
A significant line of research exploring the integration of hypernetworks into FL concentrates on mitigating the inherent challenges posed by the non-independent and identically distributed (non-IID) characteristics of client data distributions. These approaches introduce a decoupled strategy, where a subset of model parameters are disentangled from the global model and implemented via hypernetworks, allowing for adaptive personalization to the unique data distributions across individual clients \cite{ma2022layer,li2023fedtp,chen2021bridging}.
Meanwhile, pFedHN \cite{shamsian2021personalized} shares parameters via a joint hypernetwork, generating distinct network parameters for each client, with unique embedding vectors yielding personalized model weights.
\blue{Each client only shares its parameters to update the hypernetworks. These hypernetworks generate parameters for a single model architecture, which cannot handle heterogeneous model architectures.}
% Since the parameters of hypernetworks can be larger than those of the clients' networks, the generated parameters by the hypernetwork are not transferred. Consequently, communication costs remain unaffected by the size of the hypernetwork.
% Hypernetwork parameters can be larger than the clients' network parameters that hypernetwork generate are not transferred. Therefore, communication costs remain unaffected by the hypernetwork size.

\begin{table}[!t]
\centering
{ %\scriptsize
\caption{Summarization of NeFL and baselines}
\label{tab:NeFLsummary}
{\renewcommand{\arraystretch}{1}
\begin{tabular}{lccc}
\toprule
& \multirowcell{2}{\textbf{Depthwise}\\\textbf{scaling}} & \multirowcell{2}{\textbf{Widthwise}\\\textbf{scaling}} & \multirowcell{2}{\textbf{Adaptive}\\\textbf{step sizes}} \\ \\ \midrule
DepthFL \cite{kim2023depthfl} & \checkmark &  & \\ \midrule
FjORD \cite{horvath2021fjord}, HeteroFL \cite{diao2021heterofl} & & \checkmark & \\ \midrule
ScaleFL \cite{ilhan2023scalefl} & \checkmark &  \checkmark  & \\ \midrule
\textbf{NeFL (ours)} & \textcolor{red}{\checkmark} & \textcolor{red}{\checkmark} & \textcolor{red}{\checkmark} \\ \bottomrule
\end{tabular}}}
\vspace{-0.1in}
\end{table}
\subsection{Closely Related Works\label{sec:closely-related}}
% [submodels]
\blue{There have been various approaches of extracting submodels from a global model based on clients' capabilities to address system heterogeneity in FL. They do not require proxy dataset or computing server. Furthermore, they provide several submodels to be utilized during test time according to a client's environment.}
% Various approaches have been proposed to address system heterogeneity in FL by splitting the global network based on clients' capabilities.
FjORD \cite{horvath2021fjord} and HeteroFL \cite{diao2021heterofl} scale down a global model into submodels tailored to different client capabilities through widthwise pruning. On the other hand, DepthFL \cite{kim2023depthfl} introduces to scale down a single model depthwise into heterogeneous submodels. This approach incorporates an additional bottleneck layer and an independent classifier for each submodel to compensate for depthwise scaling. While these previous works mainly focus on the scaling of global model in single dimension, it has been studied that \textit{deep and narrow} models, as well as \textit{shallow and wide} models, are inefficient in terms of the number of parameters or floating-point operations (FLOPs) \cite{sergey2016wide,mingxing2019efficient}.
%Previous studies \blue{in centralized learning} have shown that carefully balancing the depth and width of a model improves its performance \cite{sergey2016wide,mingxing2019efficient,ilhan2023scalefl}. Therefore, a balanced submodel for each client should be considered for FL as well.
\blue{Furthermore, previous works \cite{horvath2021fjord,diao2021heterofl,kim2023depthfl,ilhan2023scalefl} aggregate submodel parameters without accounting for discrepancies among them.} 

\blue{Tackling these challenges,} our proposed NeFL introduces two key enhancements. First, it employs a versatile scaling method that incorporates both widthwise and depthwise scaling for submodels, as summarized in Table~\ref{tab:NeFLsummary}, ensuring a well-balanced configuration. Second, NeFL introduces the concept of inconsistency and proposes an aggregation method that explicitly considers the discrepancies between submodels during the aggregation process.

\section{Background and Motivation}\label{sec:background}

\blue{In this section, we introduce ordinary differential equations (ODE) solver, that provides insights to interpret our proposed scaling method.}
% We propose a scaling method inspired by solving ordinary differential equations (ODEs) in a numerical way (e.g., Euler method).
Consider an initial value problem to find an output $y$ at step $t$ given $\frac{dy}{dt}=f(t,y)$ and $y(t_0)=y_0$. We can compute output $y$ at any point by integration: $y = y_0 + \int_{t_0}^t f(t,y) dt$. It can be approximated by Taylor’s expansion as $ y = y_0 + f(t_0,y_0) \left(t-t_0\right)$ and approximated after more steps as follows: % + \mathcal{O}\left(\left(t-t_0 \right)^2 \right)
\begin{equation}\label{eqn:euler}
\begin{split}
y_{n+1} = y_n + hf(t_n, y_n)= y_0 + hf(t_0, y_0) + \cdots \\ + hf(t_{n-1}, y_{n-1}) + hf(t_n, y_n),
\end{split}
\end{equation}
where $h$ denotes the step size. An ODE solver can compute with less steps by using a larger step size as: $y_{n+1} = y_0 + 2hf(t_0, y_0) + \cdots  + 2hf(t_{n-1}, y_{n-1})$ when $n$ is odd and the odd steps are skipped. The results would be numerically less accurate than fully computing with a smaller step size.

\begin{figure}[!t]
\vspace{-0.3in}
  \centering
  \includegraphics[width=2.8in]{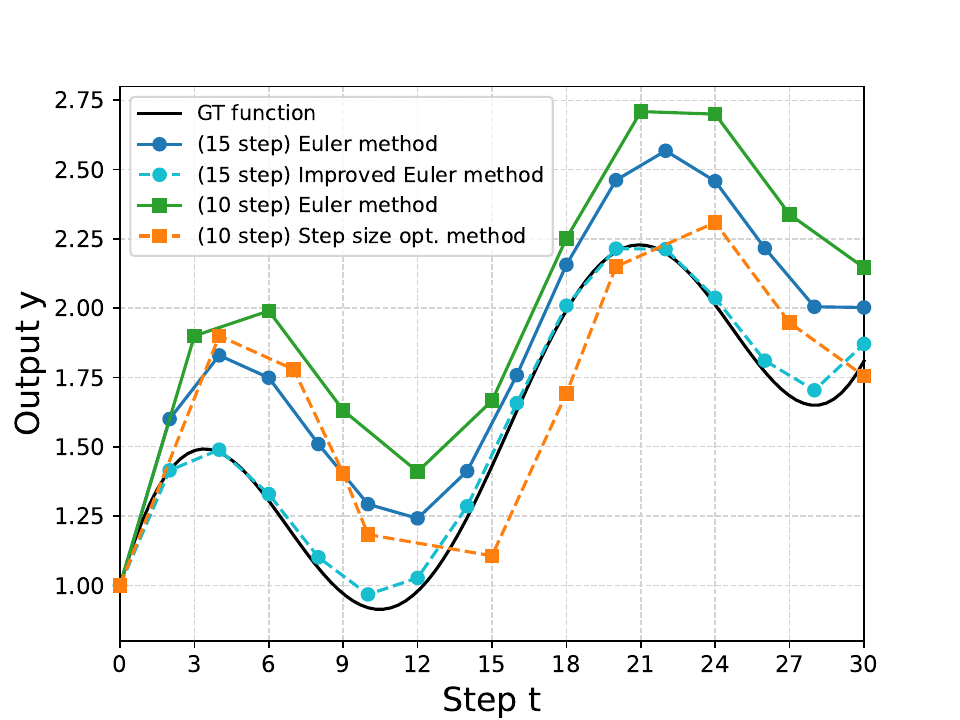}
  % \vspace{-0.1in}
  \caption{{Toy example of ODE solver showing the effect of number of steps and step sizes}}
  \label{fig:ode-toy}
  \vspace{-0.23in}
\end{figure}

For better understanding, we present a toy example of ODE solvers in Figure \ref{fig:ode-toy}. The black line represents the ground truth (GT) function $0.1t+\sin(0.2t)+\cos(0.3t)$, while the blue line denotes a discretized approximation of the GT function. This approximation was generated by an ODE solver with a step size of $h=2$ and calculated over 15 points. Although the green line, computed using 10 steps with a step size of $h=3$, exhibits larger numerical error compared to the blue line with 15 steps, the orange dashed line with 10 steps and optimized step sizes demonstrates how optimization of step sizes can mitigate numerical errors. Furthermore, contrasting the sky-blue colored dashed line implemented by the improved Euler method \cite{ODEbook} with a step size of $h=3$ to the blue line with the same number of steps, further illustrates how optimizing $dy/dt$ can contribute to reducing numerical errors. % (i.e., a full neural network) In summary, this example elucidates why each submodel in NeFL can effectively perform with fewer parameters.

Recent works have connected the ODE solver and residual networks based on their similarity in output computation \cite{chang2018multilevel,he2016deep}. Modern deep neural networks stack a few residual blocks that contain skip-connections that bypass the residual layers. An output representation of a residual block is written as $\mathbf{Y}_{j+1} = \mathbf{Y}_{j} + F_j(\mathbf{Y}_j, \phi_j)$, where $\mathbf{Y}_j$ is the feature map at the $j$-th layer, $\phi_j$ denotes the $j$-th block’s network parameters, and $F_j$ represents a residual module of the $j$-th block. An output of a residual block is rewritten as follows:
\begin{equation}
\begin{split}
\mathbf{Y}_{j+1} = \mathbf{Y}_{j} + F_j(\mathbf{Y}_j, \phi_j) = \mathbf{Y}_{0} + F_0(\mathbf{Y}_0, \phi_0) + \cdots \\
+ F_{j-1}(\mathbf{Y}_{j-1}, \phi_{j-1}) + F_{j}(\mathbf{Y}_j, \phi_j).
\end{split}
\end{equation}

\begin{figure}[!t]
  \centering
  \includegraphics[width=3.1in]{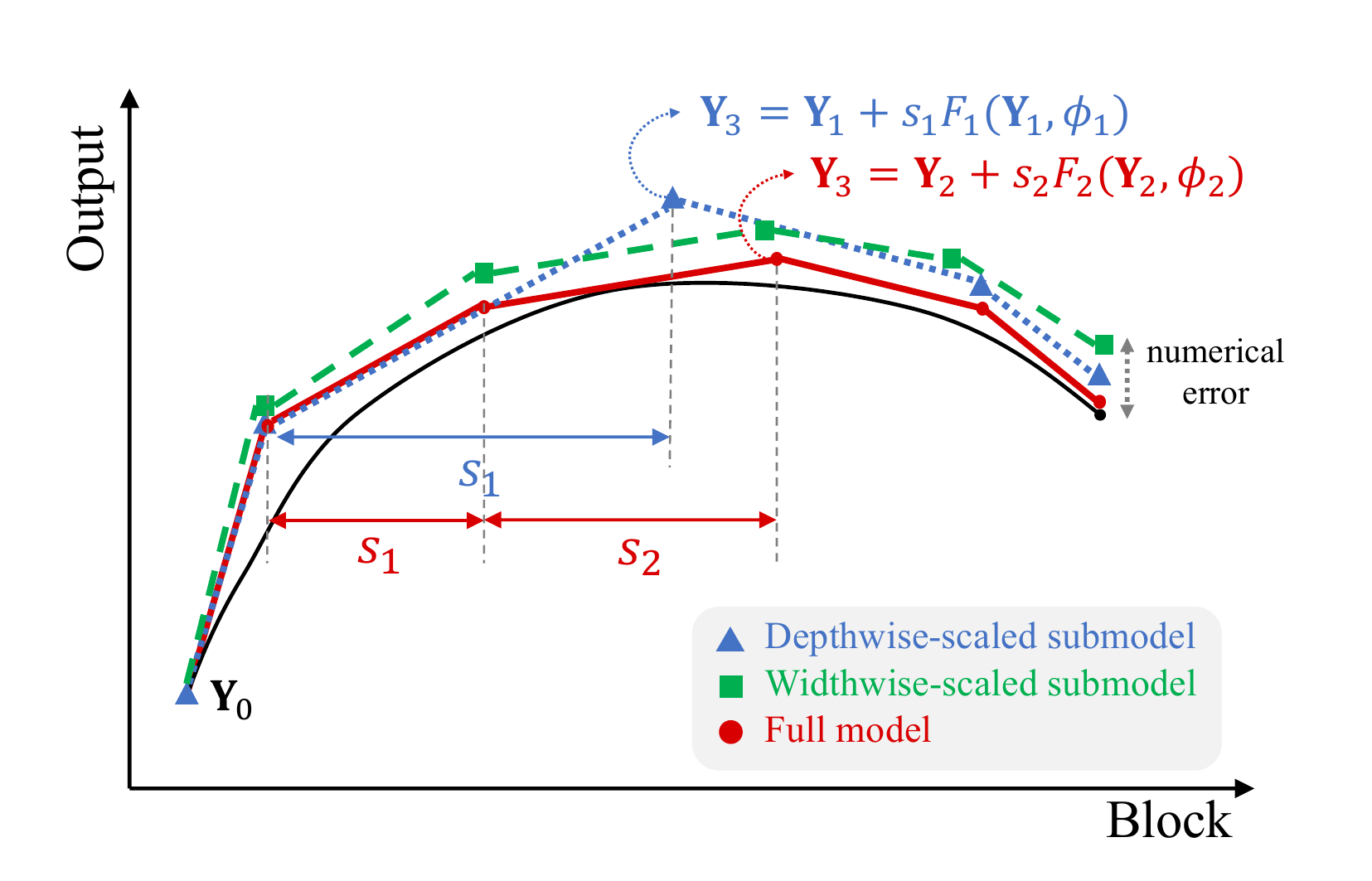}
  \caption{{Example of output computation when applying widthwise/depthwise model scaling inspired by ODE solver}}
  \label{fig:graph}
  \vspace{-0.15in}
\end{figure}
\begin{figure*}[!t]
  \centering
  \includegraphics[width=5.9in]{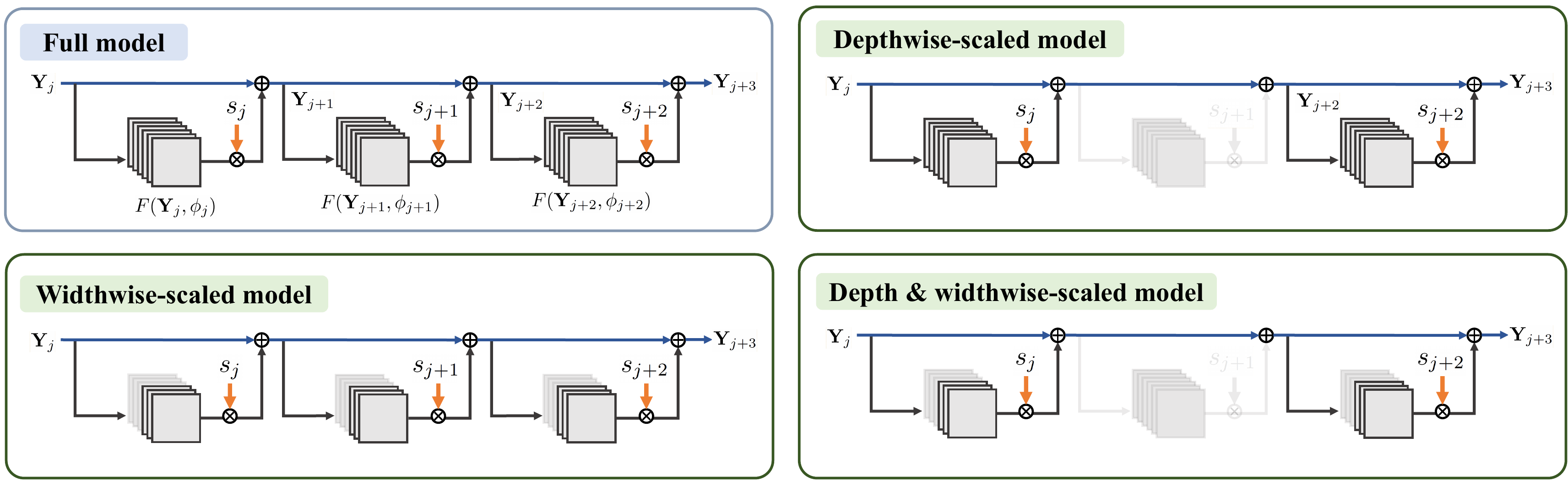}
  \caption{{Scaling method in both depth and/or width dimensions inspired by ODEs}}
  \label{fig:scaling}
  \vspace{-0.2in}
\end{figure*}

% Motivated by interpreting the neural networks as solving ODEs, some residual blocks can be omitted during forward propagation.
Our model scaling method, inspired by ODE solvers, is demonstrated in Figure \ref{fig:graph} and Figure \ref{fig:scaling}. While the black line is interpreted as the GT function to be approximated, the red line depicts the output approximated by a full neural network. Then, $\mathbf{Y}_3=\mathbf{Y}_0+F_0+F_1+F_2$ could be approximated by $\mathbf{Y}_3=\mathbf{Y}_0+F_0+2F_1$ omitting the $F_2$ block. We use this block omitting idea to our depthwise scaling.
% A width-wise scaled residual block represents the optimal $k$-rank approximation of the original (full) block \cite{horvath2021fjord}.
The blue line represents the depthwise-scaled submodel, where the model skips to compute block $F_2(\cdot)$. To compensate for the omitted block, a larger step size can be multiplied to $F_1(\mathbf{Y}_1,\phi_1)$ \cite{chang2018multilevel}. The green line represents the widthwise-scaled submodel, which has fewer parameters $\phi$ in each block (e.g., reducing the size of filters in convolutional layers). According to the theorem that a widthwise-scaled model with fewer parameters is an approximation of a model with more parameters \cite{eckart1936approximation}, while the output from a widthwise-scaled model exhibits larger numerical error. The theoretical background behind widthwise scaling, provided by the Eckart-Young-Mirsky theorem \cite{eckart1936approximation} is elaborated in Section~\ref{subsec:model_scaling}.
% Here, we introduce our proposed step size parameters for scaled submodels to effectively perform with fewer parameters.
% this example elucidates why each submodel in NeFL can effectively perform with fewer parameters.

In addition, we propose {\it learnable step size parameters} \cite{touvron2021going, bachlechner2021rezero}. Our proposed step size parameters make scaled submodels effectively perform with fewer parameters. Instead of pre-determining the step size parameters ($h$ in the (\ref{eqn:euler})), we let step size parameters trained along with the network parameters. For example, when we omit $F_2$ block, output is formulated as $\mathbf{Y}_3=\mathbf{Y}_0+s_0 F_0+ s_1 F_1$ where $s_i$’s are also optimized\footnote{\blue{In this study, we set the initial values to $s_i = 1$ for all $i$. We observed that using larger initial values led to performance degradation.}}. This enables depthwise-scaled submodel to minimize the numerical error. % without compromising numerical error. 
Note that the learnable step sizes can also alleviate numerical error induced by widthwise scaling. \blue{These parameters can be designated as inconsistent to address discrepancies between submodels, as discussed in the following section.}

\begin{figure*}[!t]
  \begin{center}
  \includegraphics[width=5.9in]{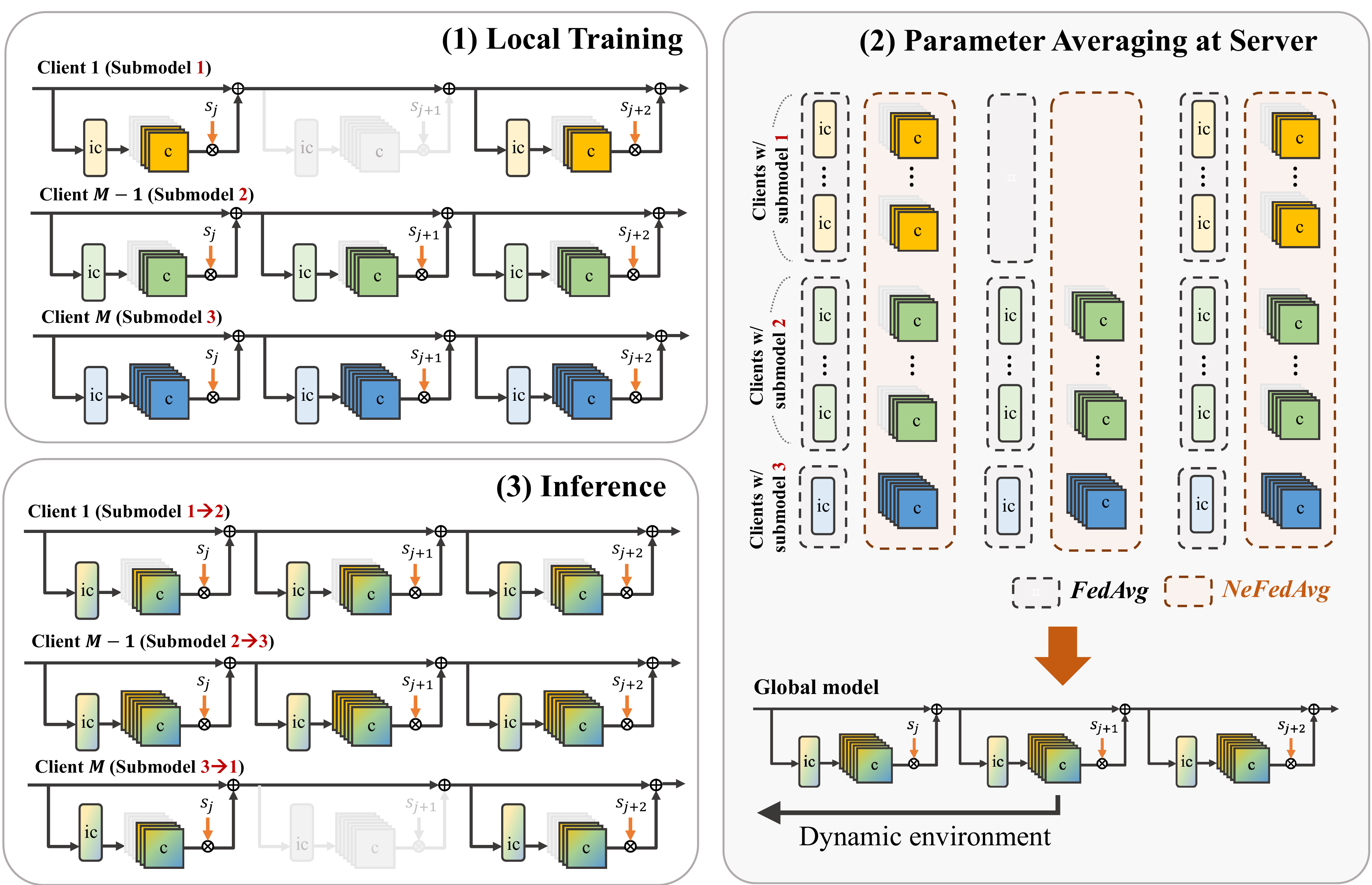}
  \end{center}
  % \vspace{-0.1in}
  \caption{\textbf{NeFL framework}. (1) Clients select submodels that are scaled in both width and/or depth dimensions. (2) The NeFL server aggregates weights of submodels by a proposed parameter averaging algorithm that addresses consistent and inconsistent parameters separately. (3) Clients can choose one of the submodels based on their varying capabilities in a dynamic environment.}
  \label{fig:head}
% \vspace{-0.18in}
\vspace{-0.1in}
\end{figure*}

\section{Proposed Algorithm}\label{sec:method}
{\it NeFL} is FL framework that accommodates resource-poor clients. Instead of requiring every client to train a single global model, NeFL allows clients to train submodels that are scaled in both width and depth dimensions based on dynamic nature of their environments. This flexibility enables more clients, even with constrained resources, to participate in the FL, thereby making the model be trained with a more data.

\begin{algorithm}[!t]
\caption{NeFL: Nested Federated Learning}\label{alg:NeFL}
{\footnotesize
\textbf{Input:} Global model $\mathbf{F}_G$, total communication round $T$, the number of local epochs $E$
\begin{algorithmic}[1]
\State NeFL server scales $\mathbf{F}_G$ into submodels $\{\mathbf{F}_k\}_{k=1}^{N_s}$ 
\For{$t \gets 0$ to $T-1$} // global iterations %\Comment{Global iterations}
    \State NeFL server broadcasts the weights $\{\theta_{\textrm{c},N_s},\theta_{\textrm{ic},1},\dots\theta_{\textrm{ic},N_s}\}$ to clients in $\mathcal{C}_t$
    \For{client $i$ in $\mathcal{C}_t$} // parallelly 
    \State Client $i$ determines a submodel among $\{\mathbf{F}_k\}_{k=1}^{N_s}$ to train
    \State $\mathbf{w}_i \gets \theta_j\in\{\theta_1,\dots,\theta_{N_s}\}$ % $\forall i \in \mathcal{C}_t$
    % \Comment{Client $i$ determines what submodel to train according to the environments $\forall i \in \mathcal{C}_t$}
    \For{$k \gets 0$ to $E-1$} // local iterations %\Comment{Local iterations}
        \State $\mathbf{w}_{i}\leftarrow \mathbf{w}_{i}-\eta \nabla_{\mathbf{w}_{i}} J(\mathbf{w}_{i})$
        % \State Client $i$ updates ${\mathbf{w}_i}$ % $\forall i \in \mathcal{C}_t$
    \EndFor
    \State Client $i$ transmits the weights $\mathbf{w}_{i}$ to the NeFL server % $\forall i \in \mathcal{C}_t$
    \EndFor
    \State $\{\theta_{\textrm{c},N_s}, \theta_{\textrm{ic},1}, \dots\theta_{\textrm{ic},N_s}\} \gets \texttt{ParamAvg}(\{\mathbf{w}_i \}_{i \in \mathcal{C}_t}$) \Comment{Algorithm \ref{alg:NeFedAvg}}
\EndFor
\end{algorithmic}}
\end{algorithm}
\blue{NeFL operates through three key stages, as illustrated in Figure~\ref{fig:head}. \textbf{(1) Local Training:} Each heterogeneous client selects a submodel at each iteration based on specific requirements, which may include factors like runtime memory constraints, CPU/GPU bandwidth, or communication conditions. The chosen submodel is trained on the client's local data, and the updated weights are then sent back to the server. \textbf{(2) Parameter Averaging at Server:} At the server, the NeFL framework aggregates the parameters from these diverse submodels using the \textit{NeFedAvg} procedure. In Figure~\ref{fig:head}, the consistent parameters, denoted by `c', are shared across all submodels and are averaged together. In contrast, the inconsistent parameters, denoted by `ic', are unique to each submodel and are averaged separately within each submodel group. This dual averaging approach allows the NeFL server to effectively combine information from various submodels while maintaining their individual characteristics. \textbf{(3) Inference:} During inference, clients can select a submodel from the trained global model based on their current operational requirements, such as memory usage, latency, or available processing power. This adaptability allows clients to balance performance with their specific constraints, ensuring functionality across diverse environments.}

Specifically, referring to Algorithm \ref{alg:NeFL}, NeFL scales a global model $\mathbf{F}_G$ into $N_s$ submodels $\mathbf{F}_1,\dots,\mathbf{F}_{N_s}$ with corresponding weights $\theta_1,\dots,\theta_{N_s}$. Without loss of generality, we suppose $\mathbf{F}_G=\mathbf{F}_{N_s}$. 
During each communication round $t$, each client in subset $\mathcal{C}_t$ (which is subset of $M$ clients) selects one of the $N_s$ submodels based on their respective environments and load the according weights.
Each client trains the model for local epochs $E$ by $\mathbf{w}_{i}\leftarrow \mathbf{w}_{i}-\eta \nabla_{\mathbf{w}_{i}} J(\mathbf{w}_{i})$ where $\eta$ denotes learning rate and $J$ denotes loss function. Then, clients transmit their trained weights $\{\mathbf{w}_i\}_{i \in \mathcal{C}_t}$ to the NeFL server, which is then aggregated into $\{\theta_{\textrm{c},N_s},\theta_{\textrm{ic},1},\dots\theta_{\textrm{ic},N_s}\}$ where $\theta_{\textrm{c},k}$ and $\theta_{\textrm{ic},k}$ denote {\it consistent} and {\it inconsistent} parameters of a submodel $k$ \blue{($\theta_k=\{\theta_{\textrm{c},k}, \theta_{\textrm{ic},k}\}$)}.
The aggregated weights are then distributed to the clients in $\mathcal{C}_{t+1}$\footnote{\blue{The server can either broadcast the global parameters or multicast only the specific submodel parameters that each client will train in the next round, in which case clients must inform the server of their chosen submodel beforehand.}}, and this process is repeated for the total number of communication rounds.

In the following section, we provide detailed explanations of our algorithm, categorizing it into \textit{model scaling} and \textit{parameter averaging}.

\begin{algorithm}[!t]
{\footnotesize 
\caption{\texttt{ParamAvg}}\label{alg:NeFedAvg}
\textbf{Input:} Trained weights from clients $\mathbf{W}=\{\mathbf{w}_i \}_{i \in \mathcal{C}_t}$
\begin{algorithmic}[1]
\For{submodel index $k$ in $\{1,\dots,N_s\}$}
    \State $\mathcal{M}_k \gets \{\mathbf{w}_i | \mathbf{w}_i=\theta_k\}_{i\in\mathcal{C}_t}$
\EndFor
\For{block $\phi_j$ in $\theta_{\textrm{c},N_s}$} // \texttt{NeFedAvg} \Comment{Consistent parameters}
    \State $\mathcal{M}' \gets \mathcal{M}=\{\mathcal{M}_k\}_{k=1}^{N_s}$
    \For{$k$ in $\{1,\dots,N_s\}$}
        \State $\mathcal{M}' \gets \mathcal{M}'\setminus \mathcal{M}_k$ if $\phi_j \notin \theta_{\textrm{c},k}$
    \EndFor
    \State $k' \gets 0$, $\phi_{j,0} = \emptyset$
    \For{$k$ in $\{k|\mathcal{M}_k \in \mathcal{M}'\}$}
        \State $ \phi_{j,k} \setminus \phi_{j,k'} \gets \frac{\sum_{\{i|\mathbf{w}_i \in \bigcup_{l\geq k, \mathcal{M}_l\in \mathcal{M}'}\mathcal{M}_l\} }  {\phi_{j,k}^i} \setminus \phi_{j,k'}^i}{\sum_{l\geq k, \mathcal{M}_l\in \mathcal{M}'}{|\mathcal{M}_l|}} $
        \State $k' \gets k$
    \EndFor
    % \State $\phi_j \gets \bigcup_{\{k|\mathcal{M}_k \in \mathcal{M}'\}} \phi_{j,k}$
\EndFor
\State $\theta_{\textrm{c},N_s} \gets \bigcup_j \phi_j$
\For{$k$ in $\{1,\dots,N_s\}$} // \texttt{FedAvg} \cite{mcmahan2017communication} \Comment{Inconsistent parameters}
    \State $\theta_{\textrm{ic},k} \gets \sum_{\{i|\mathbf{w}_i \in \mathcal{M}_k\}} \theta_{\textrm{ic},k}^i /|\mathcal{M}_k|$
\EndFor
\end{algorithmic}}
\end{algorithm}

\subsection{Model Scaling} \label{subsec:model_scaling}
\blue{We propose a global network scaling method that integrates both widthwise and depthwise scaling. We scale the model widthwise by a factor of $\gamma_W$ and depthwise by a factor of $\gamma_D$. For instance, a global model with $\gamma = \gamma_W \gamma_D = 1$ can be reduced to a submodel with $\gamma = 0.25$ by choosing $\gamma_W = 0.5$ and $\gamma_D = 0.5$. Any combination of $\gamma_W$ and $\gamma_D$ that satisfies $\gamma_W \gamma_D = 0.25$ is permissible, provided $0 < \gamma_W, \gamma_D \leq 1$. We first apply $\gamma_D$ to determine which layers or blocks to omit, and then apply $\gamma_W$ to further reduce the model's width.
This flexible widthwise and depthwise scaling offers greater freedom in adjusting model sizes. We elaborate on our scaling strategies in the following sections.}

\subsubsection{\textbf{Depthwise Scaling}} \label{subsec:depth_scaling}
% However, it has also been noted that residual connections prevent blocks to learn useful representations \cite{mingxing2019efficient}.
Residual networks, such as ResNets \cite{he2016deep} and ViTs \cite{dosovitskiy2021an}, have gained popularity due to their significant performance improvements in {\it deep} networks. \blue{Incorporating the learnable step size parameters, the output of a residual block can be seen as a function multiplied by a step size $s_j$ as $    \mathbf{Y}_{j+1} = \mathbf{Y}_{j} + s_j F(\mathbf{Y}_j, \phi_j)=\mathbf{Y}_{0} + \sum_{j} s_j F(\mathbf{Y}_j, \phi_j)$.} Note that ViTs' forward operation with skip connections can be represented in a similar way by choosing $F(\cdot)$ as either self-attention (SA) layers or feed-forward networks (FFN):
\begin{equation}
    \begin{split}
\mathbf{Y}_{j+1} &= \mathbf{Y}_{j} + s_j SA(\mathbf{Y}_j, \phi_j),\\
\mathbf{Y}_{j+2} &= \mathbf{Y}_{j+1} + s_{j+1} FFN(\mathbf{Y}_{j+1}, \phi_{j+1}).
    \end{split}
\end{equation}
\blue{Referring to Figure \ref{fig:scaling}, the depthwise scaling can be achieved by ommiting some of the blocks (e.g., residual blocks of ResNets or encoder blocks of ViTs).} For instance, ResNet architectures include a downsampling layer followed by residual blocks, with each block containing two convolutional layers. Specifically, ResNet18 has 8 residual blocks, while ResNet34 has 16 residual blocks. ViT/B-16 has 12 encoder blocks where embedding patches are used as inputs for the subsequent transformer encoder blocks.
\blue{To scale a submodel, we selectively remove blocks such that the number of parameters is adjusted by the scaling factor $\gamma_D$. It is possible to have multiple submodels that satisfy the scaling ratio $\gamma_D$. In such cases, blocks are skipped to closely match the scaling factor $\gamma_D$.}

% according to system requirements of a client.
We can observe that skipping a few blocks in a network still allows it to operate effectively \cite{chang2018multilevel}. This observation is in line with the concept of stochastic depth proposed in \cite{huang2016stochasticDepth}, where a subset of blocks is randomly bypassed during training and the all of the blocks are used during testing.
% Figure \ref{fig:l2norm} presents the L2-norm of the output of layers. The flat line across layer denotes that the submodel does not have blocks so that its output has same value until the forwarding existing blocks. We can observe the L2-norm value does not significantly vary without blocks. (Details on the figure should be added)

The step size parameters $s$'s can be viewed as dynamically training how much forward pass should be made at each step. They allow each block to adaptively contribute to useful representations, and larger step sizes are multiplied to blocks that provide valuable information. Referring to Figure \ref{fig:graph} and numerical analysis methods (e.g., Euler method, Runge-Kutta methods \cite{ODEbook}), adaptive step sizes rather than uniformly spaced steps, can effectively reduce numerical errors.
{\it Note that DepthFL \cite{kim2023depthfl} is a special case of the proposed depthwise scaling. \blue{Unlike our approach, DepthFL uses fixed step sizes set to one and employs distinct classifiers for each submodel.}}
% \footnote{We suppose a non-contiguous depthwise scaling so that we implement DepthFL without auxiliary bottleneck layers and self-distillation. The details are described in the Appendix \ref{appx:exp_details}.}

\subsubsection{\textbf{Widthwise Scaling}} \label{subsec:width_scaling}

\begin{figure}[t]
\centering
\begin{subfigure}{0.24\textwidth}
    \centering
    \includegraphics[width=1.1\textwidth]{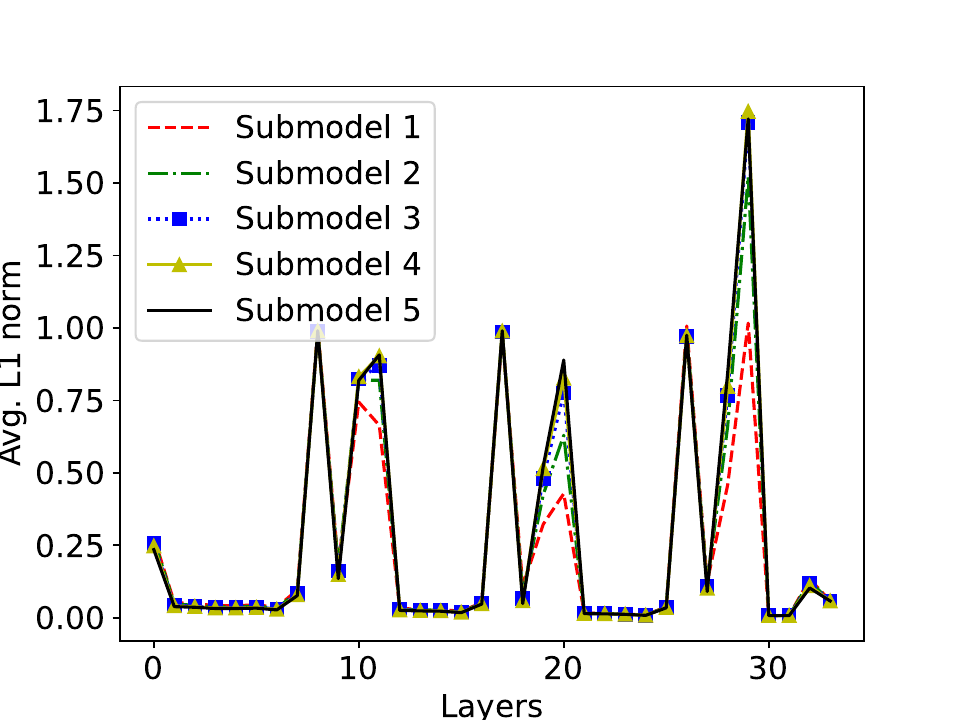}
    % \hspace{0.1cm}
    \caption{}
    \label{fig:l1norm}
\end{subfigure}
\begin{subfigure}{0.24\textwidth}
    \centering
    \includegraphics[width=1.1\textwidth]{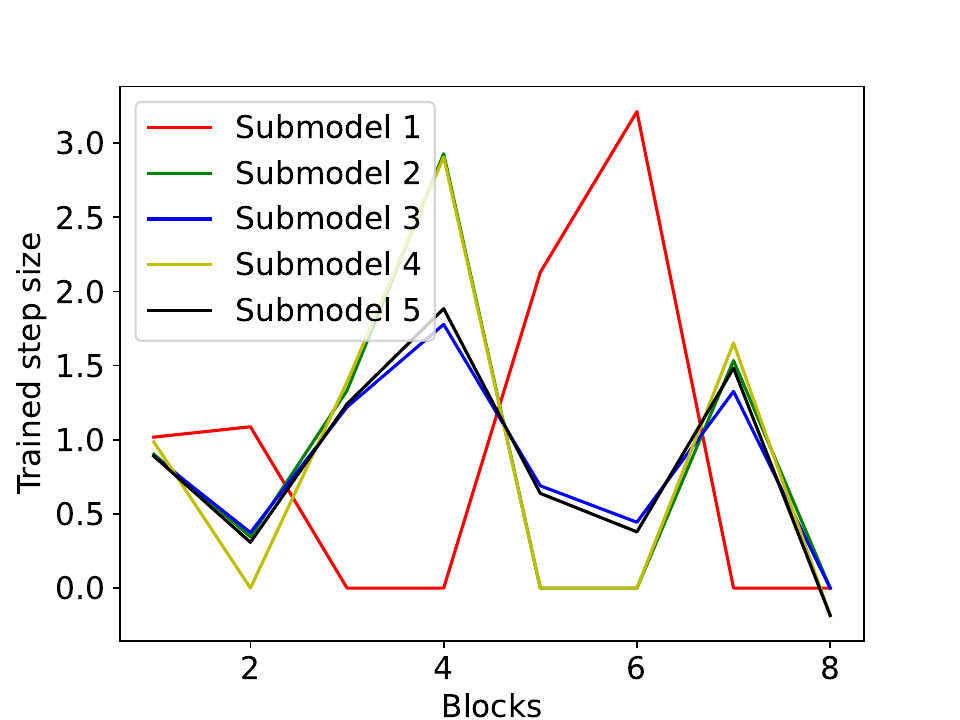}
    \caption{}
    \label{fig:steps}
\end{subfigure}
\caption{(a) Average L1 norm of trained submodel weights and (b) step sizes of trained submodels}
\label{fig:pruning}
\vspace{-0.1in}
\end{figure}
% (Each block can have different number of the parameters and computational complexity.)
Previous studies have addressed reducing the model size in width dimension \cite{li2017pruning, Liu2017learning}. In the context of NeFL, widthwise scaling is employed to slim down the network by utilizing structured contiguous pruning (ordered dropout in \cite{horvath2021fjord}) along with learnable step sizes.
\blue{Referring to Figure \ref{fig:scaling},} we apply contiguous channel-based pruning to the convolutional networks \cite{he2017channel} and node removal to the fully connected layers. \blue{A block of a submodel with a scaling factor of $\gamma_W$ includes neurons or filters indexed from $\{0, 1, \dots, \lceil \gamma_W K \rceil -1\}$ where $K$ denotes the number of parameters in the block.
While our framework permits each block to have a different widthwise scaling ratio, for simplicity and without loss of generality, we apply the same $\gamma_W$ to all blocks.}
The parameters of narrow-width submodel constitute a subset of the parameters of the wide-width model. Consequently, the parameters of the narrowest submodel are trained by every client, while the exclusive parameters of the larger-width model are trained less frequently. This approach ensures that the parameters of the narrowest model capture the most of useful representations. 
% Note that each block stated in Section \ref{subsec:depth_scaling} can have different widths, but we assume throughout the paper that every block has the same widthwise scaling ratio $\gamma_W$ without loss of generality.

We can obtain further insights from a toy example. Given an optimal linear neural network $y=Ax$ and data sampled from uniform distribution $x\sim\mathcal{X}$, the optimal widthwise-scaled submodel $y=A_W x$ is the best $k$-rank approximation\footnote{Note that we have learnable step sizes that can further improve the performance of widthwise scaled submodel over best $k$-rank approximation (refer to Figure~\ref{fig:ode-toy}).} by Eckart–Young–Mirsky theorem \cite{horvath2021fjord, eckart1936approximation}. Given a linear neural network $A$ of rank $k$, widthwise-scaled model of scaling ratio $\gamma_W$ has rank of $\lceil\gamma_W k\rceil$. Then,
$
\min \mathbb{E}_{x \sim \mathcal{X}} \left\|A_W x-A x\right\|^2_F = \min \left\|A_W -A \right\|^2_F,
$
where $F$ denotes the Frobenius norm.
In our framework, a similar tendency is observed. Inspired by the magnitude-based pruning \cite{han2015learning, li2017pruning}, we present the L1 norm of weights averaged by the number of weights in each layer of five trained submodels in Figure \ref{fig:l1norm}. The narrowest model, submodel 1, exhibits a comparable L1 norm trend to the widest model, with the gap between submodels diminishing as the model's width increases, akin to the concept of the best $k$-rank approximation.
The narrowest model likely captures the most essential representation possible, whereas larger models with additional parameters gather useful but comparatively less crucial information.

\subsection{Parameter Averaging} \label{sec:NeFedAvg}
\subsubsection{Inconsistency}

NeFL facilitates the training of multiple submodels, which are then aggregated at the server. However, submodels with different architectures—varying in width and depth—exhibit distinct characteristics \cite{acar2021federated,chen2022when,Li2020fedprox,santurkar2018how}. As illustrated in Figure \ref{fig:graph},
% if depthwise-scaled model omits a block, it can be compensated by optimizing step size of adjacent blocks. For widthwise-scaled model, numerical errors are induced from each block of less parameters, and they can be mitigated by optimizing step size for each block. 
we can infer that each submodel requires different step sizes to compensate for the numerical errors according to its respective model architecture.
% For instance, depthwise-scaled models may require adjustments in the step sizes of adjacent blocks to compensate for omitted blocks, while widthwise-scaled models may need to mitigate numerical errors that arise from reduced parameters in each block by optimizing the step size for each.
Furthermore, submodels with different model architectures have different loss landscapes.
% \cite{li2018visualizinglosslandscapeneural} Consider the situation where losses are differentiable.
A stationary point of a global model is not necessarily the stationary points of individual submodels \cite{acar2021federated, Li2020fedprox}. \blue{The varying trainability of different network architecture may lead to deviate from the optima.}

\blue{We refer to these different characteristics between submodels as {\it inconsistency}.
The inconsistency led us to introduce a decoupled set of parameters, {\it inconsistent parameters} \cite{liang2020think, arivazhagan2019federated}. A few layers or parameters of a global model can be treated as the inconsistent parameters.
Unlike previous methods that averaged parameters across all clients regardless of submodel architecture, our approach averages inconsistent parameters only among clients sharing the same submodel architecture.
In this study, we treat step size parameters and batch normalization (BN) layers as inconsistent parameters. Notably, BN can improve the Lipschitzness of loss, which is cruicial for the convergence of FL \cite{santurkar2018how, Li2020On}.
Figure \ref{fig:steps} presents the trained step size parameters of submodels, that are different across the submodels in our experiments.
% We further demonstrate the benefits of inconsistent parameters by ablation study in Section \ref{par:incons-step}.
}

In contrast, \textit{consistent parameters} are shared across all submodels. The averaging of consistent parameters differs from FedAvg \cite{mcmahan2017communication} because the consistent parameters of a submodel are the subset of the parameters of the global model.
Note that the parameters of the submodels are denoted as $\theta_1=\{\theta_{{\textrm{c}},1},\theta_{{\textrm{ic}},1}\}, \dots ,\theta_{N_s}=\{\theta_{{\textrm{c}},{N_s}},\theta_{{\textrm{ic}},{N_s}}\}$, and
% where $\theta_{\textrm{c}}$ denotes consistent parameters and $\theta_{{\textrm{ic}}}$ denotes inconsistent parameters.
\blue{the global parameters, which include all the parameters of the submodels, encompass $\theta_{{\textrm{c}},N_s}, \theta_{{\textrm {ic}},{1}}, \dots, \theta_{{\textrm{ic}},{N_s}}$. Note that $\theta_{\textrm{c},k} \subset \theta_{\textrm{c},N_s} \, \forall \, k$.}

\subsubsection{Parameter averaging}
We propose an averaging method for models scaled in both width and depth while addressing inconsistency, as detailed in Algorithm \ref{alg:NeFedAvg} (illustrated in Figure \ref{fig:head}).
The NeFL server processes locally trained weights $\mathbf{W}=\{\mathbf{w}_i\}_{i\in\mathcal{C}_t}$ from clients, verifying and organizing them according to their corresponding submodels. These weights are sorted into sets $\mathcal{M} = \{ \mathcal{M}_1, \dots, \mathcal{M}_{N_s}\}$, where $ \mathcal{M}_k $ contains the weights from the clients who trained the $k$-th submodel and $\mathcal{M}$ is a set of uploaded weights sorted by submodels.

For inconsistent parameters, \textit{FedAvg} \cite{mcmahan2017communication} is applied (lines 12-13 in Algorithm~\ref{alg:NeFedAvg}), where these parameters are averaged only among clients that share the same submodel type. This approach is feasible because the weights of clients with the same submodel type have the same size. Specifically, $\theta_{\textrm{ic},k}^{i}$ represents the inconsistent parameters of the $k$-th submodel, which have been updated by the $i$-th client.

Consistent parameters are averaged by {\it Nested Federated Averaging (NeFedAvg)}, which aggregates the parameters block by block ($\phi_{j}$).
% The parameters are averaged by weights from the clients who trained the parameters every communication round.
% since a submodel can have parameters of a block or not.
For each block, the server determines which submodels contain that block (line 6 in Algorithm \ref{alg:NeFedAvg}) and then averages the parameters in a nested, width-wise manner. Parameters shared across submodels are averaged based on the number of clients who trained them. In Algorithm \ref{alg:NeFedAvg}, $\phi_{j,k}^{i}$ denotes $j$-th block parameters of $k$-th submodel that $i$-th client updated.

% The parameters of a block with the smallest width are included in the parameters of a block with larger width. Hence, the block parameters are averaged by weights of clients ($\phi_{j,k}^{i}$; $j$-th block parameters of $k$-th submodel that $i$-th client updated) who updated the parameters.
% Meanwhile, the parameters of a block with the largest width is only contained in a submodel that has the largest width.
% Thus, the parameters are averaged by clients whose submodels have the block with the largest width. 

For example, consider five submodels. Suppose the first block is included in submodels 1, 3, and 5, with submodels 1 and 5 each trained by two clients and submodel 3 by three ($|\mathcal{M}_1|=|\mathcal{M}_5|=2$, and $|\mathcal{M}_3|=3$). The averaging process works as follows: Parameters unique to submodel 5 ($\phi_{1,5}\setminus \phi_{1,3}$) are averaged within $\mathcal{M}_5$; parameters in submodel 3 but not in submodel 1 ($\phi_{1,3}\setminus\phi_{1,1}$) are averaged within $\mathcal{M}_5\cup \mathcal{M}_3$; and parameters in submodel 1 ($\phi_{1,1}$) are averaged within ($\mathcal{M}_5\cup \mathcal{M}_3 \cup \mathcal{M}_1$). This ensures consistent parameter averaging, taking into account both depth and width across submodels.

\section{Experiments\label{sec:experiment}}

In this section, we present the performance of our proposed NeFL compared to baseline methods. We detail our experimental setup, compare performance across various model architectures, explore the use of pre-trained models, analyze statistical heterogeneity with ViTs, and conduct an ablation study on widthwise and depthwise scaling, as well as the impact of inconsistent parameters and learnable step sizes.

\subsection{Experimental Setup}
\subsubsection{Datasets and models}
We evaluated the performance of NeFL on four public datasets: \textit{CIFAR-10} \cite{cifar10}, \textit{CIFAR-100} \cite{cifar10}, \textit{CINIC-10} \cite{darlow2018cinic10}, and \textit{SVHN} \cite{svhn}.
% The CIFAR10 dataset consists of 60000 $32\times32\times3$ color images (50000 for training and 10000 for testing) divided into 10 classes . The CINIC dataset contains 270000 $32\times32\times3$ color images (90000 for training and 90000 for validation and testing) of 10 classes. The SVHN dataset is obtained from house numbers in Google Street view images in 10 classes ($32\times32\times3$ color image where 73257 and 26032 digits are used for training and testing respectively.).
We used four representative residual models, ResNet18, 34, 56, and 110 \cite{he2016deep}. ResNet18 and 34 consist of four layers, while ResNet56 and 110 have three layers. These layers are composed of blocks with different channel sizes, specifically (64, 128, 256, 512) for ResNet18/34 and (16, 32, 64) for ResNet56/101. We additionally report results of Wide ResNet101 \cite{sergey2016wide} and ViT-B/16 \cite{dosovitskiy2021an} specifically on CIFAR10 to raise a valuable discussion on model architecture. Wide ResNet101 comprises four layers of bottleneck blocks with channel sizes (128, 256, 512, 1024) and ViT-B/16 consists of twelve layers, with each containing blocks comprising self-attention (SA) and feedforward networks (FFN) \cite{dosovitskiy2021an}.
\begin{table*}[!t]
\caption{Results of NeFL and  baselines with five submodels on five datasets under {IID} settings. We report test performance including Top-1 classification accuracies (\%) for the worst-case submodel and the average performance over five submodels.} \label{tab:other-dataset}
\centering
\resizebox{.75\textwidth}{!}{
% \resizebox*{!}{0.21\textheight}{
\begin{tabular}{@{}clcccccccccccccc@{}}
\toprule
% \multicolumn{4}{c}{\textbf{CIFAR-100}} &  & \\ \midrule
\multirow{3}{*}{\textbf{Model}}    & \multirow{3}{*}{\textbf{Method}} & \multicolumn{2}{c}{\textbf{CIFAR-10}} & & \multicolumn{2}{c}{\textbf{CIFAR-100}} & & \multicolumn{2}{c}{\textbf{CINIC-10}} & & \multicolumn{2}{c}{\textbf{SVHN}}\\ \cmidrule(l){3-13}
      && Worst & Avg && Worst & Avg && Worst & Avg && Worst & Avg\\ \midrule
\multirow{7}{*}{ResNet18}
      & HeteroFL \cite{diao2021heterofl}        & 80.62 & 84.26 && 41.33 & 47.09 && 67.55 & 70.40 && 91.82 & 93.46 \\ \cmidrule(l){2-13}
      & FjORD \cite{horvath2021fjord}          & 85.12 & 87.32 && 49.29 & 52.67 && 71.95 & 74.98 && 94.31 & 93.97\\ \cmidrule(l){2-13}
      & DepthFL \cite{kim2023depthfl}        & 64.80 & 82.44 && 31.68 & 49.56 && 54.51 & 71.42 && 91.54 & 93.97\\ \cmidrule(l){2-13}
      & ScaleFL \cite{ilhan2023scalefl}        & 79.47 & 85.18 && 41.00 & 49.76 && 70.55 & 73.85 && 93.15 & 94.53 \\ \cmidrule(l){2-13}
      &  \GR{0}\textbf{NeFL (ours)} & \GR{0}\textbf{86.86} & \GR{0}  \textbf{87.88} &\GR{0}&\GR{0}\textbf{52.63}   & \GR{0}\textbf{53.62} &\GR{0}&\GR{0}\textbf{74.16} &\GR{0}\textbf{75.29} &\GR{0} & \GR{0}\textbf{94.45}   & \GR{0}\textbf{94.94}\\ \midrule
\multirow{7}{*}{ResNet34}
      & HeteroFL \cite{diao2021heterofl}       & 79.51 & 83.16 && 34.96 & 39.75 && 67.39 & 69.62 && 89.86 & 92.39\\ \cmidrule(l){2-13}
      & FjORD \cite{horvath2021fjord}         & 85.12 & 87.36 && 47.59 & 50.7 && 71.58 & 74.19 && 93.83 & 94.63\\ \cmidrule(l){2-13}
      & DepthFL \cite{kim2023depthfl}        &  25.73 & 75.30 && 14.51 & 46.79 && 32.05 & 67.04 && 74.33 & 89.96\\ \cmidrule(l){2-13}
      & ScaleFL \cite{ilhan2023scalefl}        & 54.72 & 81.05 && 22.62 & 46.41 && 49.69 & 69.43 && 86.46 & 93.21\\ \cmidrule(l){2-13}
      & \GR{0}\textbf{NeFL (ours)} & \GR{0}\textbf{87.71} &\GR{0}\textbf{89.02} &\GR{0}&\GR{0}\textbf{55.22} & \GR{0}\textbf{56.26} &\GR{0}&\GR{0}\textbf{75.02} & \GR{0}\textbf{76.68} &\GR{0}&\GR{0}\textbf{94.72}   & \GR{0}\textbf{95.22}\\ \bottomrule
\end{tabular}}
\vspace{-0.1in}
\end{table*}
\subsubsection{Baselines} We compared NeFL with SOTA FL baselines that adopt different model scaling methods to address system heterogeneity: HeteroFL \cite{diao2021heterofl} and FjORD \cite{horvath2021fjord} use widthwise scaling, DepthFL \cite{kim2023depthfl} uses depthwise scaling, while ScaleFL \cite{ilhan2023scalefl} uses both widthwise and depthwise scaling.
\blue{When comparing NeFL to baseline methods, it is important to note that none of the baselines explicitly address inconsistent parameters. Instead, HeteroFL uses static BN layers, while FjORD employs separate BN layers for each submodel. In HeteroFL, the static BN layers does not update BN layers during the training; instead, their statistics are updated in the final round using the entire dataset.
DepthFL and ScaleFL take a different approach by incorporating multiple classifiers corresponding to different exit levels within the model. Although these methods also utilize knowledge distillation, we chose to omit this in our study due to observed performance degradation.}
Instead, our DepthFL and ScaleFL models incorporate downsampling layers to adjust the feature size to match the input sizes of the classifier. Note that the auxiliary bottleneck layers for submodels in DepthFL can be interpreted as parameter decoupling, as discussed in Section \ref{sec:NeFedAvg}.

\subsubsection{Heterogeneity}
To account for system heterogeneity, each client trained one of the submodels in each iteration. We explored two scenarios: one with five submodels ($N_s=5$, where $\boldsymbol{\gamma} =\left[\gamma_1, \gamma_2, \gamma_3, \gamma_4, \gamma_5\right] =\left[0.2,0.4,0.6,0.8,1\right]$) and another with three submodels ($N_s=3$, where $\boldsymbol{\gamma} =\left[\gamma_1, \gamma_2, \gamma_3\right] =\left[0.5,0.75,1\right]$).
\blue{Clients were grouped into tiers based on their basic capabilities, with each tier corresponding to a different submodel range. In the experiments involving five submodels, during each iteration, a client within tier $x$ selects a submodel $k$ uniformly from the set $\{\gamma_k \in \boldsymbol{\gamma}\ | \ {\max(1, x-2)} \leq k \leq \min(x+2, 5)\}$, depending on the dynamically varying system conditions.} % $[\gamma_{\max(1, {x-2})}, \gamma_{\min({x+2}, 5)}]$
Statistical heterogeneity was implemented by label distribution skew following the Dirichlet distribution with concentration parameter $0.5$ \cite{yurochkin2019bayesian, li2021federated}.

\subsubsection{Training details}
Unless otherwise specified, we conducted training for a total of $T=500$ communication rounds ($T=100$ for SVHN) involving $M=100$ clients. In each round, a fraction rate of 0.1 was used, indicating that 10 clients ($|\mathcal{C}_t|=10$) transmit their weights to the server. During the client training process, we used a local batch size of 32 and a local epoch of $E=5$. We employed the stochastic gradient descent (SGD) optimizer \cite{ruder2016overview} without momentum and weight decay. The initial learning rate $\eta$ was set to 0.1 and decreases by a factor of $\frac{1}{10}$ at the halfway point and $\frac{3}{4}$ of the total communication rounds. 

% 이 밑에 부분은 따로 해당 결과에만 넣기
% The experiments in Table \ref{tab:vit-niid} are evaluated with the number of clients is $M=10$, all of whom participate in the NeFL pipeline (with a fraction rate of $1$). The experiment consists of $T=100$ communication rounds, and each client performs local training for a single epoch ($E=1$). We use a cosine annealing learning rate scheduling \cite{loshchilov2017sgdr} with 500 steps of warmup and an initial learning rate 0.03. The input images are resized to a size of $256$  and randomly cropped to a size of $224$ with a padding size of $28$. Note that utilizing layer normalization layers as consistent parameters, as opposed to BN layers that are inconsistent parameters, yields better performance.

% We begin by evaluating the performance of various NeFL methods and comparing them to models that are split widthwise or depthwise. The global model is split into $N_s=5$ submodels, where each submodel $i$ has a splitting ratio of approximately $\gamma_{W_i}\approx \gamma_{D_i}$, except for ResNet18 which has fewer degrees of freedom in terms of depthwise splitting. When we implement depthwise scaling, we heuristically determine which blocks to skip while ensuring the first blocks not to be skipped when the channel size was changed.

\subsection{Performance Evaluation} 
\subsubsection{Comparison with SOTA model scaling FL methods}
\begin{table*}[!t]
\caption{Results of NeFL utilizing \textbf{pre-trained} models as initial weights for CIFAR-10 dataset under \textbf{IID} (left) and \textbf{non-IID} (right) settings. Numbers in parentheses denote the performance difference compared to scratch, with \blue{blue} indicating improvement from pretraining, and \red{red} indicating degradation.} \label{tab:resnet-pretrained}
\centering
\resizebox{.77\textwidth}{!}{
% \resizebox*{!}{0.2 \textheight}{
\begin{tabular}{@{}clcccccccc@{}}
\toprule
\multirow{3}{*}{\textbf{Model}}    & \multirow{3}{*}{\textbf{Method}} & \multicolumn{2}{c}{\textbf{IID}} && \multicolumn{2}{c}{\textbf{non-IID}}       \\ \cmidrule(l){3-7}
      & & \textbf{Worst}   & \textbf{Avg} && \textbf{Worst}   & \textbf{Avg}  \\ \midrule
\multirowcell{7}{Pre-trained\\ ResNet18}
      & HeteroFL \cite{diao2021heterofl}          & 78.26 \red{($\downarrow$ 2.36)} & 84.48 \blue{($\uparrow$ 0.22)} && 71.95 \red{($\downarrow$ 4.30)} & 76.17 \red{($\downarrow$ 3.94)}\\ \cmidrule(l){2-7} 
      & FjORD \cite{horvath2021fjord}            & 86.37 \blue{($\uparrow$ 1.25)} & 88.91 \blue{($\uparrow$ 1.59)} && 81.81 \blue{($\uparrow$ 6.00)} & 81.96 \blue{($\uparrow$ 3.97)} \\ \cmidrule(l){2-7}
      & DepthFL \cite{kim2023depthfl}           & 47.76 \red{($\downarrow$ 17.04)} & 82.86 \blue{($\uparrow$ 0.42)} && 39.78 \red{($\downarrow$ 19.83)} & 67.71 \red{($\downarrow$ 9.18)}\\ \cmidrule(l){2-7}
      & ScaleFL \cite{ilhan2023scalefl}           & 79.34 \red{($\downarrow$ 0.13)} & 86.16 \blue{($\uparrow$ 0.98)} && 69.47 \blue{($\uparrow$ 6.00)} & 78.01 \red{($\downarrow$ 0.48)} \\ \cmidrule(l){2-7}
      &  \GR{0}\textbf{NeFL (ours)} & \GR{0}\textbf{88.61} \blue{($\uparrow$ 1.75)} &\GR{0}\textbf{89.60} \blue{($\uparrow$ 1.72)}&\GR{0}&\GR{0}\textbf{82.91} \blue{($\uparrow$ 1.65)} & \GR{0}\textbf{85.85} \blue{($\uparrow$ 4.14)}\\ \midrule
\multirowcell{7}{Pre-trained\\ ResNet34}
      & HeteroFL \cite{diao2021heterofl}         & 79.97 \blue{($\uparrow$ 0.46)} & 84.34 \blue{($\uparrow$ 1.18)} && 72.33 \red{($\downarrow$ 3.70)} & 78.20 \red{($\downarrow$ 1.43)} \\ \cmidrule(l){2-7} 
      & FjORD \cite{horvath2021fjord}             & 87.08 \blue{($\uparrow$ 1.96)} & 89.37 \blue{($\uparrow$ 2.01)} && 78.20 \blue{($\uparrow$ 3.50)} & 78.90 \blue{($\uparrow$ 2.89)} \\ \cmidrule(l){2-7}
      & DepthFL \cite{kim2023depthfl}           & 52.08 \blue{($\uparrow$ 26.35)} & 83.63 \blue{($\uparrow$ 8.33)} && 42.09 \blue{($\uparrow$ 11.67)} & 79.86 \blue{($\uparrow$ 9.10)} \\ \cmidrule(l){2-7}
      & ScaleFL \cite{ilhan2023scalefl}           & 67.66 \blue{($\uparrow$ 12.94)} & 85.77 \blue{($\uparrow$ 4.72)} && 52.59 \blue{($\uparrow$ 20.25)} & 78.29 \blue{($\uparrow$ 5.89)} \\ \cmidrule(l){2-7}
      &  \GR{0}\textbf{NeFL (ours)} & \GR{0}\textbf{88.36} \blue{($\uparrow$ 0.65)} & \GR{0}\textbf{91.14} \blue{($\uparrow$ 2.12)} &\GR{0}&\GR{0}\textbf{83.62} \blue{($\uparrow$ 2.86)} & \GR{0}\textbf{86.48} \blue{($\uparrow$ 3.18)}\\
      \bottomrule
\end{tabular}}
\vspace{-0.1in}
\end{table*}

\begin{table}[!t]
\centering
\caption{Average FLOPs over five submodels of $\boldsymbol{\gamma}=[0.2,0.4,0.6,0.8,1]$ given a similar number of parameters designed for fair comparison between different scaling methods}
\label{tab:flop}
\begin{tabular}{llccc}
\toprule
\multirow{2.5}{*}{\textbf{Model}}     & \multirow{2.5}{*}{\textbf{Metric}} & \multicolumn{3}{c}{\textbf{Scaling method}}\\ \cmidrule{3-5}
&  & \multicolumn{1}{l}{\textbf{Width/Depth}} & \multicolumn{1}{l}{\textbf{Width}} & \multicolumn{1}{l}{\textbf{Depth}} \\ \midrule
    \multirow{2}{*}{ResNet18}  
        & Param \#  & \multicolumn{1}{c}{6.71M} & \multicolumn{1}{c}{6.71M} & 6.68M \\ 
        & FLOPs & \multicolumn{1}{c}{87.8M} & \multicolumn{1}{c}{85M} & 102M \\ \midrule
    \multirow{2}{*}{ResNet34}  
        & Param \#  & \multicolumn{1}{c}{12.6M} & \multicolumn{1}{c}{12.8M} & 12.9M \\ 
        & FLOPs & \multicolumn{1}{c}{181M} & \multicolumn{1}{c}{176M} & 193M \\ \midrule
    \multirow{2}{*}{ResNet56}  
        & Param \# & \multicolumn{1}{c}{0.51M} & \multicolumn{1}{c}{0.52M} & 0.51M \\ 
        & FLOPs & \multicolumn{1}{c}{530M} & \multicolumn{1}{c}{534M} & 526M \\ \midrule
    \multirow{2}{*}{ResNet110} 
        & Param \# & \multicolumn{1}{c}{1.05M} & \multicolumn{1}{c}{1.06M} & 1.04M \\ 
        & FLOPs & \multicolumn{1}{c}{158M} & \multicolumn{1}{c}{159M} & 234M \\ \bottomrule
\end{tabular}
\vspace{-0.1in}
\end{table}

% For fair comparison across different baselines,
\blue{We designed each submodel to have the parameters and FLOPs as Table \ref{tab:flop}.
Table \ref{tab:other-dataset} presents a comparative analysis of the performance of baselines and NeFL, on four different datasets using two model architectures. The table shows both the worst-case ($\gamma=0.2$) and average accuracies for each method.}

\blue{FjORD outperformed HeteroFL, which uses static BN \cite{diao2021heterofl}. Additionally, the worst-case accuracies of widthwise-scaling methods were higher than methods that include depthwise-scaling methods (i.e., DepthFL and ScaleFL), which struggled with convergence issues.
\textit{In contrast, NeFL consistently outperformed all baseline methods across all datasets, both in terms of worst-case and average accuracies.}
% Specifically, NeFL achieved the highest accuracy demonstrating its effectiveness across diverse data.
The results highlight that NeFL not only improves the overall average performance but also significantly improves the convergence and performance of the weakest submodel.
It is worth noting that NeFL provides a federated averaging method that incorporates widthwise or/and depthwise scaled submodels. Embracing any submodel with different architecture extracted from a single global model enhances flexibility, enabling broader client participation in the FL process.}

\subsubsection{Different model architectures}
\begin{figure}[t]
\centering
\begin{subfigure}[b]{0.49\columnwidth}
\includegraphics[width=1.1\linewidth]{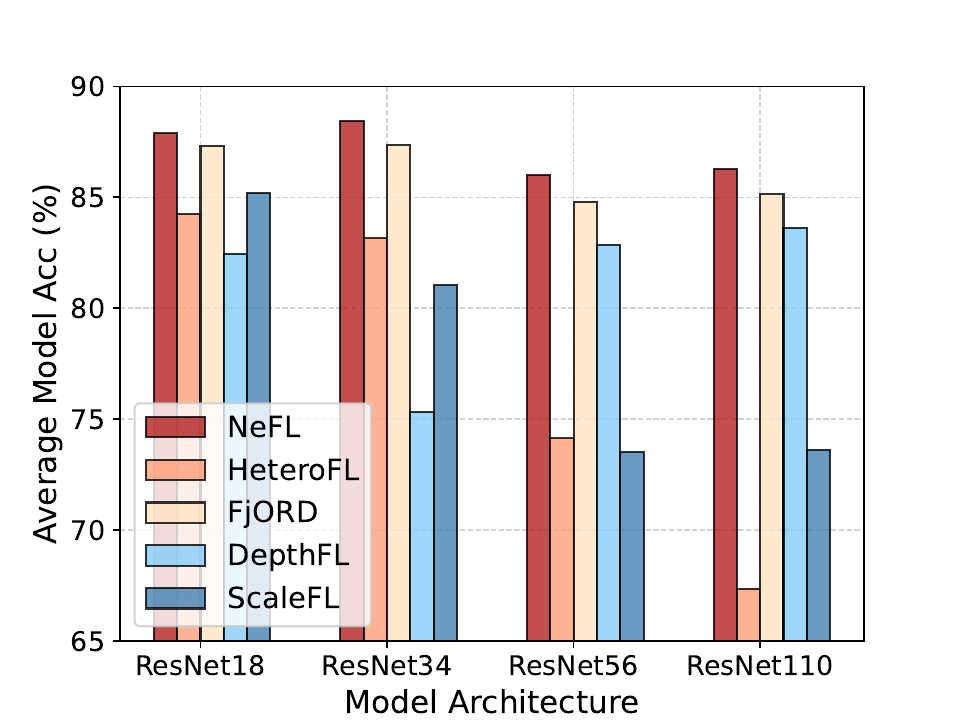}
    \caption{Average accuracy}
    \label{fig:model-size-a}
  \end{subfigure}
\begin{subfigure}[b]{0.49\columnwidth}
\includegraphics[width=1.1\linewidth]{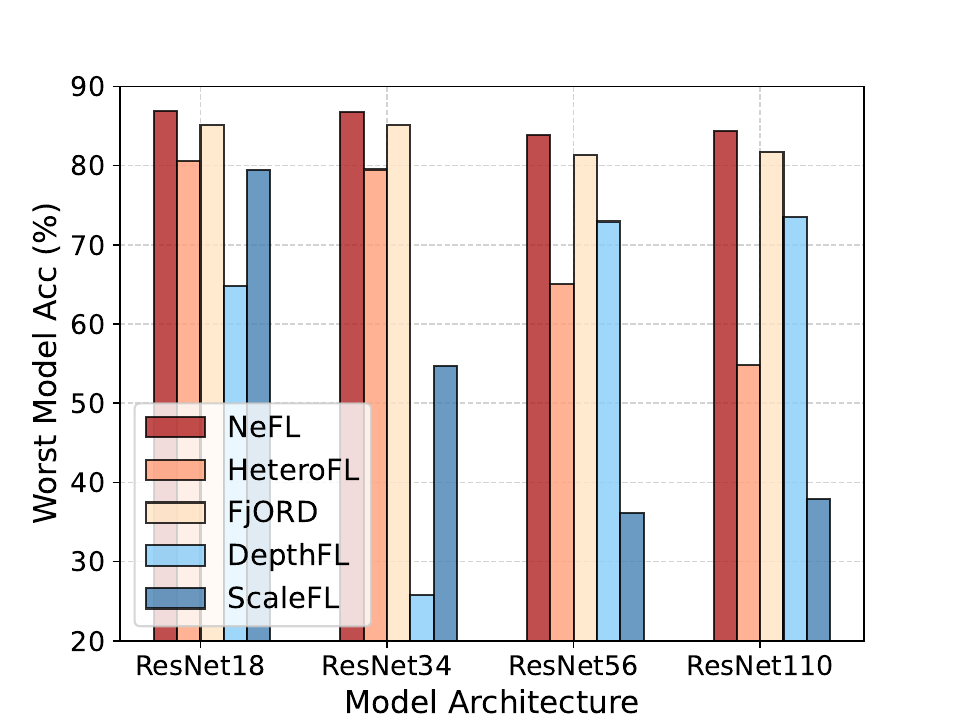}
    \caption{{Worst accuracy}}
    \label{fig:model-size-b}
  \end{subfigure}
  \caption{Results of NeFL with different models on CIFAR-10 dataset}
  \label{fig:model-size}
  \vspace{-0.1in}
\end{figure}

\blue{We evaluated performance across various ResNet architectures, including ResNet18, ResNet34, ResNet56, and ResNet110, as shown in Figure~\ref{fig:model-size}. NeFL consistently outperformed all baseline methods, regardless of the model architecture. While FjORD also showed consistent performance by separating BN layers as inconsistent layers, HeteroFL, DepthFL, and ScaleFL exhibited significant performance drops with certain ResNet configurations. When comparing HeteroFL and DepthFL, HeteroFL performed better on the wider and shallower architectures of ResNet18 and ResNet34, whereas DepthFL showed superior performance on the narrower and deeper architectures of ResNet56 and ResNet110. NeFL’s ability to balance architecture and incorporate inconsistent parameters allowed it to achieve the best performance across all model architectures.}

\subsubsection{Performance enhancement by employing pre-trained models} We investigated the performance of incorporating pre-trained models into NeFL and verified that NeFL is still effective when employing pre-trained models. Recent studies on FL have figured out that FL gets benefits from pre-trained models even more than centralized learning \cite{kolesnikov2020big, chen2023on}, and it motivated us to evaluate the performance of NeFL on pre-trained models. The pre-trained models on ImageNet-1k are loaded from PyTorch \cite{pytorch}. 
Even when \textit{a pre-trained model was trained as a full model without any submodel being trained}, NeFL made better performance with these pre-trained models. 
The results in Table \ref{tab:resnet-pretrained} show that the performance of NeFL had been enhanced through pre-training in both IID and non-IID settings. Meanwhile, baselines such as HeteroFL and DepthFL, which do not have any inconsistent parameters, had no effective performance gains when trained with pre-trained ResNet18 compared to models trained from scratch.

% Table \ref{tab:pretrained} demonstrates that pre-trained models lead to a significant increase in performance. Despite not being pre-trained specifically for NeFL, pre-trained models are still useful in NeFL framework. Upon analyzing ResNet18, we note that the worst model of NeFL-W exhibits a performance gain of 1.9\%, while the best model achieves a gain of 2.63\%. Similarly, for ResNet18, NeFL-WD shows gains of 2.53\% for the worst model and 3.03\% for the best model. Moving on to ResNet34 with NeFL-AD\textsubscript{D}, we observe gains of 2.12\% and 3.65\% for the worst and best model, respectively. The performance gain was particularly notable with NeFL-AD\textsubscript{D} compared to NeFL-WD or NeFL-W. We can attribute these interesting results to the slimming method. % Pre-trained models are not trained with contiguous pruning of widthwise splitting\footnote{Employing pre-trained model was still useful for NeFL-WD and NeFL-W.}.

\subsubsection{Impact of model architecture on statistical heterogeneity} We now present an experiment using ViTs and Wide ResNet \cite{sergey2016wide} on NeFL.
% Meanwhile, the presence of statistical heterogeneity in FL, where non-IID data is distributed across clients, poses a challenge \cite{kairouz2021advances}. This statistical heterogeneity leads to an increased divergence in FL, resulting in significant performance degradation.
Previous studies have examined the effectiveness of ViTs in FL scenarios, and it has been observed that ViTs can effectively alleviate the adverse effects of statistical heterogeneity due to their inherent robustness to distribution shifts \cite{qu2022rethinking}. Building upon this line of research, Table \ref{tab:vit-niid} demonstrates that ViTs outperformed ResNets in NeFL, with the less number of parameters, in both IID and non-IID settings. Particularly in non-IID settings, ViTs exhibited less performance degradation of average performance when compared to IID settings. Note that when comparing the performance gap between IID and non-IID settings, the worst-case ViT submodel experiences more degradation than the worst-case ResNet submodel. Nevertheless, despite this degradation, ViT still maintained higher performance than ResNet. Consequently, we verifed that ViT on NeFL is also effective following the results in \cite{qu2022rethinking}.

We conducted training for $T=100$ communication rounds involving $M=10$ clients with a fraction rate of $1$. Each client conducted local training for a single epoch ($E=1$). The cosine annealing learning rate scheduling \cite{loshchilov2017sgdr}, with a warmup period of 500 steps and an initial learning rate of 0.03 was used. Input images were resized to $256$ and then randomly cropped to $224$, with a padding size of $28$. Notably, utilizing layer normalization layers as consistent parameters resulted in superior performance.

\begin{table}[!t]
\caption{Comparison of different architectures in terms of performance on the CIFAR-10 dataset under both IID and non-IID settings. The initial weights of the global model are obtained from the pre-trained model on ImageNet-1k.} \label{tab:vit-niid}
\centering
% \resizebox{.485\textwidth}{!}{
\resizebox{.48\textwidth}{!}{
\begin{tabular}{@{}cccccccc@{}}
\toprule
\multirow{3}{*}{\textbf{Model}}    & \multirow{3}{*}{\textbf{Param. \#}} & \multicolumn{2}{c}{\textbf{IID}} && \multicolumn{2}{c}{\textbf{non-IID}}        \\ \cmidrule(l){3-7} 
                          && \textbf{Worst}   & \textbf{Avg}  && \textbf{Worst}   & \textbf{Avg} \\ \midrule
{ViT}                     & 86.4M     & 93.02  & 95.96  && 87.56 & 92.74 \\  \midrule
{Wide ResNet101}
                          & {124.8M}     & {90.9} & {91.35} && {87.17} & {87.74}\\  \bottomrule
\end{tabular}}
\vspace{-0.1in}
\end{table}
\begin{figure}[!t]
  \begin{subfigure}[b]{0.49\columnwidth}
    \includegraphics[width=1.1\linewidth]{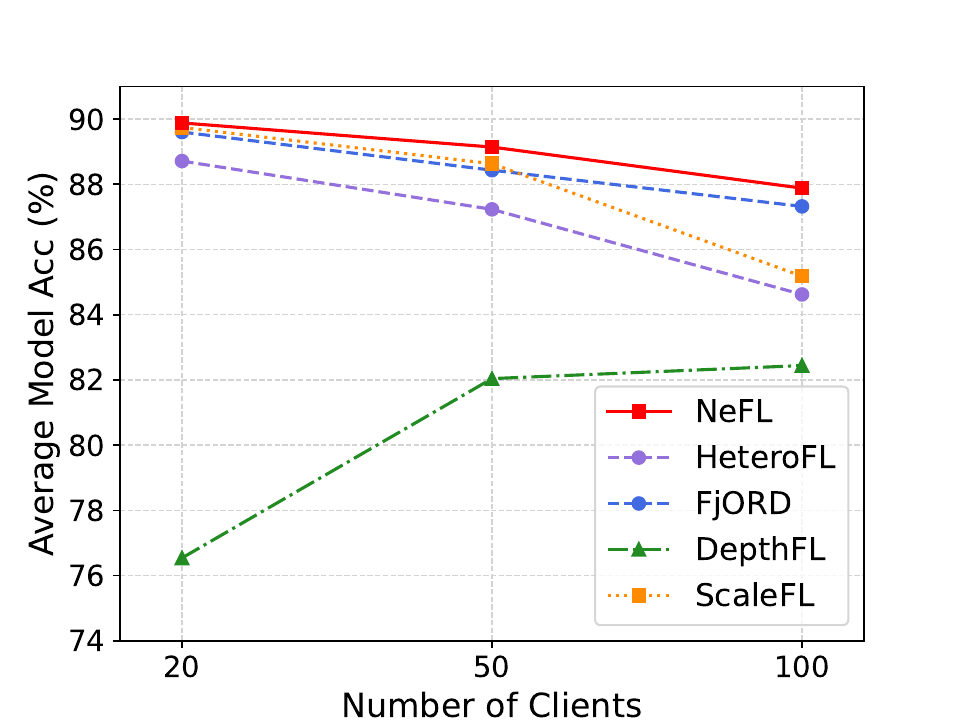}
    \caption{Average accuracy}
    \label{fig:client-num-a}
  \end{subfigure}
  \begin{subfigure}[b]{0.49\columnwidth}
    \includegraphics[width=1.1\linewidth]{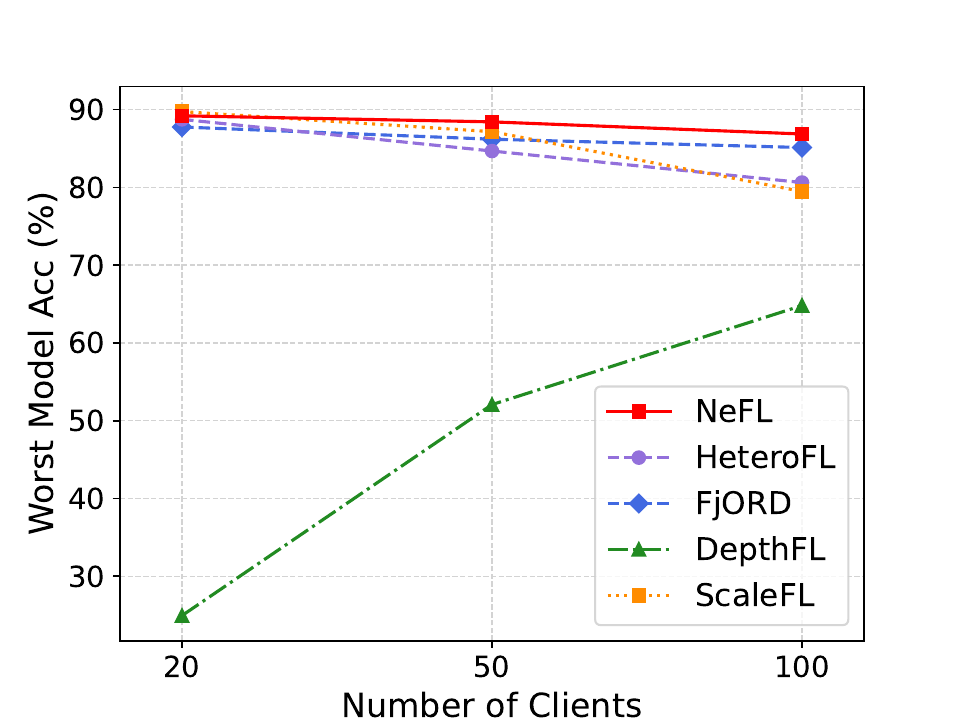}
    \caption{{Worst accuracy}}
    \label{fig:client-num-b}
  \end{subfigure}
  \caption{Results of NeFL with a global model ResNet18 on CIFAR-10 dataset under IID settings across different number of clients}
  \label{fig:client-num}
\vspace{-0.1in}
\end{figure}

\subsubsection{Robustness to number of clients}
We conducted further experiments across different numbers of clients. In Figure \ref{fig:client-num}, we observed that as the number of clients increases, the performance of NeFL as well as baselines degrades. The results align with previous studies \cite{kim2023depthfl,thapa2022splitfed, Wang2020Federated}. The more the number of clients, trained weights deviates further from the weights trained by IID data. While the IID sampling of the training data ensures the stochastic gradient to be an unbiased estimate of the full gradient, the non-IID sampling leads to non-guaranteed convergence and model weight divergence in FL \cite{Li2020fedprox, li2021model, zhao2018federated}. The local clients train their own network with multiple epochs and upload the weights so that the uploaded weights get more deviated.
In this regard, NeFL remained effective across different numbers of clients; however, the performance (e.g., accuracy and convergence) degraded by the data distribution among clients varies more as their number increases.
% We observe the similar tendency is kept with more severe statistical heterogeneity (of smaller Dirichlet distribution parameter).
% across all model sizes.
% When employing NeFL with ViTs, we observe degradation of 0.35\% (8 partitions), 0.79\% (4 partitions) for the global model, and 0.87\% (8 partitions), 9.97\% (4 partitions) for a model with scaling factor $\gamma=0.75$. It is worth mentioning that widthwise scaling, as seen in the findings on pre-trained models (Table \ref{tab:pretrained}), is not as effective as depthwise scaling.

% \paragraph{Additional experiments}
% We provide experimental results of NeFL on other datasets and with different number of clients, along with ablation study in Appendix \ref{appx:addi_exp}. The performance gain of NeFL increases when dealing with more challenging datasets. For example, the performance gain of NeFL with ResNet34 on CIFAR-100 is 7.63\% over baselines. We also verify that NeFL shows the best performance across all different number of clients. In our ablation study, we present the effectiveness of inconsistent parameters including learnable step sizes and the performance comparison of proposed scaling methods.

\subsection{Ablation Study}

\subsubsection{Widthwise \& Depthwise Scaling}
We conducted an ablation study to assess the performance gains achieved by widthwise and depthwise scaling separately and in combination. Specifically, {\it NeFL-W} refers to the widthwise scaling of all submodels through pruning, while {\it NeFL-D} refers to the depthwise scaling of all submodels by skipping certain blocks within a global model, thereby reducing its depth. {\it NeFL-WD}, on the other hand, represents the general case where submodels undergo both widthwise and depthwise scaling.
% indicates that all submodels undergo widthwise scaling via pruning, while {\it NeFL-D} indicates that all submodels undergo depthwise scaling by skipping a subset of blocks within a global model, thereby reducing its depth. On the other hand, {\it NeFL-WD} represents the general case where submodels undergo both widthwise and depthwise scaling.

\begin{table}[!t]
\caption{Ablation study on different types of scaling on CIFAR-10 dataset under IID settings} \label{tab:ablation}
\centering
\begin{subtable}[c]{0.24\textwidth}
\centering
\resizebox*{!}{0.85\textwidth}{
% \resizebox{.45\textwidth}{!}{
\begin{tabular}{@{}clccc@{}}
\toprule
\textbf{Scaling} & {\textbf{Method}} & \textbf{Worst}   & \textbf{Avg}  \\ \midrule
        \multirow{4}{*}{Width}
                & HeteroFL & 80.62 & 84.26 \\ \cmidrule(l){2-4} 
                & FjORD    & 85.12 & 87.32 \\ \cmidrule(l){2-4}
                & \textbf{NeFL-W} & \textbf{85.13} & \textbf{87.36} \\ \midrule
        \multirow{2.5}{*}{Depth}
                & DepthFL & 64.80 & 82.44 \\ \cmidrule(l){2-4}
                & \textbf{NeFL-D} &  \textbf{86.06}   & \textbf{87.94} \\ \midrule
        \multirow{2.5}{*}{W/D}
            & ScaleFL & 79.47 & 85.18 \\ \cmidrule(l){2-4}
            &  \GR{0}\textbf{NeFL-WD} & \GR{0}\textbf{86.86}   & \GR{0}\textbf{87.88} \\ \bottomrule \\
\end{tabular}}
\caption{ResNet18}
\end{subtable}
\begin{subtable}[c]{0.24\textwidth}
\centering
\resizebox*{!}{0.85\textwidth}{
% \resizebox{.45\textwidth}{!}{
\begin{tabular}{@{}clccc@{}}
\toprule
\textbf{Scaling}   & {\textbf{Method}} & \textbf{Worst}   & \textbf{Avg}  \\ \midrule
    \multirow{4}{*}{Width}
              & HeteroFL           & 79.51 & 83.16 \\ \cmidrule(l){2-4} 
              & FjORD              & 85.12 & 87.36 \\ \cmidrule(l){2-4}
              & \textbf{NeFL-W} &  \textbf{85.65}   & \textbf{87.97} \\ \midrule
    \multirow{2.5}{*}{Depth}
              &  DepthFL            & 25.73 & 75.30 \\ \cmidrule(l){2-4}
              & \textbf{NeFL-D} &  \textbf{87.71}   & \textbf{89.02} \\ \midrule
    \multirow{2.5}{*}{W/D}
              &  ScaleFL & 54.72 & 81.05 \\ \cmidrule(l){2-4}
              & \GR{0}\textbf{NeFL-WD} &  \GR{0}\textbf{86.73}   & \GR{0}\textbf{88.42} \\
              \bottomrule \\
\end{tabular}}
\caption{ResNet34}
\end{subtable}
\label{tab:ablation-scale}
\vspace{-0.1in}
\end{table}

\begin{table}[!t]
\caption{Ablation study of learnable step sizes}
\label{tab:ablation-stepsize}
\centering
\begin{tabular}{clcccc}
\toprule
\multirow{2.5}{*}{\textbf{Model}}
& \multirow{2.5}{*}{\textbf{Method}}  & \multicolumn{2}{c}{\textbf{Learnable $s$}} & \multicolumn{2}{c}{\textbf{Fixed $s$}}\\ \cmidrule{3-6}
& & \multicolumn{1}{l}{\textbf{Worst}} & \multicolumn{1}{l}{\textbf{Avg}} & \multicolumn{1}{l}{\textbf{Worst}} & \multicolumn{1}{l}{\textbf{Avg}} \\ \midrule % \multirow{3}{*}{ResNet18}
\multirowcell{3}{Pre-trained\\ ResNet18}
& NeFL-W & 86.10 & 89.13 & 86.37 & 88.91\\
& NeFL-D & 87.13 & 90.00 & 86.95 & 89.77\\
& NeFL-WD & 88.61 & 89.60 & 88.57 & 89.70\\ \midrule
\multirowcell{3}{Pre-trained\\ ResNet34} &  NeFL-W & 87.41 & 89.75 & 87.08 & 89.37\\ &  NeFL-D & 88.36 & 91.14 & 87.95 & 90.79\\ & NeFL-WD & 87.69 & 90.18 & 87.37 & 89.78\\
\bottomrule \\
\end{tabular}
\vspace{-0.1in}
\end{table}
 
% our proposed depthwise (NeFL-D) and widthwise (NeFL-W) scaling methods consistently demonstrated superior performance compared to other depthwise and widthwise scaling baselines. Furthermore, as detailed in the previous section, NeFL-WD outperformed all other baselines.
\blue{Referring to Table~\ref{tab:ablation}, while ScaleFL that scales both widthwise and depthwise (W/D) underperformed compared to widthwise scaling methods such as FjORD and HeteroFL, our NeFL-WD achieved benefits of well-balanced submodels by superior performance. Note that NeFL-WD consistently outperformed other depthwise or/and widthwise scaling baselines.}

\blue{Although NeFL-WD excelled overall, NeFL-D achieved higher average and worst-case accuracies with ResNet34 and higher average accuracy with ResNet18. However, as shown in Table~\ref{tab:flop}, NeFL-D requires more FLOPs than NeFL-WD despite having a similar number of parameters.
Usually depthwise scaled models have more FLOPs than widthwise scaled models while models scaled in both dimensions stay in between. This is due to the model architecture and limited degree-of-freedom (DoF) for scaling submodels. It indicates the another key advantage of scaling in both dimensions is the reduction in FLOPs while maintaining a similar parameter count to depthwise scaling alone. Similar trends are observed in Table~\ref{tab:ablation-stepsize}.}

In summary, our proposed method NeFL, incorporating scaling in both dimensions, yields performance enhancements over other baselines \blue{of any scaling methods and balances the performance between computational complexity.}

\subsubsection{Effect of inconsistent parameters and learnable step size}\label{par:incons-step}

% 원래 내용 
%Referring to the Table \ref{tab:ablation-incons}, the performance improvements of NeFL-WD over NeFL-WD (N/L), NeFL-W over FjORD \cite{horvath2021fjord} and NeFL-D over NeFL-D (N/L) provide {\it the effectiveness of learnable step sizes}. {\it The effectiveness of the inconsistent parameters including learn step sizes} is also verified by NeFL-D over DepthFL \cite{kim2023depthfl} and NeFL-W over HeteroFL \cite{diao2021heterofl}.

% Furthermore, we have a finding that {\it NeFL-D outperforms NeFL-D\textsubscript{O} in most cases}. The rationale comes from the empirical results that trained step sizes are not as large as initial value for NeFL-D\textsubscript{O} that large initial values for NeFL-D\textsubscript{O} degrades the trainability of depthwise-scaled submodels.

\blue{We evaluated the impact of incorporating inconsistent parameters, including learnable step sizes. Referring to Table \ref{tab:ablation}, the effectiveness of inconsistent parameters is evident across different scaling methods. Comparisons between NeFL-W and HeteroFL, NeFL-D and DepthFL, as well as ScaleFL and NeFL-WD, clearly demonstrate the advantage of inconsistent parameters. Notably, the improvement is particularly significant in the worst-case accuracies, especially for methods involving depthwise and combined width/depth scaling.}

Table \ref{tab:ablation-stepsize} further demonstrates the effectiveness of learnable step sizes \blue{when using pre-trained models. For depthwise scaling, NeFL-D showed significant performance gains with learnable step sizes on pre-trained ResNet18 and ResNet34. However, NeFL-W and NeFL-WD experienced slight performance degradation with learnable step sizes on pre-trained ResNet18. Interestingly, the benefits became more pronounced as the model depth increased, particularly with ResNet34.}
Previous studies have shown that merely increasing model depth does not always yield performance improvements over shallower models. However, adding learnable parameters can amplify these benefits \cite{touvron2021going,bachlechner2021rezero}. These results highlight the advantages of integrating inconsistent parameters and learnable step sizes within the NeFL.

\section{Conclusion}\label{sec:conclusion}
In this paper, we introduced {\it Nested Federated Learning (NeFL)}, that addresses the challenges of system heterogeneity. NeFL efficiently divides models into submodels by depthwise and widthwise scaling. The scaling is realized by employing the concept of ODE solver. We also proposed to decouple certain parameters such as inconsistent parameters, which include learnable step sizes, to address the discrepancies between the submodels.
Then, to deal with the inconsistent parameters, we introduced a parameter averaging method that combines FedAvg for inconsistent parameters with NeFedAvg for consistent parameters. Specifically, the inconsistent parameters are aggregated across the clients with the same type of submodel by FedAvg and the consistent parameters are aggregated in nested manner across all clients. The proposed method leads to improved performance and enhanced compatibility with resource-constrained clients. 

A series of experiments have verified that NeFL has performance gains over other baselines, evident in both the average performance across all submodels and specifically in the worst (i.e., smallest) submodel. Furthermore, we have verified that NeFL aligns with recent studies of FL. NeFL showed its performance enhancement through employing pretrained models and NeFL with ViTs outperformed Wide ResNets in statistically heterogeneous settings.
% Furthermore, we also explore NeFL in line with recent studies of FL such as pretraining and statistical heterogeneity. 
One primary limitation of this study was the need for more meticulous consideration in the design of the submodel architectures, specifically in determining which blocks to skip and how many parameters to prune in each block. An interesting future direction would be to apply neural architecture search (NAS) \cite{zoph2017neural} for submodel architecture design.
\blue{Additionally, while we provided a convergence analysis in the supplementary material, further exploration of convergence properties in decentralized or non-smooth optimization scenarios could also be valuable for future research \cite{pu2021distributed}.}

\bibliographystyle{IEEEtran}
\bibliography{HGref}

\newpage
\appendices
\onecolumn

\section{Additional Experiments}
In this section, we provide ablation study for evaluating the performance gains of both width/depthwise scaling, inconsistent parameters, and learnable step size parameters.
{\it NeFL-W} denotes that all submodels are scaled widthwise, {\it NeFL-D} denotes that all submodels are scaled depthwise, and {\it NeFL-WD} denotes that submodels are scaled both widthwise and depthwise.
We further refer to NeFL-D\textsubscript{O}, which has different initial step sizes with NeFL-D.
NeFL-D\textsubscript{O} has step sizes of larger magnitude aligning with the principles of ODE solver, compared to NeFL-D.
NeFL-D scales submodels by skipping a subset of blocks of a global model, thus reducing the depth of the model. NeFL-D compensates for the skipped blocks by using initial step sizes of 1, instead of using larger values. For example, a submodel in NeFL-D without Block $2$ is given the initial step sizes of $s_0=1, s_1=1, s_2=0$ and output after Block $2$ is $\mathbf{Y}_3=\mathbf{Y}_0+F_0+F_1$.
Meanwhile, NeFL-D\textsubscript{O} reduces the size of the global model by skipping a subset of block functions $F(\cdot)$ and gives larger initial step sizes to compensate it. The step sizes are determined based on the number of blocks that are skipped.
For a submodel without Block $2$, initial output after Block $2$ is initially computed as $\mathbf{Y}_3=\mathbf{Y}_0+F_0+2F_1$ given $s_0=1, s_1=2, s_2=0$. We also refer to submodels with no learnable step sizes as N/L (i.e., constant step sizes are kept with given intial values).

The performance comparison between NeFL-WD and NeFL-WD (N/L) as well as the comparison between NeFL-W and FjORD \cite{horvath2021fjord} provides the effectiveness of learnable step sizes and comparison between NeFL-W and HeteroFL \cite{diao2021heterofl} provides the effectiveness of inconsistent parameters including learnable step sizes.
Similarly, the comparison between NeFL-D and NeFL-D (N/L) provides the effectiveness of learnables step sizes and comparison between NeFL-D and DepthFL \cite{kim2023depthfl} provides the effectiveness of inconsistent parameters including learnable step sizes.
We summarized the NeFL with various scaled submodels in Table \ref{tab:NeFLsummary}. We also provide the parameter sizes and average FLOPs of submodels by scaling in Table \ref{tab:flop}.

\begin{table}[!h]
\centering
{ %\scriptsize
\caption{Summarization of NeFL and baselines for ablation study}
\label{tab:NeFLsummary}
\begin{tabular}{lccc}
\toprule
& \textbf{Depthwise scaling} & \textbf{Widthwise scaling} & \textbf{Inconsistency (step sizes)}  \\ \midrule
% NeFL-OD & \checkmark      &     &              \\ \hline
% NeFL-DD & \checkmark      &     &              \\ \hline
DepthFL & \checkmark      &     &              \\ \midrule
FjORD, HeteroFL  & & \checkmark    &           \\ \midrule
NeFL-D & \checkmark      &     & \checkmark   \\ \midrule
NeFL-W  & & \checkmark    & \checkmark         \\ \midrule
NeFL-WD & \checkmark & \checkmark & \checkmark \\ \bottomrule
\end{tabular}}
\end{table}
% \end{wraptable}

\subsection{Learnable step size parameters \& inconsistent parameters.} Referring to the Table \ref{tab:ablation-appx}, the performance improvements of NeFL-WD over NeFL-WD (N/L), NeFL-W over FjORD \cite{horvath2021fjord} and NeFL-D over NeFL-D (N/L) provide {\it the effectiveness of learnable step sizes}. {\it The effectiveness of the inconsistent parameters including learn step sizes} is also verified by NeFL-D over DepthFL \cite{kim2023depthfl} and NeFL-W over HeteroFL \cite{diao2021heterofl}.

Furthermore, we have a finding that {\it NeFL-D consistently outperforms NeFL-D\textsubscript{O} in most cases, regardless of whether the step sizes are constant (N/L) or learnable}. The rationale comes from the empirical results that trained step sizes are not as large as initial value for NeFL-D\textsubscript{O} that large initial values for NeFL-D\textsubscript{O} degrades the trainability of depthwise-scaled submodels.

\subsection{Scaling method} We observed that NeFL-D and NeFL-WD have better performance over widthwise scaling. The performance gap of depthwise scaling over widthwise scaling gets larger for narrow and deep networks. Note that ResNet56 and ResNet110 has smaller (i.e., narrower) channel sizes with more layers (i.e., deeper) than ResNet18 and ResNet34 \cite{he2016deep}.

Furthermore, we could observe that depthwise scaling shows better performance than widthwise scaling.
We evaluated the experiments by similar number of parameters for several scaling methods and depthwise scaling requires more FLOPs than widthwise scaling for ResNet18, ResNet34 and ResNet110 and slightly less FLOPs for ResNet56. Usually depthwise scaled models have more FLOPs than widthwise scaled models while scaling by both widthwise and depthwise models are in between. It is because of the model architecture and limited degree-of-freedom (DoF) of submodels. ResNets consist of convolution layers that have FLOPs of the parameters multiplied by feature sizes. The feature sizes get smaller as forwarding the layers and depthwise scaled submodels that omitted the latter layers make a model to require more FLOPs than widthwise scaled submodels that is scaled across all the layers.

It is worth noting that beyond the performance improvement (including that our proposed scaling method NeFL-W and NeFL-D over baselines in Table \ref{tab:ablation-appx}), {\it NeFL provides the more DoF for widthwise/depthwise scaling} that can be determined by the requirements of clients. It results in more clients to participate in the FL pipeline.

Also refer to Table \ref{tab:110-04} that has different scaling ratio $\gamma$. Note that in this case, FjORD \cite{horvath2021fjord} outperformed NeFL-W. In this case with severe scaling factors (the worst model has 4\% parameters of a global model), step sizes could not compensate the limited number of parameters and degraded the trainability with auxiliary parameters. However, NeFL-WD shows the best performance over other baselines that verify the well-balanced submodels show the better performance than ill-conditioned (too shallow or too narrow) submodels.

\begin{table}[!h]
\caption{Ablation study by NeFL with five submodels for CIFAR-10 dataset under IID settings. We report Top-1 classification accuracies (\%) for the worst-case submodel and the average of the performance of five submodels.} \label{tab:ablation-appx}
\centering
% \resizebox*{!}{0.25\textheight}{
% \resizebox{.45\textwidth}{!}{
\begin{tabular}{@{}clccccc@{}}
\toprule
{\textbf{Model}}    & {\textbf{Method}} & \textbf{Worst}   & \textbf{Avg}  \\ \midrule
\multirow{14}{*}{ResNet18}
      & HeteroFL          & 80.62 & 84.26 \\ \cmidrule(l){2-4} 
      & FjORD             & 85.12 & 87.32 \\ \cmidrule(l){2-4}
      &  \textbf{NeFL-W} &  \textbf{85.13}   & \textbf{87.36} \\ \cmidrule(l){2-4}
      & DepthFL          & 64.80 & 82.44 \\ \cmidrule(l){2-4}
      &  {NeFL-D (N/L)} &  {86.29}   & {88.12} \\ \cmidrule(l){2-4}
      &  {NeFL-D\textsubscript{O} (N/L)} &  {86.24}   & {88.22} \\ \cmidrule(l){2-4}
      &  \textbf{NeFL-D} &  \textbf{86.06}   & \textbf{87.94} \\ \cmidrule(l){2-4}
      &  {NeFL-D\textsubscript{O}} &  {85.98}   & {88.20} \\ \cmidrule(l){2-4}
      &  \textbf{NeFL-WD} &  \textbf{86.86}   & \textbf{87.88} \\ \cmidrule(l){2-4}
      &  {NeFL-WD (N/L)} &  {86.85}   & {88.21} \\
      \bottomrule
\end{tabular}
% }
% \resizebox*{!}{0.25\textheight}{
% \resizebox{.45\textwidth}{!}{
\begin{tabular}{@{}clccccc@{}}
\toprule
{\textbf{Model}}    & {\textbf{Method}} & \textbf{Worst}   & \textbf{Avg}  \\ \midrule
\multirow{14}{*}{ResNet34}
      & HeteroFL           & 79.51 & 83.16 \\ \cmidrule(l){2-4} 
      & FjORD              & 85.12 & 87.36 \\ \cmidrule(l){2-4}
      &  \textbf{NeFL-W} &  \textbf{85.65}   & \textbf{87.97} \\ \cmidrule(l){2-4}
      & DepthFL            & 25.73 & 75.30 \\ \cmidrule(l){2-4}
      &  {NeFL-D (N/L)} &  {87.40}   & {89.12} \\ \cmidrule(l){2-4}
      &  {NeFL-D\textsubscript{O} (N/L)} &  {86.47}   & {88.49} \\ \cmidrule(l){2-4}
      &  \textbf{NeFL-D} &  \textbf{87.71}   & \textbf{89.02} \\ \cmidrule(l){2-4}
      &  {NeFL-D\textsubscript{O}} &  {87.06}   & {88.71} \\ \cmidrule(l){2-4}
      &  \textbf{NeFL-WD} &  \textbf{86.73}   & \textbf{88.42} \\ \cmidrule(l){2-4}
      &  {NeFL-WD (N/L)} &  {86.2}   & {88.16} \\
      \bottomrule
\end{tabular}\vspace{0.05in}\\
% }
% \resizebox*{!}{0.25\textheight}{
% \resizebox{.45\textwidth}{!}{
\begin{tabular}{@{}clccccc@{}}
\toprule
{\textbf{Model}}    & {\textbf{Method}} & \textbf{Worst}   & \textbf{Avg}  \\ \midrule
\multirowcell{14}{Pre-trained\\ResNet18}
      & HeteroFL           & 78.26 & 84.06 \\ \cmidrule(l){2-4} 
      & FjORD              & 86.37 & 88.91 \\ \cmidrule(l){2-4}
      &  \textbf{NeFL-W} &  \textbf{86.1}   & \textbf{89.13} \\ \cmidrule(l){2-4}
      & DepthFL        & 47.76 & 82.85 \\ \cmidrule(l){2-4}
      &  {NeFL-D (N/L)} &  {86.95}   & {89.77} \\ \cmidrule(l){2-4}
      &  {NeFL-D\textsubscript{O} (N/L)} &  {86.24}   & {89.76} \\ \cmidrule(l){2-4}
      &  \textbf{NeFL-D} &  \textbf{87.13}   & \textbf{90.00} \\ \cmidrule(l){2-4}
      &  {NeFL-D\textsubscript{O}} &  {87.02}   & {89.72} \\ \cmidrule(l){2-4}
      &  \textbf{NeFL-WD} &  \textbf{88.61}   & \textbf{89.60} \\ \cmidrule(l){2-4}
      &  {NeFL-WD (N/L)} &  {88.57}   & {89.70} \\                          
      \bottomrule
\end{tabular}
% }
% \resizebox*{!}{0.25\textheight}{
% \resizebox{.45\textwidth}{!}{
\begin{tabular}{@{}clccccc@{}}
\toprule
{\textbf{Model}}    & {\textbf{Method}} & \textbf{Worst}   & \textbf{Avg}  \\ \midrule
\multirowcell{14}{Pre-trained\\ResNet34}
      & HeteroFL      & 79.97 & 84.34 \\ \cmidrule(l){2-4} 
      & FjORD         & 87.08 & 89.37 \\ \cmidrule(l){2-4}
      &  \textbf{NeFL-W} &  \textbf{87.41}   & \textbf{89.75} \\ \cmidrule(l){2-4}
      & DepthFL       & 52.08 & 83.63 \\ \cmidrule(l){2-4}
      &  {NeFL-D (N/L)} &  {87.95}   & {90.79} \\ \cmidrule(l){2-4}
      &  {NeFL-D\textsubscript{O} (N/L)} &  {87.44}   & {90.58} \\ \cmidrule(l){2-4}
      &  \textbf{NeFL-D} &  \textbf{88.36}   & \textbf{91.14} \\ \cmidrule(l){2-4}
      &  {NeFL-D\textsubscript{O}} &  {87.86}   & {90.90} \\ \cmidrule(l){2-4}
      &  \textbf{NeFL-WD} &  \textbf{87.69}   & \textbf{90.18} \\ \cmidrule(l){2-4}
      &  {NeFL-WD (N/L)} &  {87.37}   & {89.78} \\
      \bottomrule
\end{tabular}\vspace{0.05in}\\
% }
% \resizebox*{!}{0.25\textheight}{
% \resizebox{.45\textwidth}{!}{
\begin{tabular}{@{}clccccc@{}}
\toprule
{\textbf{Model}}    & {\textbf{Method}} & \textbf{Worst}   & \textbf{Avg}  \\ \midrule
\multirow{14}{*}{ResNet56}
      & HeteroFL           & 65.09 & 74.13 \\ \cmidrule(l){2-4} 
      & FjORD              & 81.38 & 84.77 \\ \cmidrule(l){2-4}
      &  \textbf{NeFL-W} &  \textbf{82.05}   & \textbf{85.48} \\ \cmidrule(l){2-4}
      & DepthFL            & 72.94 & 86.19 \\ \cmidrule(l){2-4}
      &  {NeFL-D (N/L)} &  {84.38}   & {86.13} \\ \cmidrule(l){2-4}
      &  {NeFL-D\textsubscript{O} (N/L)} &  {83.08}   & {85.59} \\ \cmidrule(l){2-4}
      &  \textbf{NeFL-D} &  \textbf{84.38}   & \textbf{86.13} \\ \cmidrule(l){2-4}
      &  {NeFL-D\textsubscript{O}} &  {81.97}   & {85.37} \\ \cmidrule(l){2-4}
      &  \textbf{NeFL-WD} &  \textbf{83.92}   & \textbf{86.00} \\ \cmidrule(l){2-4}
      &  {NeFL-WD (N/L)} &  {83.68}   & {85.85} \\                          
      \bottomrule
\end{tabular}
% }
% \resizebox*{!}{0.25\textheight}{
% \resizebox{.45\textwidth}{!}{
\begin{tabular}{@{}clccccc@{}}
\toprule
{\textbf{Model}}   & {\textbf{Method}} & \textbf{Worst}   & \textbf{Avg} \\ \midrule
\multirow{14}{*}{ResNet110}
      & HeteroFL           & 54.83 & 67.33 \\ \cmidrule(l){2-4} 
      & FjORD              & 81.70 & 85.16 \\ \cmidrule(l){2-4}
      &  \textbf{NeFL-W} &  \textbf{81.67}   & \textbf{85.32} \\ \cmidrule(l){2-4}
      & DepthFL            & 73.56 & 82.42 \\ \cmidrule(l){2-4}
      &  {NeFL-D (N/L)} &  {85.23}   & {86.34} \\ \cmidrule(l){2-4}
      &  {NeFL-D\textsubscript{O} (N/L)} &  {84}   & {85.97} \\ \cmidrule(l){2-4}
      &  \textbf{NeFL-D} &  \textbf{85.96}   & \textbf{86.66} \\ \cmidrule(l){2-4}
      &  {NeFL-D\textsubscript{O}} &  {82.74}   & {85.66} \\ \cmidrule(l){2-4}
      &  \textbf{NeFL-WD} &  \textbf{84.41}   & \textbf{86.28} \\ \cmidrule(l){2-4}
      &  {NeFL-WD (N/L)} &  {83.58}   & {85.73} \\
      \bottomrule
\end{tabular}
% }
\end{table}

\section{Experimental details}
\subsection{Submodel Architectures}
For the experiments summarized in Tables 
Table \ref{tab:other-dataset}, Table \ref{tab:resnet-pretrained}, Figure \ref{fig:model-size}, Figure \ref{fig:client-num}, Table \ref{tab:ablation} and Table \ref{tab:ablation-appx} we consider five submodels characterized by scaling factors $\boldsymbol{\gamma} = \left[\gamma_1, \gamma_2, \gamma_3, \gamma_4, \gamma_5\right] = \left[0.2, 0.4, 0.6, 0.8, 1\right]$. For Table \ref{tab:110-04}, we use $\boldsymbol{\gamma} = \left[0.04, 0.16, 0.36, 0.64, 1\right]$. In Table \ref{tab:vit-niid}, we evaluate three submodels with $\boldsymbol{\gamma} = \left[0.5, 0.75, 1\right]$. Detailed configurations for the ResNet and ViT submodels are provided in Table \ref{tab:resnet18detail} (ResNet18), Table \ref{tab:resnet34detail} (ResNet34), Table \ref{tab:resnet56detail} (ResNet56), Table \ref{tab:resnet110detail} (ResNet110), Table \ref{tab:wresnet101detail} (Wide ResNet101\_2) and Table \ref{tab:vitdetail} (ViT-B/16).

Submodels can be described using vectors that indicate the presence or absence of residual blocks. For example, the vector [1,1,1], [1,1,1,1], [1,1,1,1,1,1], [1,1,1] represents a fully intact ResNet34. In contrast, the vector [1,1,1], [1,1,1,1], [1,1,1,1,1,1], [1,0,0] represents a depthwise-scaled ResNet34 where the last two blocks have been omitted. Similarly, a ResNet34 with the vector [1,1,1], [1,1,1,1], [1,1,1,1,0,1], [1,0,1] reflects a model depthwise scaled by $\gamma_D = 0.72$. When combined with a uniform widthwise scaling factor of $\gamma_W = 0.77$, the overall model scales down to $\gamma = 0.5$.
In these tables, '1's and '0's indicate the initial values of step sizes, where a step size of zero means that the submodel does not include the corresponding block. For ResNet models, step size parameters are defined for each block, whereas for ViTs, separate step size parameters are applied to the Self-Attention (SA) and Feed-Forward Network (FFN) components.

We omit detailed submodel descriptions for NeFL-W beacuse they are characterized by $\boldsymbol{\gamma}_D=[1,\dots,1]$ and $\boldsymbol{\gamma}_W$ with a target size uniformly across all blocks. On the other hand, submodels in NeFL-D are characterized by $\boldsymbol{\gamma}_W=[1,\dots,1]$ and $\boldsymbol{\gamma}_D$ with a target size. Submodels in NeFL-WD are characterized by target size $\boldsymbol{\gamma}_W \boldsymbol{\gamma}_D$. The omitted blocks for NeFL-D and NeFL-WD are described. 
The corresponding number of parameters and FLOPs for these submodels are provided in Table \ref{tab:flop}.

% \begin{wraptable}{r}{7cm}
\begin{table}[!h] % Different Scaling Ratio $\gamma$
\caption{Performance evaluation with five submodels ($\boldsymbol{\gamma}=[0.04,0.16,0.36,0.64,1]$) for CIFAR-10 dataset under IID settings on ResNet110. We report Top-1 classification accuracies (\%) for the worst-case submodel and the average of the performance of five submodels.} \label{tab:110-04}
\centering
\resizebox*{!}{0.17\textheight}{
\begin{tabular}{@{}clccccc@{}}
\toprule
\multirow{3}{*}{\textbf{Model}}    & \multirow{3}{*}{\textbf{Method}} & \multicolumn{2}{c}{\textbf{Model size}}        \\ \cmidrule(l){3-4} 
      & & \textbf{Worst}   & \textbf{Avg}  \\ \midrule
\multirow{8}{*}{ResNet110}
      & HeteroFL  & 46.58 & 63.62 \\ \cmidrule(l){2-4}
      & FjORD     & 69.61 & 81.46 \\ \cmidrule(l){2-4}
      & \textbf{NeFL-W}                  & 68.27 & 80.98 \\ \cmidrule(l){2-4}  
      & DepthFL     & 11.00 & 53.91 \\ \cmidrule(l){2-4}
      & \textbf{NeFL-D} &  \textbf{75.4}   & \textbf{84.31} \\ \cmidrule(l){2-4}
      & \textbf{NeFL-WD} &  \textbf{76.60}   & \textbf{84.02} \\ \bottomrule
\end{tabular}
}
\end{table}

\begin{table}[!h]
\centering
\caption{Details of $\boldsymbol{\gamma}$ of NeFL-D and NeFL-WD on ResNet18}
\label{tab:resnet18detail}
% \resizebox{.8\textwidth}{!}{
\begin{tabular}{cccc|cccc}
\toprule
\multirowcell{2}{Model\\index} & \multirowcell{2}{Model size\\$\gamma$} & \multirow{2}{*}{$\gamma_W$} & \multirow{2}{*}{$\gamma_D$} & \multicolumn{4}{c}{NeFL-D (ResNet18)} \\ \cline{5-8} & & & & \multicolumn{1}{c|}{Layer 1 (64)} & \multicolumn{1}{c|}{Layer 2 (128)} & \multicolumn{1}{c|}{Layer3 (256)} & Layer 4 (512)\\ \hline
1&0.20  & 1 & 0.20 & \multicolumn{1}{c|}{1,1}   & \multicolumn{1}{c|}{0,0} & \multicolumn{1}{c|}{1,1} & 0,0   \\ \hline
2&0.38  & 1 & 0.38  & \multicolumn{1}{c|}{1,0}   & \multicolumn{1}{c|}{0,0} & \multicolumn{1}{c|}{1,0} & 1,0   \\ \hline
3&0.57  & 1 & 0.57  & \multicolumn{1}{c|}{1,1}   & \multicolumn{1}{c|}{1,1} & \multicolumn{1}{c|}{1,1} & 1,0   \\ \hline
4&0.81  & 1 & 0.81  & \multicolumn{1}{c|}{1,0}   & \multicolumn{1}{c|}{1,1} & \multicolumn{1}{c|}{0,0} & 1,1   \\ \hline
5&1 & 1 & 1 & \multicolumn{1}{c|}{1,1}   & \multicolumn{1}{c|}{1,1} & \multicolumn{1}{c|}{1, 1} & 1,1   \\ \bottomrule
\end{tabular}
% }
% \resizebox{.8\textwidth}{!}{
\\ \vspace{0.05in}
\begin{tabular}{cccc|cccc}
\toprule
\multirowcell{2}{Model\\index} & \multirowcell{2}{Model size\\$\gamma$} & \multirow{2}{*}{$\gamma_W$} & \multirow{2}{*}{$\gamma_D$} & \multicolumn{4}{c}{NeFL-WD (ResNet18)} \\ \cline{5-8}
& & & & \multicolumn{1}{c|}{Layer 1 (64)} & \multicolumn{1}{c|}{Layer 2 (128)} & \multicolumn{1}{c|}{Layer3 (256)} & Layer 4 (512)\\ \hline
1&0.20  & 0.34 & 0.58 & \multicolumn{1}{c|}{1,1}   & \multicolumn{1}{c|}{1,1} & \multicolumn{1}{c|}{1, 1} & 1,0   \\ \hline
2&0.4  & 0.4 & 1  & \multicolumn{1}{c|}{1,1}   & \multicolumn{1}{c|}{1,1} & \multicolumn{1}{c|}{1, 1} & 1,1   \\ \hline
3&0.6  & 0.6 & 1                       & \multicolumn{1}{c|}{1,1}   & \multicolumn{1}{c|}{1,1} & \multicolumn{1}{c|}{1, 1} & 1,1   \\ \hline
4&0.8  & 0.8 & 1                       & \multicolumn{1}{c|}{1,1}   & \multicolumn{1}{c|}{1,1} & \multicolumn{1}{c|}{1, 1} & 1,1   \\ \hline
5&1    & 1 & 1                        & \multicolumn{1}{c|}{1,1}   & \multicolumn{1}{c|}{1,1} & \multicolumn{1}{c|}{1, 1} & 1,1   \\ \bottomrule
\end{tabular}
\end{table}

\begin{table}[h]
\centering
\caption{Details of $\boldsymbol{\gamma}$ of NeFL-D and NeFL-WD on ResNet34}
\label{tab:resnet34detail}
% \resizebox{.8\textwidth}{!}{
\begin{tabular}{cccc|cccc}
\toprule
\multirowcell{2}{Model\\index} &\multirowcell{2}{Model size\\$\gamma$} & \multirow{2}{*}{$\gamma_W$} & \multirow{2}{*}{$\gamma_D$} & \multicolumn{4}{c}{NeFL-D (ResNet34)} \\ \cline{5-8} 
& & & & \multicolumn{1}{c|}{Layer 1 (64)} & \multicolumn{1}{c|}{Layer 2 (128)} & \multicolumn{1}{c|}{Layer3 (256)} & Layer 4 (512)\\ \hline
1&0.23  & 1 & 0.23 & \multicolumn{1}{c|}{1,0,0}   & \multicolumn{1}{c|}{1,0,0,0} & \multicolumn{1}{c|}{1,0,0,0,0,0} & 1,0,0   \\ \hline
2&0.39  & 1 & 0.39  & \multicolumn{1}{c|}{1,1,1}   & \multicolumn{1}{c|}{1,1,1,1} & \multicolumn{1}{c|}{1,1,0,0,0,1} & 1,0,0   \\ \hline
3&0.61  & 1 & 0.61                       & \multicolumn{1}{c|}{1,1,1}   & \multicolumn{1}{c|}{1,1,1,1} & \multicolumn{1}{c|}{1,1,0,0,0,1} & 1,0,1   \\ \hline
4&0.81  & 1 & 0.81                       & \multicolumn{1}{c|}{1,1,1}   & \multicolumn{1}{c|}{1,0,0,1} & \multicolumn{1}{c|}{1,1,0,0,0,1} & 1,1,1   \\ \hline
5&1 & 1 & 1                        & \multicolumn{1}{c|}{1,1,1}   & \multicolumn{1}{c|}{1,1,1,1} & \multicolumn{1}{c|}{1,1,1,1,1,1} & 1,1,1   \\ \bottomrule
\end{tabular}
\\ \vspace{0.05in}
% }
% \resizebox{.8\textwidth}{!}{
\begin{tabular}{cccc|cccc}
\toprule
\multirowcell{2}{Model\\index} &\multirowcell{2}{Model size\\$\gamma$} & \multirow{2}{*}{$\gamma_W$} & \multirow{2}{*}{$\gamma_D$} & \multicolumn{4}{c}{NeFL-WD (ResNet34)} \\ \cline{5-8} 
& & & & \multicolumn{1}{c|}{Layer 1 (64)} & \multicolumn{1}{c|}{Layer 2 (128)} & \multicolumn{1}{c|}{Layer3 (256)} & Layer 4 (512)\\ \hline
1& 0.20  & 0.38 & 0.53 & \multicolumn{1}{c|}{1,1,1}   & \multicolumn{1}{c|}{1,0,0,1} & \multicolumn{1}{c|}{1,0,0,0,0,1} & 1,0,1   \\ \hline
2& 0.40  & 0.63 & 0.64  & \multicolumn{1}{c|}{1,1,1}   & \multicolumn{1}{c|}{1,0,0,1} & \multicolumn{1}{c|}{1,1,1,0,0,1} & 1,0,1   \\ \hline
3& 0.60  & 0.77 & 0.78                       & \multicolumn{1}{c|}{1,1,1}   & \multicolumn{1}{c|}{1,1,1,1} & \multicolumn{1}{c|}{1,1,1,1,0,1} & 1,0,1   \\ \hline
4&0.80  & 0.90 & 0.89                       & \multicolumn{1}{c|}{1,1,1}   & \multicolumn{1}{c|}{1,1,1,1} & \multicolumn{1}{c|}{1,1,1,0,0,1} & 1,1,1   \\ \hline
5&1    & 1 & 1                        & \multicolumn{1}{c|}{1,1,1}   & \multicolumn{1}{c|}{1,1,1,1} & \multicolumn{1}{c|}{1,1,1,1,1,1} & 1,1,1   \\ \bottomrule
\end{tabular}
% }
\end{table}

\begin{sidewaystable}
\centering
\caption{Details of $\boldsymbol{\gamma}$ of NeFL-D and NeFL-WD on ResNet56}
\label{tab:resnet56detail}
\resizebox{0.65\textwidth}{!}{
\begin{tabular}{cccc|cccc}
\toprule
\multirowcell{2}{Model\\index} &\multirowcell{2}{Model size\\$\gamma$} & \multirow{2}{*}{$\gamma_W$} & \multirow{2}{*}{$\gamma_D$} & \multicolumn{3}{c}{NeFL-D (ResNet56)}\\ \cline{5-7} & & & & \multicolumn{1}{c|}{Layer 1 (16)} & \multicolumn{1}{c|}{Layer 2 (32)} & \multicolumn{1}{c}{Layer3 (64)} \\ \hline
1&0.2 & 1 & 0.2
& \multicolumn{1}{c|}{1,1,0,0,0,0,0,0,0}
& \multicolumn{1}{c|}{1,1,0,0,0,0,0,0,0}
& \multicolumn{1}{c}{1,1,0,0,0,0,0,0,0}\\ \hline
2&0.4 & 1 & 0.4
&\multicolumn{1}{c|}{1,1,1,0,0,0,0,0,0}
&\multicolumn{1}{c|}{1,1,1,0,0,0,0,0,0}
&\multicolumn{1}{c}{1,1,1,1,0,0,0,0,0}\\ \hline
3&0.6 & 1 & 0.6
&\multicolumn{1}{c|}{1,1,1,1,0,0,0,0,0}
&\multicolumn{1}{c|}{1,1,1,1,0,0,0,0,0}
&\multicolumn{1}{c}{1,1,1,1,1,1,0,0,0}\\
\hline
4&0.8 & 1 & 0.8                         &\multicolumn{1}{c|}{1,1,1,1,1,1,1,1,1}
&\multicolumn{1}{c|}{1,1,1,1,1,1,1,1,0}
&\multicolumn{1}{c}{1,1,1,1,1,1,1,0,0}\\
\hline
5&1 & 1 & 1&
\multicolumn{1}{c|}{1,1,1,1,1,1,1,1,1}
&\multicolumn{1}{c|}{1,1,1,1,1,1,1,1,1}
&\multicolumn{1}{c}{1,1,1,1,1,1,1,1,1}\\
\bottomrule
\end{tabular}
}\vspace{0.05in}
\resizebox{.65\textwidth}{!}{
\begin{tabular}{cccc|cccc}
\toprule
\multirowcell{2}{Model\\index} &\multirowcell{2}{Model size\\$\gamma$} & \multirow{2}{*}{$\gamma_W$} &\multirow{2}{*}{$\gamma_D$} & \multicolumn{3}{c}{NeFL-D\textsubscript{O} (ResNet56)} \\ \cline{5-7} & & &
& \multicolumn{1}{c|}{Layer 1 (16)} & \multicolumn{1}{c|}{Layer 2 (32)} & \multicolumn{1}{c}{Layer3 (64)} \\ \hline
1&0.2 &1 &0.2
&\multicolumn{1}{c|}{1,8,0,0,0,0,0,0,0}
&\multicolumn{1}{c|}{1,8,0,0,0,0,0,0,0}
&\multicolumn{1}{c}{1,8,0,0,0,0,0,0,0}\\
\hline
2&0.4 &1 &0.4
&\multicolumn{1}{c|}{1,1,7,0,0,0,0,0,0}
&\multicolumn{1}{c|}{1,1,7,0,0,0,0,0,0}
&\multicolumn{1}{c}{1,1,1,6,0,0,0,0,0}\\
\hline
3&0.6 &1 &0.6
&\multicolumn{1}{c|}{1,1,1,6,0,0,0,0,0}&\multicolumn{1}{c|}{1,1,1,6,0,0,0,0,0}&\multicolumn{1}{c}{1,1,1,1,1,4,0,0,0}\\
\hline
4&0.8 &1 &0.8 
&\multicolumn{1}{c|}{1,1,1,1,1,1,1,1,1}
&\multicolumn{1}{c|}{1,1,1,1,1,1,1,2,0}
&\multicolumn{1}{c}{1,1,1,1,1,1,3,0,0}\\
\hline
5&1 &1 &1
&\multicolumn{1}{c|}{1,1,1,1,1,1,1,1,1}
&\multicolumn{1}{c|}{1,1,1,1,1,1,1,1,1}
&\multicolumn{1}{c}{1,1,1,1,1,1,1,1,1}\\ \bottomrule
\end{tabular}
}\vspace{0.05in}
\resizebox{0.65\textwidth}{!}{
\begin{tabular}{cccc|cccc}
\toprule
\multirowcell{2}{Model\\index} &\multirowcell{2}{Model size\\$\gamma$} &\multirow{2}{*}{$\gamma_W$} & \multirow{2}{*}{$\gamma_D$} & \multicolumn{3}{c}{NeFL-WD (ResNet56)}\\ \cline{5-7} & & &
& \multicolumn{1}{c|}{Layer 1 (16)} & \multicolumn{1}{c|}{Layer 2 (32)} & \multicolumn{1}{c}{Layer3 (64)}\\ \hline
1&0.2 & 0.46 & 0.43
& \multicolumn{1}{c|}{1,1,1,1,0,0,0,0,0} &\multicolumn{1}{c|}{1,1,1,1,0,0,0,0,0}
&\multicolumn{1}{c}{1,1,1,1,0,0,0,0,0}\\
\hline
2&0.4&0.61&0.66
&\multicolumn{1}{c|}{1,1,1,1,1,1,0,0,0}
&\multicolumn{1}{c|}{1,1,1,1,1,1,0,0,0}
&\multicolumn{1}{c}{1,1,1,1,1,1,0,0,0}\\
\hline
3&0.6&0.77&0.77
&\multicolumn{1}{c|}{1,1,1,1,1,1,1,0,0}
&\multicolumn{1}{c|}{1,1,1,1,1,1,1,0,0}
&\multicolumn{1}{c}{1,1,1,1,1,1,1,0,0}\\
\hline
4&0.8&0.90&89
&\multicolumn{1}{c|}{1,1,1,1,1,1,1,1,0}
&\multicolumn{1}{c|}{1,1,1,1,1,1,1,1,0}
&\multicolumn{1}{c}{1,1,1,1,1,1,1,1,0}\\
\hline
5&1&1&1
&\multicolumn{1}{c|}{1,1,1,1,1,1,1,1,1}
&\multicolumn{1}{c|}{1,1,1,1,1,1,1,1,1}
&\multicolumn{1}{c}{1,1,1,1,1,1,1,1,1}\\
\bottomrule
\end{tabular}
}
% \end{sidewaystable}
% \begin{sidewaystable}
\vspace{0.05in}
\centering
\caption{Details of $\boldsymbol{\gamma}$ of NeFL-D and NeFL-W on ViT-B/16}\label{tab:vitdetail}
\resizebox{.4\textwidth}{!}{
\begin{tabular}{cccc|c}
\toprule
\multirowcell{2}{Model\\index} &\multirowcell{2}{Model size\\$\gamma$} &\multirow{2}{*}{$\gamma_W$} &\multirow{2}{*}{$\gamma_D$} & \multicolumn{1}{c}{NeFL-D (ViT-B/16)}\\ \cline{5-5} & & & & \multicolumn{1}{c}{Block} \\ \hline
1&0.5 &1 &0.50  & \multicolumn{1}{c}{1,1,1,1,1,1,0,0,0,0,0,0} \\ \hline
2&0.75 &1 &0.75 & \multicolumn{1}{c}{1,1,1,1,1,1,1,1,1,0,0,0} \\ \hline
3&1 &1 &1       & \multicolumn{1}{c}{1,1,1,1,1,1,1,1,1,1,1,1} \\ \bottomrule
\end{tabular}
}
\resizebox{.4\textwidth}{!}{
\begin{tabular}{ccccc|c}
\toprule
\multirowcell{2}{Model\\index} &\multirowcell{2}{Model size\\$\gamma$} &\multirow{2}{*}{$\gamma_W$} &\multirow{2}{*}{$\gamma_D$} & \multicolumn{1}{c}{NeFL-W  (ViT-B/16)}\\ \cline{5-5}
& & & & \multicolumn{1}{c}{Block} \\ \hline
1 &0.5 &0.5 &1  & \multicolumn{1}{c}{1,1,1,1,1,1,1,1,1,1,1,1} \\ \hline
2 &0.75 &0.75 &1 & \multicolumn{1}{c}{1,1,1,1,1,1,1,1,1,1,1,1} \\ \hline
3 &1 &1 &1 & \multicolumn{1}{c}{1,1,1,1,1,1,1,1,1,1,1,1} \\ \bottomrule
\end{tabular}
}
\end{sidewaystable}

\begin{sidewaystable} % rotating
\centering
\caption{Details of $\boldsymbol{\gamma}$ of NeFL on ResNet110}
\label{tab:resnet110detail}
\resizebox{0.85\textwidth}{!}{
\begin{tabular}{ccc|cccc}
\toprule
\multirowcell{2}{Model size\\$\gamma$} &\multirow{2}{*}{$\gamma_W$} &\multirow{2}{*}{$\gamma_D$} & \multicolumn{3}{c}{NeFL-D (ResNet110)}       \\ \cline{4-6} & &
    & \multicolumn{1}{c|}{Layer 1 (16)} & \multicolumn{1}{c|}{Layer 2 (32)} & \multicolumn{1}{c}{Layer3 (64)}\\ \hline
0.2 &1 &0.20 
& \multicolumn{1}{c|}{1,1,1,1,1,1,1,1,1,1,1,1,1,1,1,1,0,0}&\multicolumn{1}{c|}{1,1,1,1,0,0,0,0,0,0,0,0,0,0,0,0,0,0}&\multicolumn{1}{c}{1,1,1,0,0,0,0,0,0,0,0,0,0,0,0,0,0,0}   \\ \hline
0.4 &1 &0.40
&\multicolumn{1}{c|}{1,1,1,1,1,1,1,1,1,1,1,1,1,1,0,0,0,0}&\multicolumn{1}{c|}{1,1,1,1,1,1,1,0,0,0,0,0,0,0,0,0,0,0}&\multicolumn{1}{c}{1,1,1,1,1,1,1,0,0,0,0,0,0,0,0,0,0,0}\\ \hline
0.6 &1 &0.60
&\multicolumn{1}{c|}{1,1,1,1,1,1,1,1,1,1,1,1,1,1,1,1,1,0}&\multicolumn{1}{c|}{1,1,1,1,1,1,1,1,1,1,1,1,1,0,0,0,0,0}&\multicolumn{1}{c}{1,1,1,1,1,1,1,1,1,1,0,0,0,0,0,0,0,0}\\ \hline
0.8 &1 &0.80
&\multicolumn{1}{c|}{1,1,1,1,1,1,1,1,1,1,1,1,1,1,1,1,0,0}&\multicolumn{1}{c|}{1,1,1,1,1,1,1,1,1,1,1,1,1,1,1,1,0,0}&\multicolumn{1}{c}{1,1,1,1,1,1,1,1,1,1,1,1,1,1,0,0,0,0}\\ \hline
1 &1  &1 
&\multicolumn{1}{c|}{1,1,1,1,1,1,1,1,1,1,1,1,1,1,1,1,1,1}&\multicolumn{1}{c|}{1,1,1,1,1,1,1,1,1,1,1,1,1,1,1,1,1,1}&\multicolumn{1}{c}{1,1,1,1,1,1,1,1,1,1,1,1,1,1,1,1,1,1}\\ \bottomrule
\end{tabular}
}\vspace{0.05in}
\resizebox{0.85\textwidth}{!}{
\begin{tabular}{ccc|cccc}
\toprule
\multirowcell{2}{Model size\\$\gamma$} &\multirow{2}{*}{$\gamma_W$} &\multirow{2}{*}{$\gamma_D$} & \multicolumn{3}{c}{NeFL-WD (ResNet110)}  \\ \cline{4-6} & &
& \multicolumn{1}{c|}{Layer 1 (16)} & \multicolumn{1}{c|}{Layer 2 (32)} & \multicolumn{1}{c}{Layer3 (64)} \\ \hline
0.2 &0.46 &0.44
&\multicolumn{1}{c|}{1,1,1,1,1,1,1,0,0,0,0,0,0,0,0,0,0,1}&\multicolumn{1}{c|}{1,1,1,1,1,1,1,0,0,0,0,0,0,0,0,0,0,1}&\multicolumn{1}{c}{1,1,1,1,1,1,1,0,0,0,0,0,0,0,0,0,0,1}\\ \hline
0.4 &0.60 &0.66
&\multicolumn{1}{c|}{1,1,1,1,1,1,1,1,1,1,1,0,0,0,0,0,0,1}&\multicolumn{1}{c|}{1,1,1,1,1,1,1,1,1,1,1,0,0,0,0,0,0,1}&\multicolumn{1}{c}{1,1,1,1,1,1,1,1,1,1,1,0,0,0,0,0,0,1}\\ \hline
0.6 &0.77 &0.77
&\multicolumn{1}{c|}{1,1,1,1,1,1,1,1,1,1,1,1,1,0,0,0,0,1}&\multicolumn{1}{c|}{1,1,1,1,1,1,1,1,1,1,1,1,1,0,0,0,0,1}&\multicolumn{1}{c}{1,1,1,1,1,1,1,1,1,1,1,1,1,0,0,0,0,1}\\ \hline
0.8 &0.90 &0.89                         &\multicolumn{1}{c|}{1,1,1,1,1,1,1,1,1,1,1,1,1,1,1,0,0,1}&\multicolumn{1}{c|}{1,1,1,1,1,1,1,1,1,1,1,1,1,1,1,0,0,1}&\multicolumn{1}{c}{1,1,1,1,1,1,1,1,1,1,1,1,1,1,1,0,0,1}\\ \hline
1 &1 &1
&\multicolumn{1}{c|}{1,1,1,1,1,1,1,1,1,1,1,1,1,1,1,1,1,1}&\multicolumn{1}{c|}{1,1,1,1,1,1,1,1,1,1,1,1,1,1,1,1,1,1}&\multicolumn{1}{c}{1,1,1,1,1,1,1,1,1,1,1,1,1,1,1,1,1,1}\\ \bottomrule
\end{tabular}
}\vspace{0.05in}
\resizebox{0.85\textwidth}{!}{
\begin{tabular}{ccc|cccc}
\toprule
\multirowcell{2}{Model size\\$\gamma$} &\multirow{2}{*}{$\gamma_W$} &\multirow{2}{*}{$\gamma_D$} & \multicolumn{3}{c}{NeFL-D (ResNet110)}       \\ \cline{4-6} & &
    & \multicolumn{1}{c|}{Layer 1 (16)} & \multicolumn{1}{c|}{Layer 2 (32)} & \multicolumn{1}{c}{Layer3 (64)}\\ \hline
0.04 & 1& 0.04
&\multicolumn{1}{c|}{1,0,0,0,0,0,0,0,0,0,0,0,0,0,0,0,0,0}&\multicolumn{1}{c|}{1,0,0,0,0,0,0,0,0,0,0,0,0,0,0,0,0,0}&\multicolumn{1}{c}{1,0,0,0,0,0,0,0,0,0,0,0,0,0,0,0,0,0}\\ \hline
0.16 & 1& 0.16
& \multicolumn{1}{c|}{1,1,1,1,1,0,0,0,0,0,0,0,0,0,0,0,0,0}&\multicolumn{1}{c|}{1,1,1,0,0,0,0,0,0,0,0,0,0,0,0,0,0,0}&\multicolumn{1}{c}{1,1,1,0,0,0,0,0,0,0,0,0,0,0,0,0,0,0}\\ \hline
0.36 & 1& 0.37
&\multicolumn{1}{c|}{1,1,1,1,1,1,0,0,0,0,0,0,0,0,0,0,0,0}&\multicolumn{1}{c|}{1,1,1,1,1,1,0,0,0,0,0,0,0,0,0,0,0,0}&\multicolumn{1}{c}{1,1,1,1,1,1,1,0,0,0,0,0,0,0,0,0,0,0}\\ \hline
0.64 & 1& 0.65
&\multicolumn{1}{c|}{1,1,1,1,1,1,1,1,1,1,1,0,0,0,0,0,0,0}&\multicolumn{1}{c|}{1,1,1,1,1,1,1,1,1,1,1,0,0,0,0,0,0,0}&\multicolumn{1}{c}{1,1,1,1,1,1,1,1,1,1,1,1,0,0,0,0,0,0}\\ \hline
1 & 1& 1 
&\multicolumn{1}{c|}{1,1,1,1,1,1,1,1,1,1,1,1,1,1,1,1,1,1}&\multicolumn{1}{c|}{1,1,1,1,1,1,1,1,1,1,1,1,1,1,1,1,1,1}&\multicolumn{1}{c}{1,1,1,1,1,1,1,1,1,1,1,1,1,1,1,1,1,1}\\ \bottomrule
\end{tabular}
}\vspace{0.05in}
\resizebox{0.85\textwidth}{!}{
\begin{tabular}{ccc|cccc}
\toprule
\multirowcell{2}{Model size\\$\gamma$} &\multirow{2}{*}{$\gamma_W$} &\multirow{2}{*}{$\gamma_D$} & \multicolumn{3}{c}{NeFL-WD (ResNet110)}      \\ \cline{4-6} & & 
    & \multicolumn{1}{c|}{Layer 1 (16)} & \multicolumn{1}{c|}{Layer 2 (32)} & \multicolumn{1}{c}{Layer3 (64)}         \\ \hline
0.04 & 0.26 & 0.16
&\multicolumn{1}{c|}{1,1,1,0,0,0,0,0,0,0,0,0,0,0,0,0,0,0}&\multicolumn{1}{c|}{1,1,1,0,0,0,0,0,0,0,0,0,0,0,0,0,0,0}&\multicolumn{1}{c}{1,1,1,0,0,0,0,0,0,0,0,0,0,0,0,0,0,0}\\ \hline
0.16 & 0.42 & 0.38                       &\multicolumn{1}{c|}{1,1,1,1,1,1,1,0,0,0,0,0,0,0,0,0,0,0}&\multicolumn{1}{c|}{1,1,1,1,1,1,1,0,0,0,0,0,0,0,0,0,0,0}&\multicolumn{1}{c}{1,1,1,1,1,1,1,0,0,0,0,0,0,0,0,0,0,0}\\ \hline
0.36 & 0.59 & 0.61
&\multicolumn{1}{c|}{1,1,1,1,1,1,1,1,1,1,1,0,0,0,0,0,0,0}&\multicolumn{1}{c|}{1,1,1,1,1,1,1,1,1,1,1,0,0,0,0,0,0,0}&\multicolumn{1}{c}{1,1,1,1,1,1,1,1,1,1,1,0,0,0,0,0,0,0}\\ \hline
0.64 & 0.77 & 0.83
&\multicolumn{1}{c|}{1,1,1,1,1,1,1,1,1,1,1,1,1,1,1,0,0,0}&\multicolumn{1}{c|}{1,1,1,1,1,1,1,1,1,1,1,1,1,1,1,0,0,0}&\multicolumn{1}{c}{1,1,1,1,1,1,1,1,1,1,1,1,1,1,1,0,0,0}\\ \hline
1 & 1 & 1
&\multicolumn{1}{c|}{1,1,1,1,1,1,1,1,1,1,1,1,1,1,1,1,1,1}&\multicolumn{1}{c|}{1,1,1,1,1,1,1,1,1,1,1,1,1,1,1,1,1,1}&\multicolumn{1}{c}{1,1,1,1,1,1,1,1,1,1,1,1,1,1,1,1,1,1}\\
\bottomrule
\end{tabular}
}
\vspace{0.3in}
\caption{Details of $\boldsymbol{\gamma}$ of NeFL on Wide ResNet101\_2}
\label{tab:wresnet101detail}
\resizebox{.8\textwidth}{!}{
\begin{tabular}{ccc|cccc}
\toprule
\multirowcell{2}{Model size\\$\gamma$} &\multirow{2}{*}{$\gamma_W$} &\multirow{2}{*}{$\gamma_D$} & \multicolumn{4}{c}{NeFL-D (Wide ResNet101\_2)}
    \\ \cline{4-7} & & 
    & \multicolumn{1}{c|}{Layer 1 (128)} & \multicolumn{1}{c|}{Layer 2 (256)} & \multicolumn{1}{c|}{Layer3 (512)} & Layer 4 (1024)\\ \hline
0.5 &1 &0.51 &
\multicolumn{1}{c|}{1,1,1}   & \multicolumn{1}{c|}{1,1,1,1} & \multicolumn{1}{c|}{1,1,1,1,1,1,1,1,1,0,0,0,0,0,0,0,0,0,0,0,0,0,0} & 1,1,0   \\ \hline
0.75 &1 &0.75                        & \multicolumn{1}{c|}{1,1,1}   & \multicolumn{1}{c|}{1,1,1,1} & \multicolumn{1}{c|}{1,1,1,1,1,1,1,1,1,1,1,1,1,1,0,0,0,0,0,0,0,0,0} & 1,1,1   \\ \hline
1 &1 &1                          & \multicolumn{1}{c|}{1,1,1}   & \multicolumn{1}{c|}{1,1,1,1} & \multicolumn{1}{c|}{1,1,1,1,1,1,1,1,1,1,1,1,1,1,1,1,1,1,1,1,1,1,1} & 1,1,1   \\ \bottomrule
\end{tabular}}
\end{sidewaystable}

\clearpage
\subsection{Pseudocode for Parameter Averaging}

We provide psuedocode for parameter averaging of NeFL. The NeFL server processes locally trained weights $\mathbf{W}=\{\mathbf{w}_i\}_{i\in\mathcal{C}_t}$ from clients, verifying and organizing them according to their corresponding submodels. These weights are sorted into sets $\mathcal{M} = \{ \mathcal{M}_1, \dots, \mathcal{M}_{N_s}\}$, where $ \mathcal{M}_k $ contains the weights from the clients who trained the $k$-th submodel, and $\mathcal{M}$ is a set of uploaded weights sorted by submodels. Then, the consistent parameters are averaged in a nested manner by \texttt{NeFedAvg} while the inconsistent parameters are averaged by FedAvg (\texttt{InconsistentParamAvg}). \texttt{card} evaluates how many times a portion of the consistent parameters (e.g., $\phi_{1,5}\setminus\phi_{1,3}, \phi_{1,3}\setminus\phi_{1,1}$) has been updated by clients.

\begin{minted}[
frame=lines,
baselinestretch=1.1,
fontsize=\footnotesize,
]{python}
import numpy as np

M = [[] for _ in range(Ns)]
for i in range(len(uploaded_weights)):
    for k in range(Ns):
        if uploaded_weights[i]==theta[k]: # parameters of submodel k
            M[k].append(uploaded_weights[i])

def NeFedAvg(M): # consistent parameters
    for block in theta_c[-1]: # global model parameters
        for key in block: # depthwise access by block by block
        num_submodel_uploaded=[], submodel_idx=[], gamma_W_block=[0]
            for k in range(Ns):
                if key in theta_c[k]:
                    submodel_idx.append(k)
                    num_submodels.append(len(M[k]))
                    card = np.cumsum(num_submodels[::-1])[::-1] # cardinality
                    gamma_W_block.append(gamma_W[k])
            for i in range(len(submodel_idx)): # widthwise access
                start = math.ceil(param_size*gamma_W_block[i])
                end = math.ceil(param_size*gamma_W_block[i+1])
                for w in M[submodel_idx[i]]:
                    theta_c_avg[key][start:end]+=w[key][start:end]/card[i]
    return theta_c_avg
    
def InconsistentParamAvg(M): # inconsistent parameters
    for k in range(Ns):
        for key in theta_ic[k]:
            for w in M[k]:
                theta_ic_avg[k][key] += w[key]/len(M[k])
    return theta_ic_avg
\end{minted}

\subsection{Learning Curves}
\begin{figure}[!h]
\centering
\subfloat[{NeFL}]
{\includegraphics[width=0.33\textwidth]{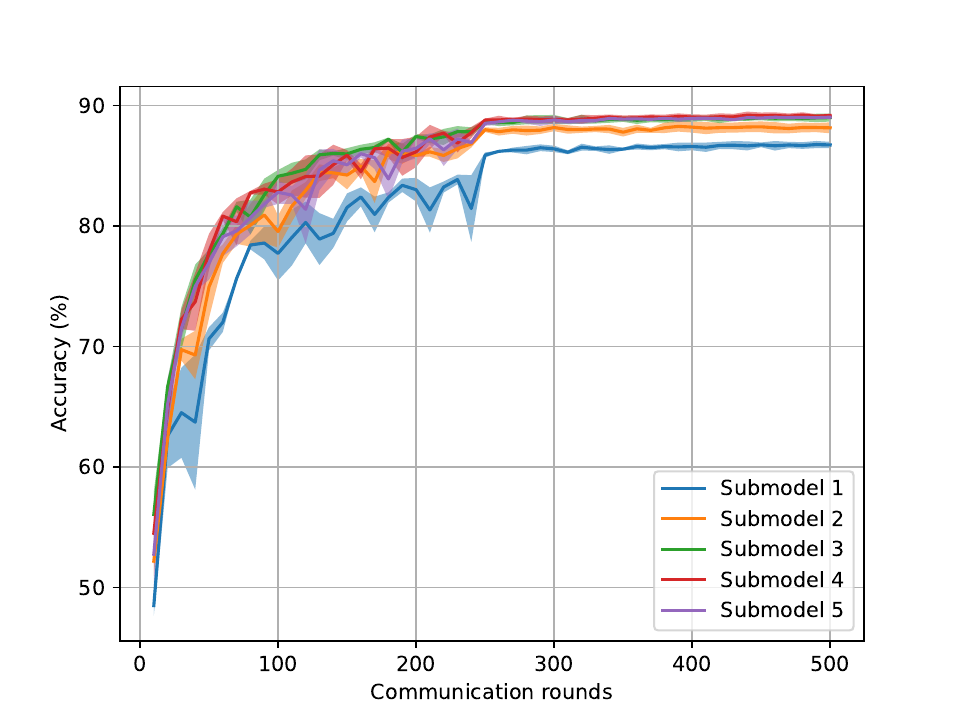}\label{nefl}}
\subfloat[{DepthFL}]{\includegraphics[width=0.33\textwidth]{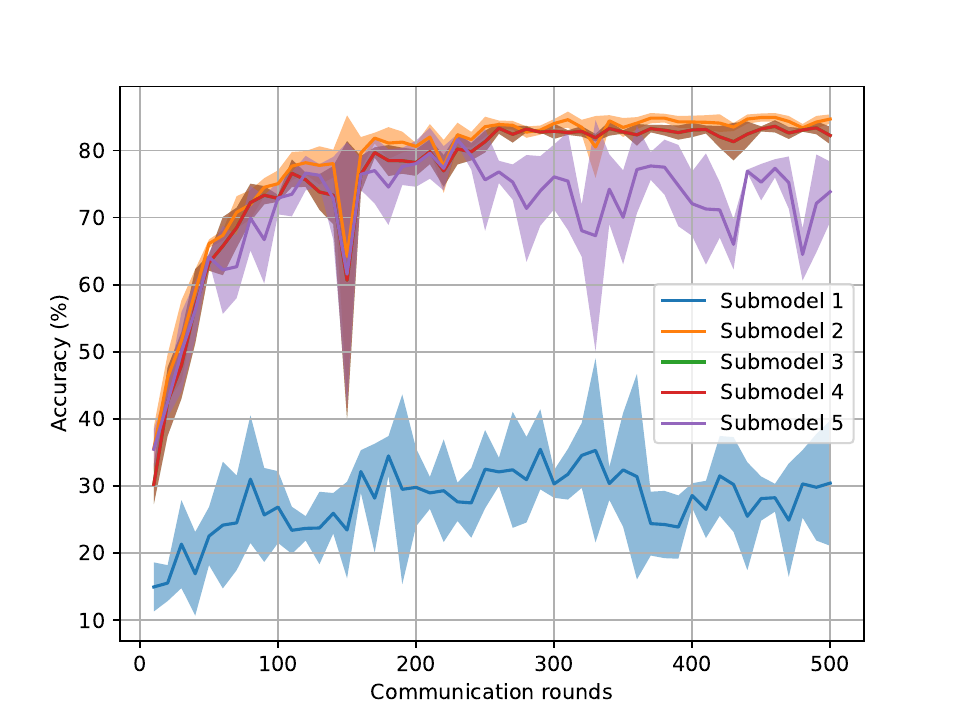}\label{depthfl}}
\subfloat[{HeteroFL}]{\includegraphics[width=0.33\textwidth]{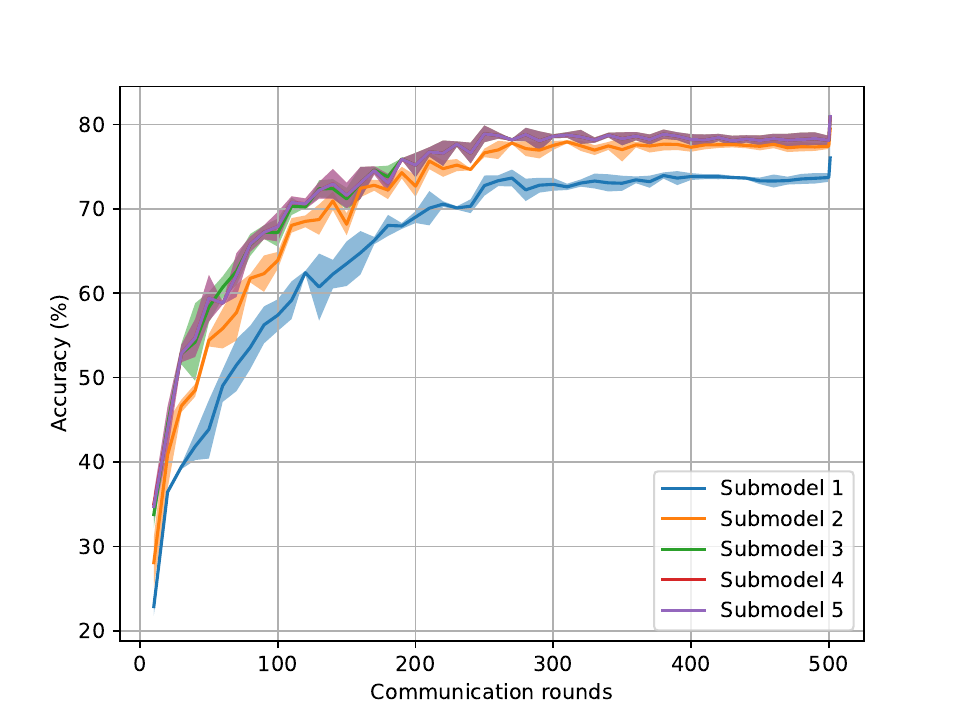}\label{heterofl}}\\
\subfloat[{FjORD}]{\includegraphics[width=0.33\textwidth]{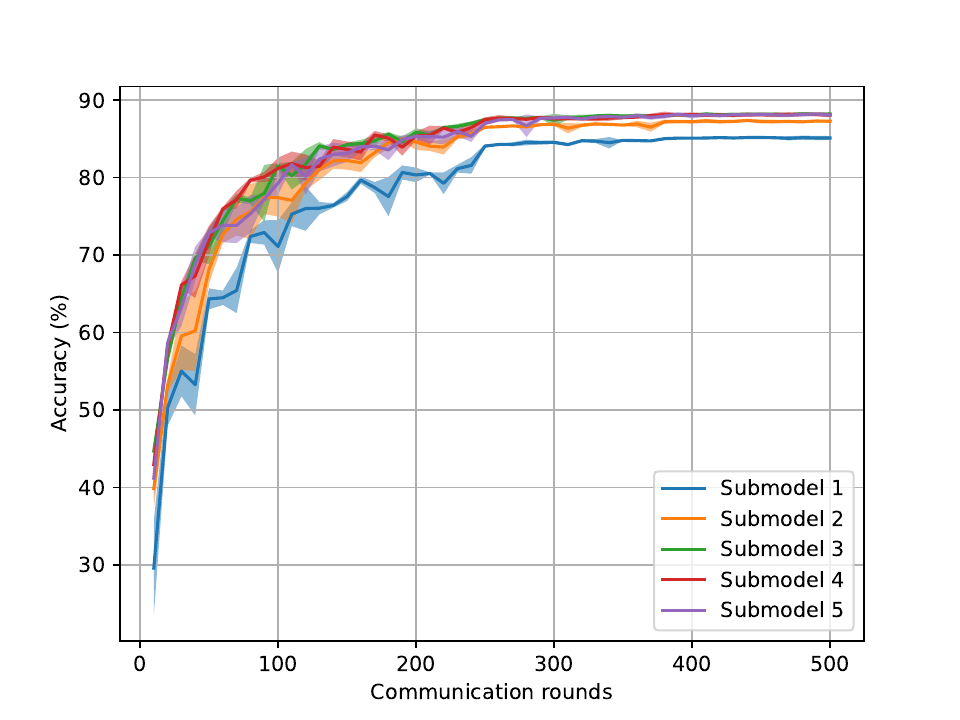}\label{fjord}}
\subfloat[{ScaleFL}]{\includegraphics[width=0.33\textwidth]{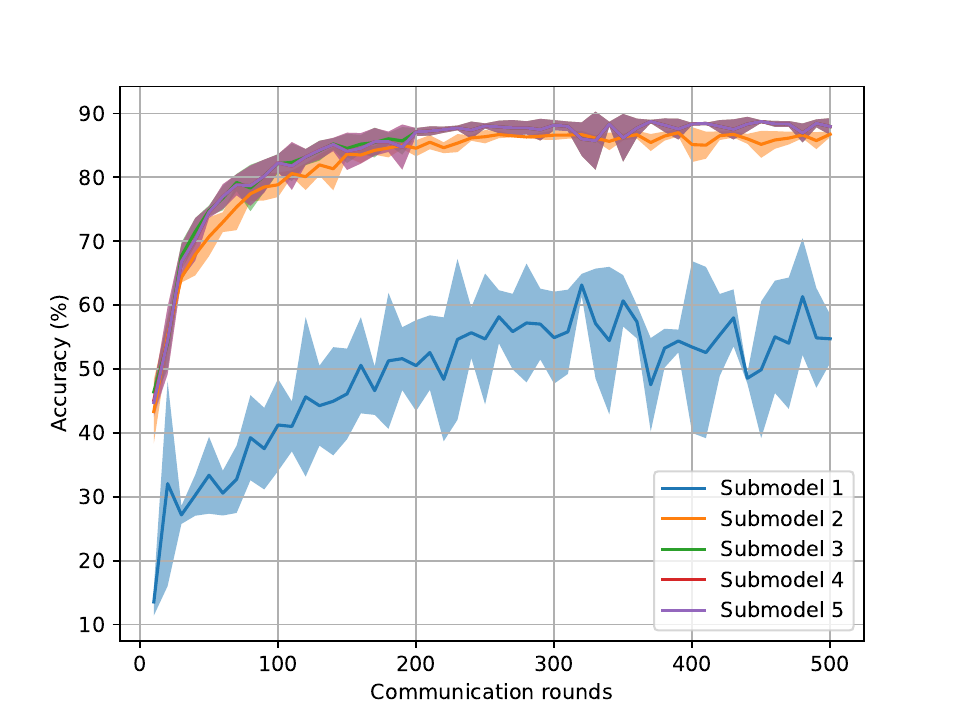}\label{scalefl}}
\caption{Learning curves of NeFL and baselines on CIFAR10 dataset by a global model of ResNet34\label{fig:learning-curves}}
\end{figure}

We have conducted evaluations on all baseline models using 500 communication rounds. As shown in Figure \ref{fig:learning-curves}, we provide the learning curves for NeFL and the baseline models across these communication rounds. The results demonstrate that 500 rounds are sufficient for both NeFL and the baselines to converge.

It is noteworthy that DepthFL and ScaleFL exhibit divergence for specific submodels, and this divergence persists even with an increased number of communication rounds. This suggests that while NeFL achieves stable convergence within the allotted 500 rounds with depthwise scaling, other methods may require different stabilization techniques to achieve similar levels of performance.
It is evident that widthwise scaling methods, such as FjORD and HeteroFL, display good convergence properties comparable to NeFL. Additionally, while HeteroFL shows a noticeable rise in performance at round 501 due to static batch normalization updates, its overall performance remains lower compared to FjORD and NeFL. \newpage

\section{Convergence analysis}

\newtheorem{theorem}{Theorem}
\newtheorem{lemma}{Lemma}
\newtheorem{assumption}{Assumption}
\let\origtheassumption\theassumption

We present convergence analysis of NeFL following a recent work showing the convergence of heterogeneous FL with general model reduction method \cite{zhou2024every}. Our width and depthwise scaling can be interpreted as a form of model pruning, where depthwise scaling can be achieved by applying all-zero masks. We use the concept of a parameter region to denote a specific set of consistent parameters that are defined by model scaling. For each round, the parameters $\theta_t^{(r)}$ associated with region $r$ are exclusively contained within set of submodels represented by $\mathcal{M}_t^{(r)}$. Consequently, $m_{t,i}^{(r)}=\mathbf{1}$ if $i\in\mathcal{M}_t^{(r)}$, and $m_{t,i}^{(r)}=\mathbf{0}$ otherwise. Our submodels are designed to satisfy minimum coverage index $\Gamma^* = \min_{t,r}\Gamma_t^{(r)}$ to be at least $1$ for every block, ensuring every parameter is trained at least once during training rounds. As a result, our algorithm converges, as demonstrated in \cite{zhou2024every}. To clarify the incorporation of learnable step sizes, we provide proofs for the relevant lemmas and demonstrate convergence below. Notations used in this section are summarized in Table~\ref{tab:conv-not}. 

% \usepackage{multirow}
% \usepackage{graphicx}
% \usepackage{graphicx}
\begin{table}[ht]
\centering
\resizebox{0.8\textwidth}{!}{%
\begin{tabular}{ll}
\hline
Symbol               & Explanation                   \\ \hline
$t, T $              & Current, total communication round           \\
$k, E $              & Current, total local epochs           \\
$i, N $              & Local client, total client number                  \\
$r, R $              & Region (or set of parameters), total region \\
$\theta_t$           & Global model at $t$-th round    \\
$m_{t,i}$            & Model reduction mask          \\
$\mathcal{M}_t^{(r)}$ & Parameter set, whose local models contain the $i$th modeling parameter or $r$-th region in round $t$ \\
$\mathbb{P}$         & Model reduction method        \\
$F$                  & Cost function   \\ 
$\nabla F_i(\theta)$ & Local stochastic gradient     \\
$\xi$                & Sampled training data         \\
$D_i$                & Data distribution             \\
$\mathcal{M}^{(r)}$  & Number of local models that containing the $r$-th parameter/region                                         \\
$\gamma$           & Model reduction ratio         \\
$\sigma^{2}$         & Gradient variance bound       \\
$G$                  & Stochastic gradients bound    \\
$\Gamma^{*}$       & Minimum covering index        \\ \hline
\end{tabular}%
}
\caption{}
\label{tab:conv-not}
\end{table}

% \begin{itemize}
%     \item cost function
%     \item mask function 
%     \item depthwise scaling can be considered as mask=0 
%     \item denote parameter for region $r$ as $\theta^{(r)}$
%     \item $\gamma^2=\gamma_W\gamma_D$
% \end{itemize}

\subsection{Assumptions} 
% \begin{itemize}
%     \item Assume 1 block resnet as conv layer 
%     \item conv layer can be represented as matrix multiplication
%     \item step size parameters are bounded
% \end{itemize}

% We make several key assumptions to facilitate the analysis and ensure the validity of our approach. First, we assume a simplified model architecture where a ResNet consists of a single block, where the block is a convolutional layer. This assumption allows us to represent our cost function incorporating learnable step size as $ F(\theta;s)=F(s\theta)$. This formulation is possible because the convolution operation can be represented as matrix multiplication of input and parameter vectors, such that $s(X\theta)=X(s\theta)$. Additionally, we assume that the step size parameters used in the optimization process are bounded, as Fig.~\ref{fig:pruning} shows their value remains within a certain reasonable range.

% \begin{assumption} (Bounded step size).
% Learnable step size parameters are bounded: $\forall s^* \in \mathcal{R}$, we assume there exists $s^*>0$:
% \begin{eqnarray}
%     s^* \le s
% \end{eqnarray}
% \end{assumption}

\begin{assumption} (Smoothness).
Cost functions $F_{1}, \dots ,F_{N}$ are all L-smooth: $\forall \theta,\phi\in \mathcal{R}^d$, for any step size parameters $\forall s_1, s_2 \in \mathcal{R}$ and any $n$, we assume that there exists $L>0$:
\begin{eqnarray}
\| \nabla F_i(s_1\theta) - \nabla F_i(s_2\phi) \| \le L \|s_1\theta-s_2\phi\|.
\end{eqnarray}
\end{assumption}

\begin{assumption}(Model Reduction Noise). 
We assume that for some $\delta^{2} \in[0,1)$ and any $t,i$, the model reduction error is bounded by
\label{assumption:two}
\begin{eqnarray}
\left\|\theta_t-\theta_t \odot m_{t,i}\right\|^{2} \leq \gamma\left\|\theta_t\right\|^{2}.
\end{eqnarray}
\end{assumption}

\begin{assumption} (Bounded Gradient).
The expected squared norm of stochastic gradients is bounded uniformly, i.e., for constant $G>0$ and any $t, i, k$:
\begin{eqnarray}
\mathbb{E}_{\xi_{t,i,k}} \left\| \nabla F_{i}(s_{t,i,k}\theta_{t,i,k},\xi_{t,i,k})\right\|^{2} \leq G.
\end{eqnarray}
\end{assumption}

\begin{assumption} (Gradient Noise for IID data).
Under IID data distribution, for any $t,i,k$, we assume that
\begin{eqnarray}
& \mathbb{E}[\nabla F_i(s_{t,i,k}\theta_{t,i,k}, \xi_{i,k})] = \nabla F(s_{t,i,k}\theta_{t,i,k}) \\
& \mathbb{E}\| \nabla F_i(s_{t,i,k}\theta_{t,i,k}, \xi_{i,k}) - \nabla F(s_{t,i,k}\theta_{t,i,k}) \|^2 \le \sigma^2
\end{eqnarray}
where $\sigma^2>0$ is a constant and $\xi_{i,k}$ are independent samples for different $i,k$.
\end{assumption}

\begin{assumption} (Gradient Noise for non-IID data).
Under non-IID data distribution, we assume that for constant $\sigma^2>0$ and any $t,i,k$:
\begin{eqnarray}
& \mathbb{E}\left[\frac{1}{|\mathcal{M}_t^{(r)}|} \sum_{i\in \mathcal{M}_t^{(r)} } \nabla F_i^{(r)}(s_{t,i,k}\theta_{t,i,k}, \xi_{i,k})\right] = \nabla F^{(r)}(s_{t,i,k}\theta_{t,i,k}) \\
& \mathbb{E}\left\| \frac{1}{|\mathcal{M}_t^{(r)}|} \sum_{i\in \mathcal{M}_t^{(r)} } \nabla F_i^{(r)}(s_{t,i,k}\theta_{t,i,k}, \xi_{i,k}) -\nabla F^{(r)}(s_{t,i,k}\theta_{t,i,k}) \right\|^2 \le \sigma^2.
\end{eqnarray}
%where $\sigma^2>0$ is a constant and $\xi_{i,k})$ are independent samples for different $i,k$.
\end{assumption}

\subsection{Useful Lemmas} 

% Lemma 1 % 
\begin{lemma}
Under Assumption~2 and 3, for any round $t$:
\begin{eqnarray}
\sum_{k=1}^{E} \sum_{i=1}^N \mathbb{E} \| s_{t,i,k-1}\theta_{t,i,k-1} - s_t\theta_t \|^2 \le \frac{2 \eta^2 E^3 N G}{3} + 2 \gamma  NE \cdot \mathbb{E} \|\theta_t \|^2 .
\end{eqnarray}
\end{lemma}
\begin{proof}
\begin{eqnarray}
& & \sum_{k=1}^E \sum_{i=1}^N \mathbb{E} \| s_{t,i,k-1}\theta_{t,i,k-1} - s_t\theta_t \|^2 \nonumber \\
& & \ \ \ \ \ = \sum_{k=1}^E \sum_{i=1}^N  \mathbb{E} \| \left( s_{t,i,k-1}\theta_{t,i,k-1} - s_{t,i,0}\theta_{t,i,0} \right) + \left( s_{t,i,0}\theta_{t,i,0} - s_t\theta_t \right)  \|^2 \nonumber \\
& & \ \ \ \ \ \le \sum_{k=1}^E \sum_{i=1}^N 2 \mathbb{E} \| s_{t,i,k-1}\theta_{t,i,k-1} - s_{t,i,0}\theta_{t,i,0} \|^2 + \sum_{k=1}^E \sum_{i=1}^N 2 \mathbb{E} \| s_{t,i,0}\theta_{t,i,0} - s_t\theta_t \|^2  \label{l1_1}
\end{eqnarray}
where we use $\|\sum_{i=1}^c a_i \|^2\le c \sum_{i=1}^c \|a_i \|^2$. The first term of Eq.~\eqref{l1_1} can be expressed as 

\begin{eqnarray}
&&  \sum_{k=1}^E \sum_{i=1}^N \mathbb{E} \left\|s_{t,i,k-1}\theta_{t, i, k-1}-s_{t,i,0}\theta_{t, i, 0}\right\|^2 \nonumber\\
& & \ \ \ \ \ = \sum_{k=1}^E \sum_{i=1}^N \mathbb{E} \left\|\sum_{j=1}^{k-1}-\eta \nabla F_i\left(s_{t,i,k-1}\theta_{t,i,k-1} ; \xi_{i, j-1}\right) \odot m_{t, i}\right\|^2 \nonumber\\
& & \ \ \ \ \  \le  \sum_{k=1}^E \sum_{i=1}^N(k-1) \sum_{j=1}^{k-1} \mathbb{E} \left\|-\eta \nabla F_i\left(s_{t,i,k-1}\theta_{t,i,k-1} ; \xi_{i, j-1}\right) \odot m_{t, i}\right\|^2  \nonumber\\
& & \ \ \ \ \  \le \sum_{k=1}^E \sum_{i=1}^N(k-1) \sum_{j=1}^{t-1} \eta^2 G \nonumber\\
& & \ \ \ \ \  \le \eta^2 N G \sum_{k=1}^E(k-1)^2 \nonumber\\
& & \ \ \ \ \  \le \frac{ \eta^2 E^3 N G}{3}, \label{l1_2}
\end{eqnarray}
where we use $\|\sum_{i=1}^c a_i \|^2\le c \sum_{i=1}^c \|a_i \|^2$ again in the second line and Assumption~3 in the third line. The second term of Eq.(\ref{l1_1}) is satisfies
\begin{eqnarray}
& \sum_{k=1}^E \sum_{i=1}^N \mathbb{E} \| s_{t,i,0}\theta_{t,i,0} - s_t\theta_t \|^2 &  = \sum_{k=1}^E \sum_{i=1}^N \mathbb{E} \| s_t\theta_t\odot m_{t,i} - s_t\theta_t \|^2 \nonumber \\ 
& & \le \sum_{k=1}^E \sum_{i=1}^N \gamma  \mathbb{E} \|s_t\theta_t \|^2 \nonumber \\
& & = \gamma  NE \cdot \mathbb{E} \|s_t\theta_t \|^2, \label{l1_3}
\end{eqnarray}
where $\theta_{t,i,0}=\theta_t\odot m_{t,i}, \,s_{t,i,0}=s_t$ and Assumption~2 is used in the second line.
\end{proof}

% Lemma 2 %
\begin{lemma}
Under Assumptions~1-3, for any $t$, we have:
\begin{eqnarray}
& & \sum_{r=1}^R \mathbb{E} \left\| \frac{1}{\Gamma_t^{(r)}E} \sum_{k=1}^E \sum_{i\in\mathcal{M}_t^{(r)}} \left[ \nabla F_i^{(r)}({s_{t,i,k-1}\theta_{t,i,k-1}})-\nabla F_i^{(r)} (s_t\theta_t)  \right] \right\|^2  \nonumber \\ 
& &  \ \ \ \ \ \ \ \ \ \ \ \ \ \ \ \ \  \ \ \ \ \ \ \ \ \ \ \ \ \ \ \ \ \ \ \ \ \ \ \le  \frac{2 L^2 \eta^2 E^2 NG}{3\Gamma^*} + \frac{2 L^2\gamma  N}{\Gamma^*}  \mathbb{E} \|s_t\theta_t \|^2
\label{l2_0}.
\end{eqnarray}
\end{lemma}
\begin{proof}
Using the fact $\|\sum_{i=1}^c a_i \|^2\le c \sum_{i=1}^c \|a_i \|^2 $, we have:
\begin{eqnarray}
& & \sum_{r=1}^R \mathbb{E} \left\| \frac{1}{\Gamma_t^{(r)}E} \sum_{k=1}^E \sum_{n\in\mathcal{M}_t^{(r)}} \left[ \nabla F_i^{(r)}({s_{t,i,k-1}\theta_{t,i,k-1}})-\nabla F_i^{(r)} (s_t\theta_t)  \right] \right\|^2 \nonumber \\
& & \ \ \ \ \ \le \sum_{r=1}^R \frac{1}{\Gamma_t^{(r)}E} \sum_{k=1}^E \sum_{n\in\mathcal{M}_t^{(r)}} \mathbb{E} \left\|  \nabla F_i^{(r)}({s_{t,i,k-1}\theta_{t,i,k-1}})-\nabla F_i^{(r)} (s_t\theta_t)  \right\|^2 \nonumber \\
& &  \ \ \ \ \ \le \frac{1}{E\Gamma^{*}}  \sum_{k=1}^E \sum_{i=1}^N \sum_{r=1}^R \mathbb{E} \left\|  \nabla F_i^{(r)}({s_{t,i,k-1}\theta_{t,i,k-1}})-\nabla F_i^{(r)} (s_t\theta_t)  \right\|^2 \nonumber \\
& &  \ \ \ \ \ = \frac{1}{E\Gamma^{*}}  \sum_{k=1}^E \sum_{i=1}^N \mathbb{E} \left\|  \nabla F_i({s_{t,i,k-1}\theta_{t,i,k-1}})-\nabla F_i (s_t\theta_t)  \right\|^2 \nonumber \\
& & \ \ \ \ \ \le \frac{1}{E\Gamma^{*}}  \sum_{k=1}^E \sum_{i=1}^N L^2 \mathbb{E} \left\|  {s_{t,i,k-1}\theta_{t,i,k-1}} - s_t\theta_t  \right\|^2 {\label{l2_2}}, 
\end{eqnarray}
where $\Gamma_t^{(r)}=|\mathcal{M}_t^{(r)}|$ and $\Gamma^*=\min_{t,r} \Gamma_t^{(r)}$. We use Assumption~1 in the last line and we can get Lemma~2 by applying Lemma~1 in the final term. 

\end{proof}

% Lemma 3 %
\begin{lemma}
For IID data distribution under Assumptions~3, for any $t$, we have:
\begin{eqnarray}
\sum_{r=1}^R \mathbb{E} \left\| \frac{1}{\Gamma_t^{(r)}E} \sum_{k=1}^E \sum_{i\in\mathcal{M}_t^{(r)}} \left[ \nabla F_i^{(r)}({s_{t,i,k-1}\theta_{t,i,k-1}},\xi_{i,k-1})-\nabla F^{(r)} (s_{t,i,k-1}\theta_{t,i,k-1})  \right] \right\|^2   \le  \frac{N\sigma^2}{E({\Gamma^*})^2}
\nonumber \label{l3_0}.
\end{eqnarray}
For non-IID data distribution under Assumption~4, for any $q$, we have:
\begin{eqnarray}
\sum_{r=1}^R \mathbb{E} \left\| \frac{1}{\Gamma_t^{(r)}E} \sum_{k=1}^E \sum_{i\in\mathcal{M}_t^{(r)}} \left[ \nabla F_i^{(r)}({s_{t,i,k-1}\theta_{t,i,k-1}},\xi_{i,k-1})-\nabla F^{(r)} (s_{t,i,k-1}\theta_{t,i,k-1})  \right] \right\|^2   \le  \frac{R\sigma^2}{E}
\nonumber \label{l3_0}.
\end{eqnarray}
\end{lemma}
\begin{proof}

For IID distribution, we use the fact that $\nabla F_i^{(r)}({s_{t,i,k-1}\theta_{t,i,k-1}},\xi_{i,k-1})-\nabla F_i^{(r)} (s_{t,i,k-1}\theta_{t,i,k-1})$ is independent and zero mean stemming from independence between $\xi_{i,k-1}$'s and Assumption~4. Applying $\mathbb{E}\|\sum_i \mathbf{x}_i\|^2 = \sum_i \mathbb{E} \|\mathbf{x}_i^2\| $ for zero-mean and independent $\mathbf{x}_i$'s and Assumption~4 to bound the gradient noise, we have: 

\begin{eqnarray}
& & \sum_{r=1}^R \mathbb{E} \left\| \frac{1}{\Gamma_t^{(r)}E} \sum_{k=1}^E \sum_{i\in\mathcal{M}_t^{(r)}} \left[ \nabla F_i^{(r)}({s_{t,i,k-1}\theta_{t,i,k-1}},\xi_{i,k-1})-\nabla F_i^{(r)} (s_{t,i,k-1}\theta_{t,i,k-1})  \right] \right\|^2  \nonumber \\ 
& &  \ \ \ \ \ \le \sum_{r=1}^R \frac{1}{(\Gamma_t^{(r)}E)^2} \sum_{k=1}^E \sum_{i\in\mathcal{M}_t^{(r)}} \mathbb{E} \left\| \nabla F_i^{(r)}({s_{t,i,k-1}\theta_{t,i,k-1}},\xi_{i,k-1})-\nabla F_i^{(r)} (s_{t,i,k-1}\theta_{t,i,k-1})  \right\|^2  \nonumber \\
& & \ \ \ \ \ \le \frac{1}{(E\Gamma^*)^2}\sum_{r=1}^R \sum_{k=1}^E \sum_{i=1}^N \mathbb{E} \left\| \nabla F_i^{(r)}({s_{t,i,k-1}\theta_{t,i,k-1}},\xi_{i,k-1})-\nabla F_i^{(r)} (s_{t,i,k-1}\theta_{t,i,k-1})  \right\|^2  \nonumber \\
& & \ \ \ \ \ = \frac{1}{(E\Gamma^*)^2} \sum_{k=1}^E \sum_{i=1}^N \mathbb{E} \left\| \nabla F_i({s_{t,i,k-1}\theta_{t,i,k-1}},\xi_{i,k-1})-\nabla F_i (s_{t,i,k-1}\theta_{t,i,k-1})  \right\|^2 \nonumber \\
& & \ \ \ \ \ \le \frac{1}{(E\Gamma^*)^2} \cdot EN\sigma^2. 
\end{eqnarray}

Similarly for non-IID data distributions, using the fact that 
$\mathbb{E}\left[\frac{1}{|\mathcal{M}_t^{(r)}|} \sum_{i\in \mathcal{M}_t^{(r)} } \nabla F_i^{(r)}(s_{t,i,k-1}\theta_{t,i,k-1}, \xi_{i,k-1})\right] = \nabla F^{(r)}(s_{t,i,k-1}\theta_{t,i,k-1})$ is an unbiased estimate for any $k$ and independence of samples and applying Assumption~5, we get:
\begin{eqnarray}
& & \sum_{r=1}^R \mathbb{E} \left\| \frac{1}{\Gamma_t^{(r)}E} \sum_{k=1}^E \sum_{i\in\mathcal{M}_t^{(r)}} \left[ \nabla F_i^{(r)}({s_{t,i,k-1}\theta_{t,i,k-1}},\xi_{i,k-1})-\nabla F_i^{(r)} (s_{t,i,k-1}\theta_{t,i,k-1})  \right] \right\|^2  \nonumber \\ 
& &  \ \ \ \ \ \le \frac{1}{E^2} \sum_{r=1}^R  \sum_{k=1}^E  \mathbb{E} \left\| \frac{1}{\Gamma_t^{(r)}} \sum_{i\in\mathcal{M}_t^{(r)}} \nabla F_i^{(r)}({s_{t,i,k-1}\theta_{t,i,k-1}},\xi_{i,k-1})-\nabla F_i^{(r)} (s_{t,i,k-1}\theta_{t,i,k-1})  \right\|^2  \nonumber \\
& & \ \ \ \ \ \le \frac{1}{E^2} \sum_{r=1}^R \sum_{k=1}^E \sigma^2  \nonumber \\
& & \ \ \ \ \ = \frac{R\sigma^2}{E}.
\end{eqnarray}
\end{proof}

% Main proof % 
\subsection{Main proof}
We use L-smoothness property in Assumption~1, which implies 
\begin{eqnarray}
F(s_{t+1}\theta_{t+1}) - F(s_t\theta_t) \le \left< \nabla F(s_t\theta_t) , \  s_{t+1}\theta_{t+1} - s_t\theta_t \right> + \frac{L}{2 }\left\| s_{t+1}\theta_{t+1} - s_t\theta_t \right\|^2.
\label{mainprf1}
\end{eqnarray}
By taking expectations on both sides of Eq.~\eqref{mainprf1}, we get:
\begin{eqnarray}
\mathbb{E} [ F(s_{t+1}\theta_{t+1})] - \mathbb{E}[F(s_t\theta_t)] \le \mathbb{E}\left< \nabla F(s_t\theta_t) , \ s_{t+1}\theta_{t+1} - s_t\theta_t \right> + \frac{L}{2 } \mathbb{E}\left\| s_{t+1}\theta_{t+1} - s_t\theta_t \right\|^2. \label{mm_0}
\end{eqnarray}
We separately find bounds for two terms on RHS and show the convergence as final result. 

% Upperbound 1
\vspace{0.07in}
\paragraph{Bound for $\mathbb{E}\left< \nabla F(s_t\theta_t) , \ s_{t+1}\theta_{t+1} - s_t\theta_t \right>$} For any parameter region $r$, we have:
\begin{eqnarray}
& s_{t+1}\theta_{t+1}^{(r)} - s_t\theta_t^{(r)} & =  \left(\frac{1}{\Gamma_t^{(r)}} \sum_{i\in\mathcal{M}_t^{(r)}} s_{t,i,E}\theta_{t,i,E}^{(r)} \right)- s_t\theta_t^{(r)} \nonumber \\
& & = \frac{1}{\Gamma_t^{(r)}} \sum_{i\in\mathcal{M}_t^{(r)}}  \left[ s_{t,i,0}\theta_{t,i,0}^{(r)} - \sum_{k=1}^E \eta \nabla F_i^{(r)} (s_{t,i,k-1}\theta_{t,i,k-1},\xi_{i,k-1})\cdot m_{t,i}^{(r)} \right] - s_t\theta_t^{(r)} \nonumber \\
& &  = - \frac{1}{\Gamma_t^{(r)}} \sum_{i\in\mathcal{M}_t^{(r)}} \sum_{k=1}^E \eta \nabla F_i^{(r)} (s_{t,i,k-1}\theta_{t,i,k-1},\xi_{i,k-1})\cdot m_{t,i}^{(r)} + s_t\theta_t^{(r)}\cdot m_{t,i}^{(r)} - s_t\theta_t^{(r)} \nonumber \\
& & = - \frac{1}{\Gamma_t^{(r)}} \sum_{i\in\mathcal{M}_t^{(r)}} \sum_{k=1}^E \eta \nabla F_i^{(r)} (s_{t,i,k-1}\theta_{t,i,k-1},\xi_{i,k-1}), \label{m_1}
\end{eqnarray}

Now, we can express $\mathbb{E}\left< \nabla F(s_t\theta_t) , \ s_{t+1}\theta_{t+1} - s_t\theta_t \right>$ by separating the inner product into $R$ regions as 
\begin{eqnarray}
& & \mathbb{E}\left<  \nabla F(s_t\theta_t) , \ s_{t+1}\theta_{t+1} - s_t\theta_t \right> \nonumber \\
& & \ \ \ \ \ \ \ \ \ \ = \sum_{r=1}^R \mathbb{E}\left< \nabla F^{(r)}(s_t\theta_t) , \ s_{t+1}\theta_{t+1}^{(r)} - s_t\theta_t^{(r)} \right> \nonumber \\
% & & \ \ \ \ \ \ \ \ \ \ = \sum_{r=1}^R s \mathbb{E}\left< \nabla F^{(r)}(s_t\theta_t) , \ \theta_{t+1}^{(r)} - \theta_t^{(r)} \right> \nonumber \\
& & \ \ \ \ \ \ \ \ \ \ = \sum_{r=1}^R  \mathbb{E}\left< \nabla F^{(r)}(s_t\theta_t) , \ - \frac{1}{\Gamma_t^{(r)}} \sum_{i\in\mathcal{M}_t^{(r)}} \sum_{k=1}^E \eta \nabla F_i^{(r)} (s_{t,i,k-1}\theta_{t,i,k-1},\xi_{i,k-1}) \right> \nonumber \\
& & \ \ \ \ \ \ \ \ \ \ = \sum_{r=1}^R  \mathbb{E}\left< \nabla F^{(r)}(s_t\theta_t) , \ - \frac{1}{\Gamma_t^{(r)}} \sum_{i\in\mathcal{M}_t^{(r)}} \sum_{k=1}^E \eta \mathbb{E} \left[ \nabla F_i^{(r)} (s_{t,i,k-1}\theta_{t,i,k-1},\xi_{i,k-1}) | s_t\theta_t \right]\right> \nonumber \\
& &  \ \ \ \ \ \ \ \ \ \ = \sum_{r=1}^R  \mathbb{E}\left< \nabla F^{(r)}(s_t\theta_t) , \ - \frac{1}{\Gamma_t^{(r)}} \sum_{i\in\mathcal{M}_t^{(r)}} \sum_{k=1}^E \eta  \nabla F_i^{(r)} (s_{t,i,k-1}\theta_{t,i,k-1})  \right> \nonumber \\
& & \ \ \ \ \ \ \ \ \ \ = - \sum_{r=1}^R  \mathbb{E}\left< \nabla F^{(r)}(s_t\theta_t) , \   \eta E\nabla F^{(r)}(s_t\theta_t) \right> \label{m_2} \\
& & \ \ \ \ \ \ \ \ \ \  \ \ \  -\sum_{r=1}^R  \mathbb{E}\left< \nabla F^{(r)}(s_t\theta_t) , \   \frac{1}{\Gamma_t^{(r)}} \sum_{i\in\mathcal{M}_t^{(r)}} \sum_{k=1}^E \eta \left[ \nabla F_i^{(r)} (s_{t,i,k-1}\theta_{t,i,k-1}) - \nabla F^{(r)}(s_t\theta_t)\right] \right> \nonumber
%& & \ \ \ \ \ \ \ \ \ \ = -\sum_{r=1}^R \mathbb{E} \left\| \nabla F^{(r)}(\theta_t) \right\|^2 
\end{eqnarray}
where we get the final term by splitting the earlier line with respect to a reference point $\eta E \nabla F^{(r)}(s_t\theta_t)$. Next, the first term on the RHS of Eq.~\eqref{m_2} is expressed as 
\begin{eqnarray}
&  - \sum_{r=1}^R \mathbb{E}\left< \nabla F^{(r)}(s_t\theta_t) , \   \eta E \nabla F^{(r)}(s_t\theta_t) \right> & = - \eta E \sum_{r=1}^R \left\| \nabla F^{(r)}(s_t\theta_t)  \right\|^2  \nonumber \\ 
& & = - \eta E \mathbb{E} \left\| \nabla F(s_t\theta_t)  \right\|^2. \label{m_3}
\end{eqnarray}
For the second term on the RHS of Eq.~\eqref{m_2}, we use the inequality $\langle a,b \rangle \le \frac{1}{2} \|a\|^2 +  \frac{1}{2} \|b\|^2 $ for any vectors $a,b$, and it satisfies
\begin{eqnarray}
& & -\sum_{r=1}^R \mathbb{E}\left< \nabla F^{(r)}(s_t\theta_t) , \   \frac{1}{\Gamma_t^{(r)}} \sum_{i\in\mathcal{M}_t^{(r)}} \sum_{k=1}^E \eta \left[ \nabla F_i^{(r)} (s_{t,i,k-1}\theta_{t,i,k-1}) - \nabla F^{(r)}(s_t\theta_t)\right] \right> \nonumber \\
& & \ \ \ \ \ = -\sum_{r=1}^R E\eta \cdot \mathbb{E}\left< \nabla F^{(r)}(s_t\theta_t) , \   \frac{1}{E\Gamma_t^{(r)}} \sum_{i\in\mathcal{M}_t^{(r)}} \sum_{k=1}^E  \left[ \nabla F_i^{(r)} (s_{t,i,k-1}\theta_{t,i,k-1}) - \nabla F^{(r)}(s_t\theta_t)\right] \right> \nonumber \\
& &  \ \ \ \ \ \le  \frac{E\eta}{2} \sum_{r=1}^R  \mathbb{E} \left\| \nabla F^{(r)}(s_t\theta_t) \right\|^2 + \frac{E\eta}{2} \sum_{r=1}^R  \mathbb{E} \left\|  \frac{1}{E\Gamma_t^{(r)}} \sum_{i\in\mathcal{M}_t^{(r)}} \sum_{k=1}^E  \left[ \nabla F_i^{(r)} (s_{t,i,k-1}\theta_{t,i,k-1}) - \nabla F^{(r)}(s_t\theta_t)\right] \right\| \nonumber \\
& & \ \ \ \ \ = \frac{E\eta}{2} \mathbb{E} \left\| \nabla F(s_t\theta_t) \right\|^2 + \frac{E\eta}{2} \left( \frac{2 L^2 \eta^2 E^2 NG}{3\Gamma^*} + \frac{2 L^2\gamma  N}{\Gamma^*}  \mathbb{E} \|s_t\theta_t \|^2 \right) \label{m_4}
\end{eqnarray}
where Lemma~2 is used in the third line. Finally, we plug Eq.~\eqref{m_3} and Eq.~\eqref{m_4} into Eq.~\eqref{m_2} and get:
\begin{eqnarray}
 \mathbb{E}\left< \nabla F(s_t\theta_t) , \ s_{t+1}\theta_{t+1} - s_t\theta_t \right>  \le - \frac{E\eta}{2} \mathbb{E} \left\| \nabla F(s_t\theta_t) \right\|^2 + \frac{E\eta}{2} \left( \frac{2 L^2 \gamma E^2 NG}{3\Gamma^*} + \frac{2 L^2\gamma  N}{\Gamma^*}  \mathbb{E} \|s_t\theta_t \|^2 \right). \label{mm_1}
\end{eqnarray}

\vspace{0.07in}
\paragraph{Bound for $\frac{L}{2 } \mathbb{E}\left\| s_{t+1}\theta_{t+1} - s_t\theta_t \right\|^2$}
We apply Eq.~\eqref{m_1} and this leads to:
\begin{eqnarray}
 & & \frac{L}{2 } \mathbb{E}\left\| s_{t+1}\theta_{t+1} - s_t\theta_t \right\|^2  \nonumber  \\ 
% & &  \ \ \ \ \le \frac{s^2 L}{2 } \mathbb{E}\left\| \theta_{t+1} - \theta_t \right\|^2  \nonumber  \\ 
 & &  \ \ \ \ = \frac{L}{2 } \mathbb{E}\left\| \frac{1}{\Gamma_t^{(r)}} \sum_{i\in\mathcal{M}_t^{(r)}} \sum_{k=1}^E \eta \nabla F_i^{(r)} (s_{t,i,k-1}\theta_{t,i,k-1},\xi_{i,k-1})\right\|^2  \nonumber \\
& & \ \ \ \ \le \frac{3 L}{2 } \mathbb{E}\left\| \frac{1}{\Gamma_t^{(r)}} \sum_{i\in\mathcal{M}_t^{(r)}} \sum_{k=1}^E \eta \left[ \nabla F_i^{(r)} (s_{t,i,k-1}\theta_{t,i,k-1},\xi_{i,k-1}) - \nabla F_i^{(r)} (s_{t,i,k-1}\theta_{t,i,k-1})\right]\right\|^2 \nonumber \\
& & \ \ \ \ \ \ \ \ + \frac{3 L}{2} \mathbb{E}\left\| \frac{1}{\Gamma_t^{(r)}} \sum_{i\in\mathcal{M}_t^{(r)}} \sum_{k=1}^E \eta  \left[ \nabla F_i^{(r)} (s_{t,i,k-1}\theta_{t,i,k-1}) - \nabla F_i^{(r)} (s_t\theta_t)  \right] \right\|^2 \nonumber \\
& & \ \ \ \ \ \ \ \ + \frac{3 L}{2} \mathbb{E}\left\| \frac{1}{\Gamma_t^{(r)}} \sum_{i\in\mathcal{M}_t^{(r)}} \sum_{k=1}^E \eta \nabla F_i^{(r)} (s_t\theta_t)   \right\|^2, \label{mm_2}
%-F^{(r)} (\theta_t)
\end{eqnarray}
where we use the inequality $\|\sum_{i=1}^c a_i \|^2\le c \sum_{i=1}^c \|a_i \|^2 $ in the second line. Here, the third term on the RHS of Eq.~\eqref{mm_2} can be further simplified as:
\begin{eqnarray}
& \frac{3L}{2} \mathbb{E}\left\| \frac{1}{\Gamma_t^{(r)}} \sum_{i\in\mathcal{M}_t^{(r)}} \sum_{k=1}^E \eta \nabla F_i^{(r)} (s_t\theta_t)   \right\|^2 & \le \frac{3LE^2\eta^2}{2}\sum_{r=1}^R \mathbb{E}\|  \nabla F^{(r)} (s_t\theta_t)  \|^2  \nonumber \\
&  & = \frac{3LE^2\eta^2}{2} \mathbb{E}\| \nabla F (s_t\theta_t)  \|^2. \label{mm_3}
\end{eqnarray}

Since the first and second terms of Eq.~\eqref{mm_2} are bounded by Lemma~2 and Lemma~3, respectively, we have:
\begin{eqnarray}
 & \frac{L}{2 } \mathbb{E}\left\| s_{t+1}\theta_{t+1} - s_t\theta_t \right\|^2 
 & \le \frac{3LEN\eta^2\sigma^2}{2({\Gamma^*})^2} {\rm \ (for \ IID) \ or \ }  \frac{3LER\eta^2\sigma^2}{2} {\rm \ (for \ non-IID)} \nonumber  \\ 
 & & \ \ \ \ + \frac{L^3 \eta^4 E^4 NG}{\Gamma^*} + \frac{3 L^3 E^2 \eta^2 \gamma N}{\Gamma^*} \mathbb{E} \|s_t\theta_t \|^2 \nonumber \\
 & & \ \ \ \ + \frac{3LE^2\eta^2}{2} \mathbb{E}\| \nabla F (s_t\theta_t)  \|^2. \label{mm_4}
\end{eqnarray}

% \begin{eqnarray}
% {\theta}_{t,i,k} =  {\theta}_{t,i,k-1} - \eta \nabla  F_{i}(s_{t,i,k-1}{\theta}_{t,i,k-1};\xi_{i,k-1})
% \end{eqnarray}

% Combine two bounds 
\vspace{0.07in}
\noindent {\bf Combining the two Upperbounds}. We first sum both sides of Eq.~\eqref{mm_0} over round $t=1,\ldots,T$ and we obtain:
\begin{eqnarray}
 & & \mathbb{E} [ F(s_{T+1}\theta_{T+1})] - \mathbb{E} [ F(s_0\theta_{0})] \nonumber \\
& &  \ \ \ \ \ \ \ \  = \sum_{t=1}^T \mathbb{E} [ F(s_{t+1}\theta_{t+1})] - \sum_{t=1}^T \mathbb{E}[F(s_t\theta_t)] \nonumber \\
 & &   \ \ \ \ \ \ \ \ \le \sum_{t=1}^T \mathbb{E}\left< \nabla F(s_t\theta_t) , \ s_{t+1}\theta_{t+1} - s_t\theta_t \right> + \sum_{t=1}^T  \frac{L}{2 } \mathbb{E}\left\| s_{t+1}\theta_{t+1} - s_t\theta_t \right\|^2. \label{mm_5}
\end{eqnarray}
For IID distribution, by plugging in Eq.~\eqref{mm_1} and Eq.~\eqref{mm_4}, we get:
\begin{eqnarray}
 & & \mathbb{E} [ F(s_{T+1}\theta_{T+1})] - \mathbb{E} [ F(s_0\theta_{0})] \nonumber \\
& &  \ \ \ \ \ \ \ \  \le - \frac{E\eta}{2} \left( 1-3LE\eta \right) \sum_{t=1}^T \mathbb{E}\| \nabla F(s_t\theta_t)\|^2  \nonumber \\ 
& & \ \ \ \ \ \ \ \  \ \ \ \ + \frac{E\eta T}{2} \left( \frac{2 E^2 L^2 \eta^2 NG}{3\Gamma^*} +\frac{3LN\eta \sigma^2}{(\Gamma^*)^2} + \frac{2 L^3\eta^3E^3NG}{\Gamma^*} \right)   \nonumber \\
& & \ \ \ \ \ \ \ \  \ \ \ \ +\frac{E\eta}{2} \left( \frac{2 L^2\gamma N}{\Gamma^*} +\frac{6 L^3 E\eta \gamma N}{\Gamma^*}  \right) \sum_{t=1}^T \mathbb{E} \| s_t\theta_t \|^2.
\end{eqnarray}
We set learning rate $\eta\le 1/(6LE)$ and use $\mathbb{E} [ F(s_{T+1}\theta_{T+1})]\ge 0$, then we obtain:
\begin{eqnarray}
& \frac{E\eta}{4} \sum_{t=1}^T \mathbb{E}\| \nabla F(s_t\theta_t)\|^2 & \le \mathbb{E} [ F(s_0\theta_{0})] + \frac{E\eta T}{2}\left( \frac{3LN\eta \sigma^2}{(\Gamma^*)^2} +  \frac{E^2 L^2\eta^2NG}{\Gamma^*}\right) \nonumber \\
& & \ \ \ \ + \frac{E\eta}{2}\left( \frac{3 L^2\gamma N}{\Gamma^*} \right)\sum_{t=1}^T \mathbb{E}  \| s_t\theta_t \|^2.
\end{eqnarray}
Dividing both sides by $(E\eta T)/4$ and choosing $\eta \le 1/(E\sqrt{T})$ leads to:
\begin{eqnarray}
& \frac{1}{T} \sum_{t=1}^T \mathbb{E}\| \nabla F(s_t\theta_t)\|^2 & \le \frac{ 4\mathbb{E} [ F(s_0\theta_{0})]}{\sqrt{T}} +  \frac{6LN \sigma^2}{E\sqrt{T}(\Gamma^*)^2}  \\
& & \ \ \ \ + \frac{2L^2NG}{T\Gamma^*} + \frac{6 L^2\gamma N}{\Gamma^*} \cdot \frac{1}{T}\sum_{t=1}^T \mathbb{E}  \| s_t\theta_t \|^2 \nonumber \\
& & = \frac{A_0}{\sqrt{T}}+  \frac{B_0}{E\sqrt{T}} + \frac{C_0}{T} +\frac{D_0}{\Gamma^*} \cdot \frac{1}{T}\sum_{t=1}^T \mathbb{E}  \| s_t\theta_t \|^2,
\end{eqnarray}
where $A_0=4\mathbb{E} [ F(s_0\theta_{0})]$,   $B_0=6LN\sigma^2/(\Gamma^*)^2$, $C_0 = 2L^2NG/\Gamma^*$, and $D_0=6 L^2\gamma^2 N$. This completes the proof of Theorem~1.

For non-IID data distribution, we plug the two upperbounds into Eq.~\eqref{mm_5} and re-arrange the terms. We follow a similar procedure and choose learning rate $\eta \le 1/(6LE)$ and $\eta \le 1/E\sqrt{T}$, then we have:
\begin{eqnarray}
\frac{1}{T} \sum_{t=1}^T \mathbb{E}\| \nabla F(s_t\theta_t)\|^2 \le \frac{A_0}{\sqrt{T}} + \frac{B_1}{E\sqrt{T}} + \frac{C_0}{T} + \frac{D_0}{\Gamma^*} \cdot \frac{1}{T}\sum_{t=1}^T \mathbb{E}  \| s_t\theta_t \|^2,
\end{eqnarray}
where $B_1=6LR\sigma^2$.

% @inproceedings{zhou2024every,
%   title={Every parameter matters: Ensuring the convergence of federated learning with dynamic heterogeneous models reduction},
%   author={Zhou, Hanhan and Lan, Tian and Venkataramani, Guru Prasadh and Ding, Wenbo},
%   booktitle={Advances in Neural Information Processing Systems (NeurIPS)},
%   year={2024}
% }

 

% \end{IEEEbiography}



\vfill

\end{document}